\documentclass[pdflatex,sn-mathphys-num]{sn-jnl}% Math 

\usepackage{amsmath,graphicx,fullpage}
\usepackage{booktabs}
\usepackage{multirow}
\usepackage{array}
\usepackage{enumerate}
\usepackage{pdflscape}
\usepackage{color,soul}
\usepackage{xcolor}
\definecolor{green}{rgb}{0,0.5,0}
\definecolor{red}{rgb}{0.8,0,0}
\definecolor{blue}{rgb}{0,0,0.8}
\definecolor{orange}{rgb}{0.85,0.35,0}
\usepackage{url}
\newcolumntype{x}[1]{%
>{\centering\hspace{0pt}}p{#1}}%

\usepackage{fancyhdr}
\begin{document}

\title[Advances in Artificial Intelligence: A  Review for the Creative Industries]{Advances in Artificial Intelligence: A  Review for the Creative Industries}


\author*[1]{\fnm{Nantheera} \sur{Anantrasirichai}}\email{N.Anantrasirichai@bristol.ac.uk}
\author[1]{\fnm{Fan} \sur{Zhang}}
\author[1]{\fnm{David} \sur{Bull}}
\affil[1]{\orgname{Visual Information Laboratory, University of Bristol, Bristol, UK}}

%%==================================%%

\abstract{{ Artificial intelligence (AI) has undergone transformative advances since 2022, particularly through generative AI, large language models (LLMs), and diffusion models, fundamentally reshaping the creative industries. However, existing reviews have not comprehensively addressed these recent breakthroughs and their integrated impact across the creative production pipeline. This paper addresses this gap by providing a systematic review of AI technologies that have emerged or matured since our 2022 review, examining their applications across content creation, information analysis, post-production enhancement, compression, and quality assessment. We document how transformers, LLMs, diffusion models, and implicit neural representations have established new capabilities in text-to-image/video generation, real-time 3D reconstruction, and unified multi-task frameworks—shifting AI from support tool to core creative technology. Beyond technological advances, we analyze the trend toward unified AI frameworks that integrate multiple creative tasks, replacing task-specific solutions. We critically examine the evolving role of human-AI collaboration, where human oversight remains essential for creative direction and mitigating AI hallucinations. Finally, we identify emerging challenges including copyright concerns, bias mitigation, computational demands, and the need for robust regulatory frameworks. This review provides researchers and practitioners with a comprehensive understanding of current AI capabilities, limitations, and future trajectories in creative applications.}}




\maketitle

\section{Introduction}

The influence of artificial intelligence (AI) has grown dramatically over the past few years, particularly due to the rise of generative AI and large language models (LLMs). These advancements are widely regarded as beneficial by many countries, creating significant opportunities for growth (e.g. as outlined in the UK, by the Authority of the House of Lords \cite{UK:Large:2024}). These advances have also had significant direct and indirect impacts on the creative industries, influencing the direction of their growth. Generative AI, for instance, primarily focuses on generating new data that is not identical to the training data yet shares similarities with it. However, the cardinality of the training data can be huge,  larger than what any individual human has ever encountered. The resulting output may therefore act as a new source of inspiration. 

AI tools also provide opportunities for a wider range of users to work more efficiently and  effectively, with even greater creativity. Moreover, these new technologies not only influence creators,  but they also enable new ways for audiences to experience art and culture \cite{Jeary2024}.

A major breakthrough in generative AI has been led by OpenAI\footnote{\url{https://openai.com/}}, an AI research and deployment company, with their introduction of Generative Pre-trained Transformer (GPT) models for LLMs. LLMs are specifically designed to understand and generate human language. They are characterized by their vast size in terms of parameters and the amount of training data used to create them. This breakthrough was particularly impactful when the company released ChatGPT %\footnote{\url{https://chat.openai.com/}}
in 2022, which was fine-tuned from a model in the GPT-3.5 series. ChatGPT is a conversational model that includes advanced safety features that mitigate the generation of inappropriate content. Several other LLM platforms were also developed contemporaneously, such as LaMDA and PaLM by Google AI, Ernie Bot by Baidu, and BLOOM by BigScience. Additionally, Anthropic launched Claude, the LLM trained specifically to be harmless and honest, leveraging reinforcement learning from human feedback (RLHF) - a technique used to train AI systems to appear more human \cite{Bai:Train:2021}. Nonetheless, ChatGPT stands out as the most renowned, thanks to its quick and efficient responses, and notably its public accessibility, being available for free. 

Another breakthrough in 2022 was in the area of text-to-image models. OpenAI achieved a significant milestone with DALL·E 2, producing impressive artworks and photorealistic images despite its limited language understanding. Midjourney by Midjourney, Inc., another well-known text-to-image generator, supports {  image resolutions of 1024$\times$1024 pixels and also provides upscaling tools.} Stable Diffusion by Stability AI, for which the code and model weights are publicly available\footnote{\url{https://github.com/Stability-AI/stablediffusion}}, allows developers and artists to further adapt AI to suit their own specific applications.

The next breakthrough occurred in 2023 when OpenAI unveiled GPT-4, a significantly larger model with {  third-party estimates of 1.8 trillion parameters; OpenAI has not disclosed an official figure.} It also demonstrated improved performance compared to its predecessors \cite{openai:gpt4:2023}. However, this still {  many orders of magnitude smaller than the human brain’s estimated 100–1,000} trillion synaptic connections\footnote{\url{https://www.rsb.org.uk/biologist-features/ai-versus-the-brain}}. GPT-4 is a multimodal large language model that can generate responses to both text and images. {  In the ChatGPT interface, GPT-4 is integrated with} DALL·E 3, enabling it to comprehend a much broader range of nuances and details than earlier versions. In March 2024, Claude 3 Opus by Anthropic was released, boasting multimodal capabilities in generating images, tables, graphs, and diagrams. Moreover, Anthropic claims that Claude 3 Opus outperforms GPT-4 in generating human-like dialog and contextually aware responses. These rapid advances have, in turn,  led the creative industries to face significant challenges. For example, DMG Media, the Financial Times, and Guardian Media Group have highlighted concerns about the potential impact on print journalism, particularly if AI tools reduce the need for users to click through to news websites, affecting advertising and subscription revenues \cite{UK:Large:2024}. There is also concern about `AI-generated slop'—low-quality, mass-produced content created by AI that often lacks coherence or originality\footnote{\url{https://reutersinstitute.politics.ox.ac.uk/news/ai-generated-slop-quietly-conquering-internet-it-threat-journalism-or-problem-will-fix-itself}}. It is typically used for spam, search engines, or clickbait, and is criticized for cluttering the internet and undermining genuine human-created content.

The generation of videos is significantly more challenging for AI than generating images. In February 2024, Google announced Gemini 1.5, which had the capability to process approximately 8 times more data than GPT-4{  ; this comparison refers to a 1M-token context window versus 128k tokens.} This opening opportunities for  video  and audio processing\footnote{\url{https://blog.google/technology/ai/google-gemini-next-generation-model-february-2024/}}. In the same month, OpenAI provided its first preview of Sora,  %\footnote{\url{https://openai.com/sora}}, 
a model capable of generating impressive realistic videos up to 1 minute long. Based on the videos released by OpenAI, Sora appears to outperform other text-to-video models. Sora is currently available to ChatGPT subscribers{  , but new Plus accounts may face a wait-list or monthly credit limits.} A month later, Gemini 1.5 announced its support for native audio understanding in 180$+$ countries.  With the emergence of these tools, together with the prospect of further advances, it is clear that video content creation will be a major beneficiary. This will further open up the media landscape for creativity and provide more opportunities for diverse storytellers, while also reducing production time. A recent example is the AI-generated Christmas commercial by Coca-Cola\footnote{\url{https://www.youtube.com/watch?v=4RSTupbfGog}}{  , which elicited mixed responses in the press \cite{Yang2025}, thereby pre-empting questions regarding potential hype}. Such advertisements overcome the limitations of current technologies by using very short videos with rapid scene transitions, ensuring that any artifacts, such as unnatural fingers, are less apparent.

For the case of post-production workflows, generative AI may not have a direct impact, but the neural networks originally proposed for generative AI have been widely adapted to serve this purpose. This has led to significant improvements in both output quality and computational speed. Moreover, there is  a noticeable trend  towards adopting a unified framework rather than addressing individual tasks, as it better reflects real-world scenarios. For instance, natural history filmmaking involves challenging acquisition environments and high production standards. Filming often takes place in low light conditions, in the presence of heat haze, underwater or in adverse weather conditions. This often results in increased noise levels, focus issues, low contrast, color balance problems, and blurriness in the footage. In such cases, unified models can offer advantages in generalizing to diverse tasks and providing flexibility. Take Painter by BAAI Vision \cite{Wang:Painter:2023} as an example, which employs an image pair as a task prompt (similar to a text prompt in LLMs), their model transfers the input image to produce a similar output as the task prompt, enabling it to undertake various tasks such as segmentation, low-light enhancement or rain removal.

While generative AI can facilitate and accelerate the creation and post-processing of digital media, there is an equivalent need to transmit or stream it efficiently to users. Although AI-based solutions have been proposed both for the enhancement of conventional video coding tools and for new compression frameworks, they are yet to be  adopted in practical applications due to hardware constraints, complexity issues and a lack of standardization. Despite this,  the latest learning-based video codecs have already demonstrated their potential to compete with conventional standardized video codecs and are being actively investigated in various standards bodies such as MPEG and AOM.

Furthermore, in recent years,  AI has also impacted our ability to assess and monitor the perceptual quality of visual media. Advances have included new model architectures based on different attention mechanisms and the application of LLMs, which evidently improve model generalization. New training methodologies have also been proposed based on weakly/unsupervised learning, which address issues associated with the limited availability of labeled training content.

One of the exciting aspects of using LLMs in the creative sector is that `The human in the loop' \cite{chung:human-loop:2021} is simplified through text prompts, with sophisticated, multilingual language capabilities enabling artists to convey complex emotions and narratives. This is important because generative AI does produce mistakes, known as hallucinations. Human oversight is thus essential to correct this through reinforcement learning with feedback \cite{Wu:brief:2023}.

In this paper, the objective is to present the major technological advancements that have emerged since our previous review on AI in the creative industries (2022) \cite{Anantrasirichai:AI:2022}. Whereas the earlier paper was written at a time when AI tools primarily served supportive roles, this updated review captures the disruptive shifts driven by generative AI and related technologies over the past few years. { Adopting an integrative narrative rather than a systematic approach, we aim to synthesize major developments and identify cross-domain themes, rather than conduct an exhaustive quantitative survey. Sources published between 2022 and mid-2025 were selected from leading AI conferences (CVPR, ICCV, ECCV, NeurIPS, ICLR, IEEE TPAMI, IEEE TIP), verified arXiv preprints highlighting emerging developments, and industry releases, based on their technical significance and impact (e.g., OpenAI, Google DeepMind, Meta, Adobe). Selection was guided by recency, originality, and relevance to creative-industry applications, ensuring representative coverage of key models and trends.} {\color{blue}In summary, we initially screened more than 1,000 documents published between 2022 and mid-2025, including major AI conference papers ($\approx$60\%), journal articles ($\approx$30\%), and industry or policy reports ($\approx$10\%). From these, around 450 sources were retained for detailed analysis across creation, enhancement, immersive media, and AI-driven production workflows.}
The review first outlines recent advances in AI technologies (Section \ref{sec:overview}), then highlights their applications in creative domains (Section \ref{sec:existing}), and finally discusses emerging challenges and future directions (Section \ref{sec:discussion}).

% ====================================================
\section{Current Advanced AI Technologies}
\label{sec:overview}

This paper provides a review of AI in the creative industries,  building on our previous publication in 2022 \cite{Anantrasirichai:AI:2022}. The reader is referred to that work for an introduction to AI, basic neurons, convolutional neural networks (CNNs), generative adversarial networks (GANs), recurrent neural networks (RNNs) and deep reinforcement learning (DRL). In this paper,  we emphasize four key technologies that have grown in importance since 2022 that have had a significant impact on the creative industries. These are Transformers, Large language models (LLMs), Diffusion Models (DMs), and Implicit Neural Representations (INRs). It is important to note that, while these newer technologies are gaining prominence, those from previous generations remain in widespread use, often in conjunction with the newer ones. For instance, CNNs complement transformers since CNNs effectively capture local features and semantic meaning, while the attention mechanism in transformers captures global dependencies.

One important class within AI that has become dominant since our previous review comprises Foundation models (FMs). These were described by The Stanford Institute for Human-Centered Artificial Intelligence in 2021 \cite{Bommasani2021FoundationModels} as ``any model that is trained on broad data (generally using self-supervision at scale) that can be adapted (e.g., fine-tuned) to a wide range of downstream tasks". Foundation models have been enabled by rapid advances in AI-oriented computing power and have underpinned the emergence and success of Large Language Models, particularly following the launch of ChatGPT by OpenAI in  2022. ChatGPT has become the fastest-growing consumer software application in history\footnote{\url{https://www.reuters.com/technology/chatgpt-sets-record-fastest-growing-user-base-analyst-note-2023-02-01/}}.
These technologies are expanded on below.

\subsection{Transformers}
\label{ssec:transformers}

In 2017, Google AI introduced the concept of  `Transformer' architectures in their publication `Attention Is All You Need' \cite{Vaswani:attention:2017}. This work has since been instrumental in the development and success  of large language models alongside many other applications, including vision understanding \cite{Dosovitskiy:image:2021}, and multiple modality learning (e.g., Gato \cite{Reed:Generalist:2022}). 

Before the advent of transformers, natural language processing (NLP) was performed using recurrent neural networks (RNNs), processing data sequences sequentially. In contrast, the ability of transformers to capture long-range dependencies through self-attention mechanisms that extend across all words in the sequence meant that the importance of different words could be established globally,  understanding relationships regardless of their positions. This context-aware representation enables parallel processing of the entire sequence, making the transformers computationally efficient. A set of several attention layers running in parallel is called Multi-Head Attention. 

\begin{figure}
    \centering
    \includegraphics[width=\textwidth]{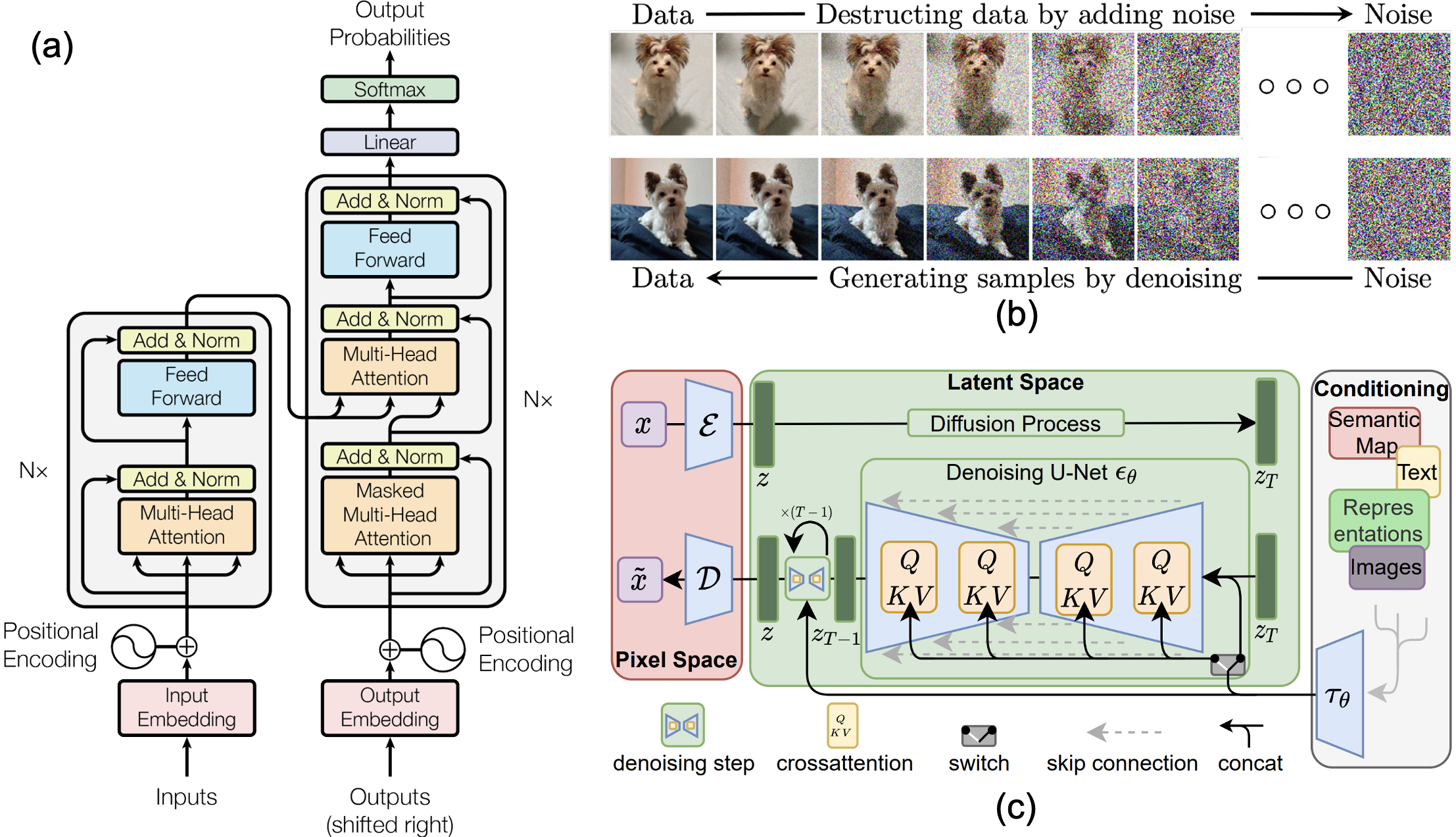}
    \caption{Generative AI. (a) Transformer architecture \cite{Vaswani:attention:2017}. (b) The top row represents the diffusion process and the bottom row represents the generation process of the new image \cite{Yang:diffusion:2023}. (c) Latent Diffusion Models (LDM) \cite{Rombach:LDM:2022}. }
    \label{fig:generativemodel}
\end{figure}

The Transformer architecture, shown in Fig. \ref{fig:generativemodel} (a), comprises Encoder and Decoder sections, similar to many CNN-based generators. However, the encoder is now a stack of identical layers, concatenating a multi-head self-attention mechanism and a fully connected feed-forward network. The decoder is also a stack of identical layers, in which each layer has an additional sub-layer to perform multi-head attention over the output of the encoder stack.

Mathematically, the attention function is computed from inputs: query $Q$, keys $K$, and values $V$. The matrix of outputs of  attention function is
\begin{equation}
    \text{Attention}(Q, K, V ) = \text{softmax}(\frac{QK^T}{\sqrt{d_k}}) V,
    \label{eqn:attention}
\end{equation}
\noindent where $d_k$ is a  dimension of $K$. The term ${QK^T}$ is Dot-Product Attention, which yields a high similarity value when the two words are closely related.  If $Q$ and $K$ are from the same sentence, Eq.~\ref{eqn:attention} refers to self-attention, but if $Q$ and $K$ are from different sentences, it is referred to as cross-attention. Within the network, multi-head attention is actually employed to concurrently process attention and enable the model to collectively focus on information from distinct representation subspaces at various positions through the learnable parameters $W$s.
\begin{equation}
\begin{split}
        \text{MultiHead}(Q, K, V ) &= \text{Concat}(\text{head}_1, ..., \text{head}_h) W^O,\\
        \text{head}_i &= \text{Attention}(QW_i^Q, KW_i^K, VW_i^V ). 
\end{split}
\label{eqn:MultiHead}
\end{equation}
It should be noted that attention modules are not solely used in transformers, but have also been successfully integrated into other deep learning architectures such as CNNs, used for image classification \cite{Li:HAM:2022}, object detection \cite{Woo:CBAM:2018}, and other computer vision tasks \cite{Guo:Attention:2022}.

In 2020, the first successful training of a transformer encoder for image recognition was published \cite{Dosovitskiy:image:2021}, referred to as a Vision Transformer (ViT). The ViT decomposes an input image into patches, similar to words in a sentence, and processes them through multi-head attention. Additionally, a Multilayer Perceptron (MLP) is employed as the feedforward network.  In later work Microsoft introduced  a hierarchical division of image inputs and a shifted window approach in their Swin Transformer \cite{Liu:Swin:2021}. This was reported to outperform ViT by 2.4\% in ImageNet-22K classification (21,841 different categories). Its version 2 \cite{Liu:Swinv2:2022} applied a cosine function in the attention module. enabling the scaling of capacity and resolution. More detail on transformer-based object detection is discussed in Section \ref{sssec:recog}. To date, Swin Transformers have been widely adopted in a range of applications including image restoration \cite{Fan:SUNet:2022}. 
Comprehensive surveys on the use of transformers for image and video processing can be found in \cite{Khan:Transformers:2022} and \cite{Selva:video:2023}, respectively.
 
Transformers have been widely used and offer better performance across many tasks. One reason for this widespread adoption has been the availability of open-source Transformer libraries such as Hugging   Face\footnote{\url{https://huggingface.co/}}, a platform that assists developers to build applications for tasks including computer vision, NLP, audio, tabular data, multimodal tasks, and reinforcement learning. The platform also provides access to model zoo\footnote{Such as \url{https://modelzoo.co/} and \url{https://pytorch.org/serve/model_zoo.html}} pretrained networks and datasets. 

In recent years, state space models \cite{gu2023mamba, zhu2024vision}, commonly known as `Mamba' have emerged. These are  a linear variant of Transformers distinguished by their linear complexity in attention modeling. They are acknowledged to offer an equivalent or better performance than traditional Transformers, while demanding fewer computational resources and less memory.

\subsection{Large language models}
\label{ssec:llms}

{ LLMs are based on transformer models using self-attention mechanisms as their core modules. Training comprises two steps: i) pre-training with large amounts of unlabelled text data in an unsupervised manner to learn word meanings and relationships, and ii) task adaptation through either fine-tuning or prompt-tuning. This pre-training approach is why OpenAI refers to their model as a Generative Pre-trained Transformer (GPT). 

Fine-tuning involves training the model on new datasets, which must be large enough to ensure generalization to new tasks. Prompt-tuning and prompt engineering are emerging disciplines focused on developing and optimizing prompts for efficient model use. Prompts guide how AI models interpret and respond to user queries. Prompt engineering structures text or phrasing to steer the model toward desired outputs, relying heavily on experimentation and understanding model behavior. Prompt-tuning instead trains a small set of parameters before using the LLM, requiring relatively little new data. This approach converts text inputs into task-specific virtual tokens while keeping the pre-trained model unchanged \cite{Lester:power:2021}. Its main drawback is reduced interpretability, though the paradigm has extended to other domains such as visual prompt tuning \cite{Jia:VPT:2022}. For a comprehensive survey of LLMs, see \cite{zhao:survey:2023}.}

\begin{figure}
    \centering
    \includegraphics[width=\textwidth]{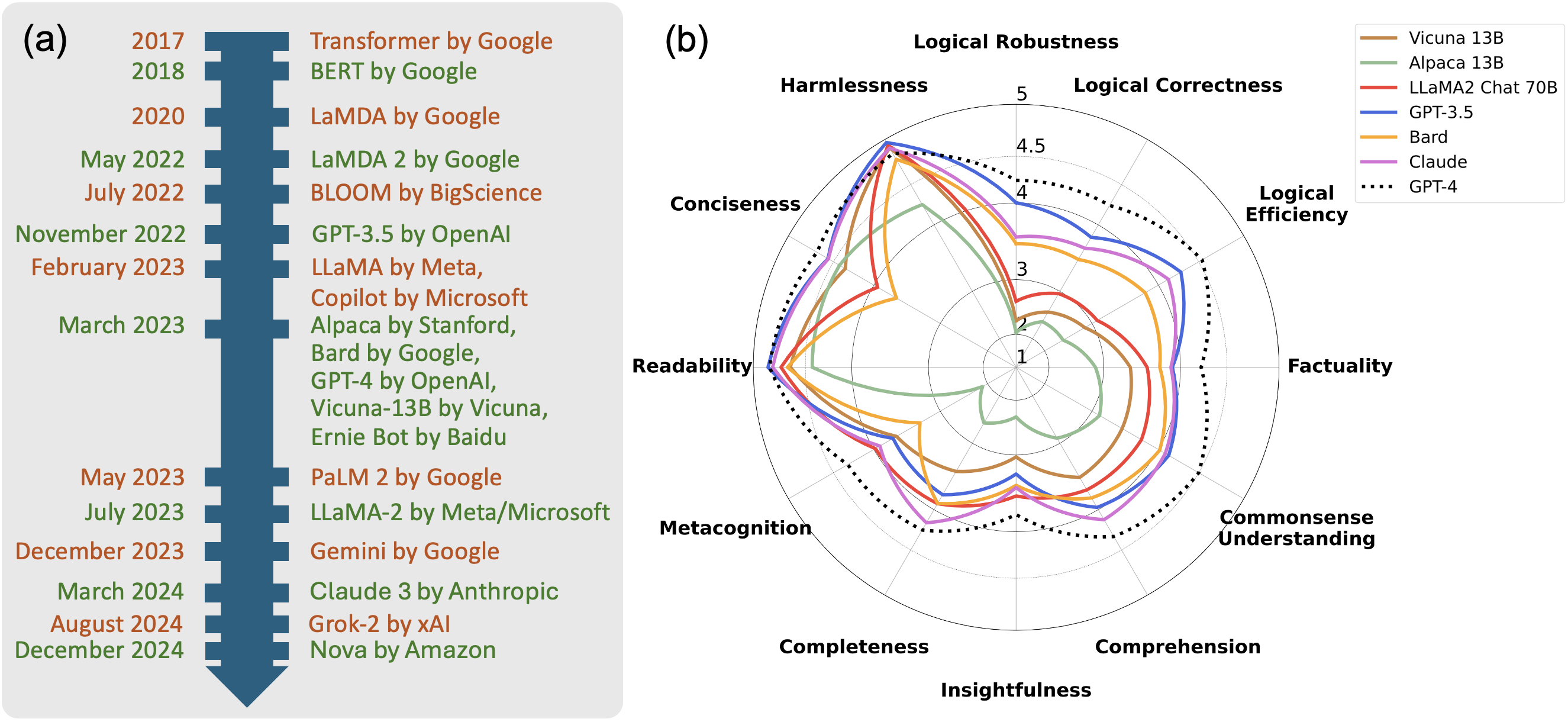}
    \caption{(a) Timeline of large language models. (b) Performance comparison evaluated by FLASK \cite{Ye:FLASK:2024}.}
    \label{fig:FLASK_LLM_and_history}
\end{figure}

To date, there are many LLM platforms as shown in Fig. \ref{fig:FLASK_LLM_and_history}. Fig. \ref{fig:FLASK_LLM_and_history}(a) shows their timeline. Many surveys and evaluations of LLMs are also available \cite{zhao:survey:2023,Chang:Survey:2024,YAO:Survey:2024}. These include FLASK (Fine-grained Language Model Evaluation based on Alignment Skill Sets) \cite{Ye:FLASK:2024} which evaluates LLMs based on 12 fine-grained skills for comprehensive language model evaluation: logical correctness, logical robustness, logical efficiency, factuality, commonsense understanding, comprehension, insightfulness, completeness, metacognition, conciseness, readability, and harmlessness. Evaluation results from FLASK are shown in Fig. \ref{fig:FLASK_LLM_and_history} (b).

{ A clear divide exists between open-source and proprietary LLMs. Open-source models such as LLaMA 3, Mistral, and Falcon promote transparency, reproducibility, and creative customization, allowing researchers to fine-tune or deploy them locally with improved data control. Proprietary systems like GPT-4, Gemini 1.5, and Claude 3, meanwhile, deliver stronger multimodal integration, safety alignment, and reliability—benefiting from access to large, private datasets and compute resources. Yet their closed architectures limit interpretability and independent evaluation. Together, these ecosystems form a complementary landscape balancing openness and innovation against performance and safety, shaping accessibility and creative autonomy in AI-driven practice.}

% --------------------------------------------------------------
%\subsection{Zero/Few-shot learning}
%\label{ssec:fewshot}

% --------------------------------------------------------------
\subsection{Diffusion Models}
\label{ssec:DMs}

A generative model, in the context of AI, exploits machine learning to learn a probability distribution of the training data to generate new data samples. The very first models were based on Autoencoders (AEs) that learn to encode input data into a lower-dimensional representation (latent space) and then decode it back to its original form. A specific type of AE, a variational autoencoders (VAE) \cite{Kingma:auto:2014}, learns the latent space as statistical parameters of probabilistic distributions, leading to significant improvement of the generated results. Concurrently, Goodfellow et al. \cite{Goodfellow:GAN:2014} introduced an alternative architecture known as a Generative Adversarial Network (GAN). GANs comprise two competing AI modules: a generator, which creates a sample, and a discriminator, which determines whether the received sample is real or generated. When comparing VAEs to GANs, VAEs exhibit greater stability during training, whereas GANs excel at producing realistic images. More details about AEs and GANs for creative technologies can be found in our previous review \cite{Anantrasirichai:AI:2022}.

An important factor in driving the rapid growth of generative AI has been  the development of diffusion probabilistic models (referred to as diffusion models (DMs)). The first DM was introduced in 2015 by Sohl-Dickstein et al. \cite{Dickstein:Deep:2015}, using Nonequilibrium Thermodynamics. However, it took a further 5 years for DMs to generate desirable results: the era of DMs began with Denoising Diffusion Probabilistic Models (DDPMs) proposed by Ho et al. \cite{Ho:DDPM:2020} in 2020 and Score-based diffusion models proposed by Song et al. \cite{Song:Score:2021} in 2021. These involve a simplified process using a denoising autoencoder to approximate Bayesian inference. In brief, the models leverage a diffusion process to learn a probability distribution of the input data. As the name suggests, the data is diffused by gradually adding noise at each iteration step as shown in Fig. \ref{fig:generativemodel} (b). A deep neural network (DNN) is then trained to remove this noise, called the denoising process or reverse process. Consequently, the trained model uses random noise to generate data with characteristics similar to those of the training samples. Comparing to GANs, DMs provide higher diversity samples \cite{Dhariwal:Diffusion:2021} and a training process that is much more stable and does not suffer from mode collapse. DMs are however computationally intensive and require longer training times compared to GANs. The complexity can be significantly reduced by training the DMs in latent space. Latent Diffusion Models (LDM) \cite{Rombach:LDM:2022} use pretrained networks to convert images to feature maps, and perform training on a low-dimensional space. The diagram of LDM is shown in Fig. \ref{fig:generativemodel} (c). 

Generating a synthesized sample at random might not be particularly useful, especially for creative industry applications. Therefore, conditional diffusion models have been proposed, supporting a wide range of applications such as text-to-sound, text-to-images, and image-to-videos. For DMs, the conditional distributions are modeled using a conditional denoising autoencoder. Classifier guidance was introduced in \cite{Dhariwal:Diffusion:2021} to improve the generation of images of a desired class. For example, when we provide the model with information, such as `a flower', the DM will synthesize a variety of flower images, as the word `flower' guides the model toward the latent distribution that is formed by various images of flowers. The work in \cite{Choi:ILVR:2021} simply refines the latent space of well-trained unconditional DDPM so that the higher-level semantics of the synthetic samples are similar to the reference (conditioning).
The LDM \cite{Rombach:LDM:2022} offers more flexible conditional image generators by adding cross-attention layers (referred to Transformers in Section \ref{ssec:transformers}) to the denoising autoencoder. A survey on the methods and applications of DMs prior to 2024 can be found in \cite{Cao:survey:2024}.

\subsection{Implicit Neural Representations}

Implicit Neural Representations (INR), also called neural fields, neural implicits or coordinate-based neural networks, represent input content implicitly through learned functions $F$, as shown in Eq.~\ref{eqn:inr}. They can be considered as fields $x$ (represented by a scalar, vector, or a tensor with a value, such as a magnetic field in physics) that are fully or partially parameterized by a neural network $\Phi$, typically an MLP \cite{xie2022neural}.
\begin{equation}
F(x, \Phi, \nabla_x \Phi, \nabla_x^2 \Phi, \dots) = 0, \quad \Phi : x \mapsto \Phi(x).
\label{eqn:inr}
\end{equation}

 Although this concept appears complex,  the process is actually very straightforward. For example, in the case of an image, the coordinates of each pixel $(x, y)$ contain color information $(r, g, b)$. The INR inputs $(x, y)$ to the MLP and learns to provide the output $(r, g, b)$. The weights and biases of the MLP now represent such an image. Usually, the number of parameters of the MLP is smaller than the total number of pixels multiplied by 3, accounting for the 3 color channels. Hence, one of its emerging applications is in data compression \cite{kwan2024hinerv}. Moreover, the INR can handle complex and high-dimensional data efficiently, attracting attention for visual computing applications such as  3D scene reconstruction.

Traditional MLPs employ ReLU (rectified linear unit) for non-linear activation due to its simplicity. However, Sitzmann et al. \cite{sitzmann:siren:2020} demonstrated that periodic functions, such as sinusoids, are more suitable for representing complex natural signals, offering a better fit to the first- and second-order derivatives of the signals. However, this activation can cause ringing artifacts. Saragadam et al. instead proposed using complex Gabor wavelets \cite{Saragadam:wire:2023}, which learn to represent high frequencies better and simultaneously are robust to noise.

One of the fastest-growing areas that exploits INRs is \textbf{Neural Radiance Fields (NeRF)}, evidenced by 57 papers presented at CVPR, the largest annual conference in computer vision, in 2022 growing to 175 papers in 2023\footnote{\url{https://markboss.me/post/nerf_at_cvpr23/}}, before dropping to 71 in 2024, largely due to competition from 3D Gaussian Splatting\footnote{\url{https://github.com/Yubel426/NeRF-3DGS-at-CVPR-2024}}. First introduced in 2020 by Mildenhall et al. \cite{Mildenhall:NeRF:2020}, NeRF is a form of neural rendering, a subset of generative AI, that generates novel views of a scene based on a partial set of 2D images. It achieves this by learning a mapping from 3D spatial coordinates and view directions $(x,y,z,\theta,\phi)$ to colors and density $(r,g,b,\sigma)$. This implicit representation allows NeRF to handle complex scenes with varying geometry and appearance,  resulting in highly realistic renderings that include accurate lighting, shadows, and reflections. More detail can be found in Section \ref{sssec:nerf}.

% ====================================================

%\begin{landscape}
\begin{table}
\caption{Creative applications and corresponding AI-based methods mentioned in this paper}
%\tiny
 %\hskip-5.0cm
 \begin{tabular}{>{\raggedright\arraybackslash}p{2cm}>{\raggedright\arraybackslash}p{2.5cm}|>{\raggedright\arraybackslash}p{3.2cm}>{\raggedright\arraybackslash}p{3.2cm}>{\raggedright\arraybackslash}p{3.2cm}}
% \resizebox{\linewidth}{!}{
% \begin{tabular}{lr|lll}
 \\
 \toprule
\multicolumn{2}{c}{\multirow{2}{*}{Application}} & \multicolumn{3}{|c}{Technology} \\ \cmidrule{3-5}
& & Trans./Attn.$^1$ & Diffusion model$^2$  & INR\\
\midrule
{\bf Creation} & Text & \cite{Vaswani:attention:2017,Wang:SIMVLM:2022, openai:gpt4:2023, wei2024vary}  & \\
& audio/music & \cite{Alayrac:Flamingo:2022, Huang:GenerSpeech:2022} & \cite{Li:Diffusion:2022, Yang:Diffsound:2023, Evans:stable:2025} & \\
& Image &  \cite{Alayrac:Flamingo:2022,esser:scaling:2024} & \cite{Rombach:LDM:2022, Brooks:InstructPix2Pix:2023, gal:Image:2023, Gandikota:Unified:2024, Lian:LLMG:2024,esser:scaling:2024,ren:hypersd:2024, Feng_Ma_2025, Liu_Ma_2025} & \\
% ---------------------------------------------------------------
& Animation/video & \cite{hong:cogvideo:2023,villegas:phenaki:2023,Azadi:Make:2023,Yu:Bidirectionally:2023,Liu:FETV:2023,wang:disco:2024, xu:VASA-1:2024,corona:vlogger:2024, Gupta:Photorealistic:2024, Hu_2024_CVPR, Zhu:INFP:2024} & \cite{singer:Make:2023, molad:dreamix:2023, wang:modelscope:2023, wu:tune:2023,Liu:FETV:2023, Gupta:Photorealistic:2024, Zhu:INFP:2024, wang2025lavie, wu2025customcrafter} &\\
% ---------------------------------------------------------------
 & 3D/AR/VR & \cite{yang:Holodeck:2024} & \cite{Xu:NeuralLift:2023, Melas:RealFusion:2023, Qian:Magic123:2024,tang:dreamgaussian:2024} & \cite{tang:dreamgaussian:2024, ren:dreamgaussian4d:2023, zhao2024clear} \\
% ---------------------------------------------------------------
\midrule
{\bf Information Analysis} & Text categorization & \cite{sun:text:2023, shi:chatgraph:2023, Hou:promptboosting:2023, AI2025125952} \\
% ---------------------------------------------------------------
 & Film analysis & \cite{Mao:Biases:2023,krugmann:sentiment:2024,Hartmann:More:2023}  \\
 % ---------------------------------------------------------------
 & Content retrieval & \cite{Metzler:Rethinking:2021, Yan:Universal:2023, Lu:content:2023, Rajput:recommender:2023, li:unigen:2024, li2024learning} & \cite{Jin:DiffusionRet:2023} \\
 % ---------------------------------------------------------------
 & Intelligent assistants & \cite{King:Sasha:2024} \\
 % ---------------------------------------------------------------
 \midrule
{\bf Content} & Enhancement  & \cite{Xu:SNR:2022, liang:RVRT:2022, Wang:Ultra:2023, Lin:SPATIO:2024, Youk:FMA:2024, Liang:VRT:2024} & \cite{HOU:Global:2023, Yi:Diff:2023, Jiang:Low:2023, lin2024lowlight} & \cite{Yang:Implicit:2023} \\
{\bf  Enhancement  } & Style transfer & \cite{Deng:StyTr2:2022, Moon:generalizable:2023, Chung_2024_CVPR} & \cite{Zhang:Inversion:2023,Chai:StableVideo:2023} & \cite{Moon:generalizable:2023,Kim:Controllable:2024} \\
% ---------------------------------------------------------------
 {\bf  and Post} & Super-resolution & \cite{Liang:SwinIR:2021, Lu:Transformer:2022, Liu:Learning:2022, Chen:Activating:2023,li:GRL:2023, kang:gigagan:2023, Liang:VRT:2024, xu:videogigagan:2024,wang2025seedvr} & \cite{Saharia:image:2023, Moliner:solving:2023, Gao:Implicit:2023, cao2025zero, wang2025seedvr} & \cite{Chen:Learning:2021, Saharia:image:2023, Fei:Generative:2023, Gao:Implicit:2023, Yin:CLE:2023}\\
% ---------------------------------------------------------------
{\bf Production} & {Restoration} & \cite{Wang:Uformer:2022,Zamir:Restormer:2022, li:GRL:2023,yang:ldp:2023, Liang:VRT:2024, Morris:DaBiT:2024, Liang:SwinIR:2021,Fan:SUNet:2022, Yu:DBT:2022, Wang:Painter:2023, Song:vision:2023,Xu:Video:2023, mao:single:2022,Zhang:Image:2024,zou2024deturb, fang2025guided, Yue:RViDeformer:2025, Jin:Masked:2025,  Yue:RViDeformer:2025, Shi:VmambaIR:2025} & \cite{Jiang:Low:2023,Fei:Generative:2023, yang:realworld:2023, Nair:AT-DDPM:2023,Jaiswal:Physics:2023, cao2025zero, feng2025residual} & \cite{Jiang:NeRT:2023} \\
  % ---------------------------------------------------------------
 & Inpainting & \cite{Li:MAT:2022,Liu:Reduce:2022,Ren:DLFormer:2022,Zhou:ProPainter:2023,HUANG:Sparse:2024} & \cite{Moliner:solving:2023, Fei:Generative:2023} & \\
 % ---------------------------------------------------------------
& Fusion & \cite{Ma:SwinFusion:2022, Rao:TGFuse:2023,Liu:Multi:2023, LI2024102147} & \cite{Zhao:DDFM:2023} & \\
 & Editing/VFX & \cite{Shi:Motion-I2V:2024} & \cite{Shi:Motion-I2V:2024,guo2024liveportrait} \\
 \midrule
 % ---------------------------------------------------------------
{\bf Information} & Segmentation & \cite{Bowen:Marked:2022, Kirillov:SAM:2023, Ke:SAM-HQ:2023,Wang:Painter:2023,Wang:SegGPT:2023,Zou:Segment:2023, Oquab:DINOv2:2024, ravi2024sam2,Zhang:DVISp:2025} & \cite{Wu:DiffuMask:2023, Xu:Open:2023, Gu:Diffusioninst:2024} & \cite{Gong:Continuous:2023, Cen:Segment:2023} \\
 % ---------------------------------------------------------------
{\bf  Extraction} & Recognition &  \cite{Carion:DERT:2020, Dosovitskiy:image:2021,Zhu:Deformable:2021, Liu:Swin:2021,  Neimark:video:2021, Liu:Swinv2:2022, Huang:MonoDTR:2022, Oquab:DINOv2:2024, zhao:videoprism:2024, im2025gate3d, tian2025yolov12} & \cite{Li:Your:2023,Chen:DiffusionDet:2023, Zhang:DiffAD:2025, WU2025102965}\\
 % ---------------------------------------------------------------
{\bf  and}  & Tracking &  \cite{Meinhardt:TrackFormer:2022,zeng:motr:2022,cui:mixformer:2022,Mayer:Transforming:2022, yang:track:2023,Chen:SeqTrack:2023,Zhang:MOTRv2:2023, Yi:Comprehensive:2024, kang2025exploring} & \cite{Luo:DiffusionTrack:2024, Xie:DiffusionTrack:2024, Zhang:DiffusionTracker:2024} & \cite{Jung:AnyFlow:2023}\\
 % ---------------------------------------------------------------
{\bf  Understanding}  & 3D Reconstruction  & \cite{Wang:multi:2021, Zhang:Lite:2023, Chen:Vision:2023,Yang:depthanything:2024,Oquab:DINOv2:2024,Yang:depthanythingv2:2024, LIU:DSEM:2025} & \cite{Barron:Mip-NeRF360:2022, Ji:DDP:2023, wynn:diffusionerf:2023, Ke:Repurposing:2024} & \cite{Mildenhall:NeRF:2020,pumarola:DNeRF:2020, mueller:instant:2022,Barron:Mip-NeRF360:2022,Mildenhall:NeRFDark:2022,Fang:Fast:2022, Guo:neural:2022, 9879447, Liu_2024_CVPR, azzarelli:waveplanes:2023, zhan2024kfd, Tang_2024_CVPR, LIU:DSEM:2025},\cite{ Fridovich:kplanes:2023,  kerbl:3Dgaussians:2023, wu:4dgaussians:2024, Yu:CoGS:2024, Huang:SCGS:2024, Wang2025, junkawitsch2025eva, kong2025efficient}$^\dag$  \\
\midrule

% ---------------------------------------------------------------
{\bf Compression} & Image$^\ast$ & \cite{zhu2022transformer,zou2022devil,liu2023learned}& \cite{careil2023towards,yang2024lossy,hoogeboom2023high,ghouse2023residual} & \cite{sitzmann2020implicit,dupont2021coin,dupontcoin++,strumpler2022implicit}\\
& Video& \cite{xiang2022mimt,mentzer2022vct}& \cite{li2024extreme}&\cite{chen2021nerv,bai2023ps,kwan2024hinerv,kim2024c3,leguay2024cool,kwan2024nvrc,gao2024pnvc,ruan2024point,kwan2024immersive}\\
&Audio$^\ast$ & & &\\
\midrule
% ---------------------------------------------------------------
{\bf Quality}  &  Image$^\ast$ & \cite{cheon2021perceptual,golestaneh2022no, shi2024transformer}& &\\
\textbf{Assessment}& Video$^\ast$ & \cite{wu2022fast,feng2024rankdvqa,wu2023exploringvideo,he2024cover,peng2024rmt}& & \\
\bottomrule
\multicolumn{5}{l}{$^1$ Trans./Attn. include transformers, mamba and CNN-based architectures that use attention module.} \\
\multicolumn{5}{l}{$^2$ Some diffusion models employ the transformer in their denoising autoencoders.}
 \end{tabular}
 \footnotesize
 $^\dag$ 
These methods are based on explicit neural representations. \\
$^*$ It is noted that for some compression and quality assessment tasks, there are other dominant network architectures in existing works. For example, LLMs have been used for image and audio compression, and visual quality assessment. Many neural audio codecs are also based on VQ-VAE models.
 
\label{tab:gather}
\end{table}
 %\end{landscape}
% ====================================================
\section{Advanced AI for the creative industries}
\label{sec:existing}

Similarly to our previous (2021) review of AI for the creative industries~\cite{Anantrasirichai:AI:2022}, Table \ref{tab:gather} categorizes applications and corresponding AI-based solutions. These areas are explored in more detail below.

% --------------------------
\subsection{Content creation}

Content creation is a fundamental activity of artists and designers and the term `\textit{AI art}' refers to artforms created with the assistance of an AI algorithm or entirely by an AI system. This can refer to various digital forms including images, texts, audio, and videos. The roots of AI art can be traced back to the 20th century, exemplified by AARON, a computer program initiated in 1972 to autonomously produce paintings and drawings \cite{encyclopedia_ai_v1}. The practicality of AI art has been enhanced with advancements in deep learning, particularly GANs from 2014 and, more recently, transformers, DMs and INRs. 

% --------------------------
\subsubsection{Text generation, script and journalism}

In the era of LLMs, AI writing tools have been widely used to assist various writing tasks, including generation of written articles, blog posts, essays, and reports. These tools go beyond mere grammar and spelling checks; they boast advancements enabling them to analyze the style and tone of written material, adding images, videos and tables, offering suggestions to enhance clarity, coherence, and overall readability \cite{ippolito:creative:2022}. Moreover, AI tools extend their utility beyond content generation by automating tasks like keyword generation, meta tags, and descriptions, thereby increasing search rankings using search engine optimization (SEO). Additionally, they support the process of publishing across multiple online platforms. Transformers have been used to generate image captions by combining information from the images with a word prefix or questions \cite{Wang:SIMVLM:2022}.

AI script generators serve as beneficial aids for writers, filmmakers, and game developers, offering inspiration, idea generation, and assistance in crafting entire scripts \cite{Jeary2024, Azzarelli:Reviewing:2024}. Human-AI brainstorming is helpful and saves time \cite{guo:exploring:2024}. Presently, there are numerous software and websites providing both free and paid script generation services. However, many of these tools are still constrained when it comes to longform creative writing. Dramatron, developed by Google \cite{Mirowski:cowriting:2023}, introduces hierarchical language generation, enabling the creation of cohesive scripts and screenplays spanning long ranges. This includes elements such as titles, characters, story beats, location descriptions, and dialogue.

As discussed earlier, chatbots are now powered by LLMs, effectively simulating human conversation. These fundamental LLMs are specialized for specific tasks. For instance, journalist AI and blog AI writers\footnote{For example, see \url{https://tryjournalist.com/}} generate content with layouts suitable for print or online publication. Additionally,  AI tools exist that are designed to detect AI-generated content (e.g., for checking for copyright), AI-writing styles, content originality, and to ensure the naturalness and flow of articles. Undoubtedly, generative AI is reshaping the way artists and journalists operate. For an in-depth exploration of the impact and implications of these technological advancements on news organizations, refer to the survey conducted by Beckett et al. \cite{Beckett:Generating:2023}.

Generating text and scripts automatically can also be done through image and video inputs without text prompts (e.g., image captioning \cite{Stefanini:From:2023}) and with text prompts. These approaches are referred to as Vision Language Models (VLMs):  multimodal models that learn from images and text. The most common and prominent models often consist of an image encoder, an embedding projector to align image and text representations, often via a dense neural network, and a text decoder stacked in this order. The most well-known technique is Contrastive Language-Image Pre-training (CLIP) \cite{radford2021learning}. More recent work in \cite{wei2024vary} scales up the vision vocabulary by incorporating new image features into the existing CLIP model, resulting in improved content understanding. A comprehensive survey of VLMs for vision tasks can be found in \cite{Zhang:vision:2024}.

% --------------------------
\subsubsection{Audio and music generation}
\label{sssec:musicgen}

Similar to language models, AI-based music generation has rapidly advanced due to unsupervised learning on large datasets and the use of transformers (see Section \ref{ssec:llms}). Examples of such systems include MuseNet\footnote{\url{https://openai.com/research/musenet}}, Magenta Studio\footnote{\url{https://magenta.tensorflow.org/studio}}, and Musicfy\footnote{\url{https://musicfy.lol/}}. These tools assist in music composition by learning complex musical patterns, predicting the next word or music note in a sequence, and mixing specified instruments. Moreover, AI tools can convert one type of sound into another, such as from whistling to violin or from flute to saxophone\footnote{See an example by Ummet Ozcan at \url{https://www.youtube.com/watch?v=lI1LCfTx2lI}}. This capability is invaluable for artists who may not be proficient in playing all the instruments they wish to incorporate, saving both time and costs. In  2024, Suno has released a model capable of producing radio-quality music that can be created in 2 minutes\footnote{\url{https://www.suno.ai/blog/v3}}. Later, Udio\footnote{\url{https://www.udio.com/}} was launched. This offers a prompt to create lyrics and music with a maximum duration of 90 seconds, and also appears to have, at least some, awareness of copyright.

AI voice software changes vocalizations from one person to another, for example enabling users to train the model to convert other people's voices into their own, e.g., lalals\footnote{\url{https://lalals.com/}}, Kits\footnote{\url{https://www.kits.ai/}}, Media.io\footnote{\url{https://www.media.io/online-voice-changer.html}}, etc. Certain software, such as Voice.ai\footnote{\url{https://voice.ai/}}, even offers real-time voice changing capabilities. The technologies behind this use a transformer to learn voice features and patterns in mel-spectrogram form. For example, the framework proposed in \cite{Yang:Diffsound:2023} uses a DM-based method with a transformer backbone to turn text input into a mel-spectrogram using the vector quantized variational autoencoder (VQ-VAE) \cite{Oord:Neural:2017}. Next, this mel-spectrogram is transformed into a sound wave.  Unlike a regular spectrogram, the mel-spectrogram is based on the mel-frequency scale, which offers higher resolution for lower frequencies. Voice style transfer often uses zero-shot learning (a model is trained to recognize classes or categories that it has never encountered during training) \cite{Huang:GenerSpeech:2022} or few-shot learning (a model trained with only one or a few examples per class) \cite{Wang:One:2022}. Stable Audio Open \cite{Evans:stable:2025} introduces a text-conditioned generative model for non-speech audio, trained on Creative Commons licensed data, capable of producing state-of-the-art 44.1kHz stereo audio.

Another emerging AI technology application is in the field of spatial audio. In 2022, Apple Music revealed that, in just over a year, more than 80\% of its worldwide subscribers were enjoying the spatial audio experience, with monthly plays in spatial audio increasing by over 1,000\%\footnote{\url{https://www.apple.com/uk/newsroom/2023/01/apple-celebrates-a-groundbreaking-year-in-entertainment/}}. With head tracking, this technology significantly enhances the immersive experience. Masterchannel has launched SpatialAI\footnote{\url{https://platform.masterchannel.ai/spatial}}, claiming it to be the world's first spatial mastering AI. This processes audio files and returns an optimized track for streaming platforms, along with an individually optimized stereo version for traditional distribution. All these advancements leverage transformer-based technologies.

% -----------------------------------------------------
\subsubsection{Image generation}

\begin{figure}
    \centering
    \includegraphics[width=\textwidth]{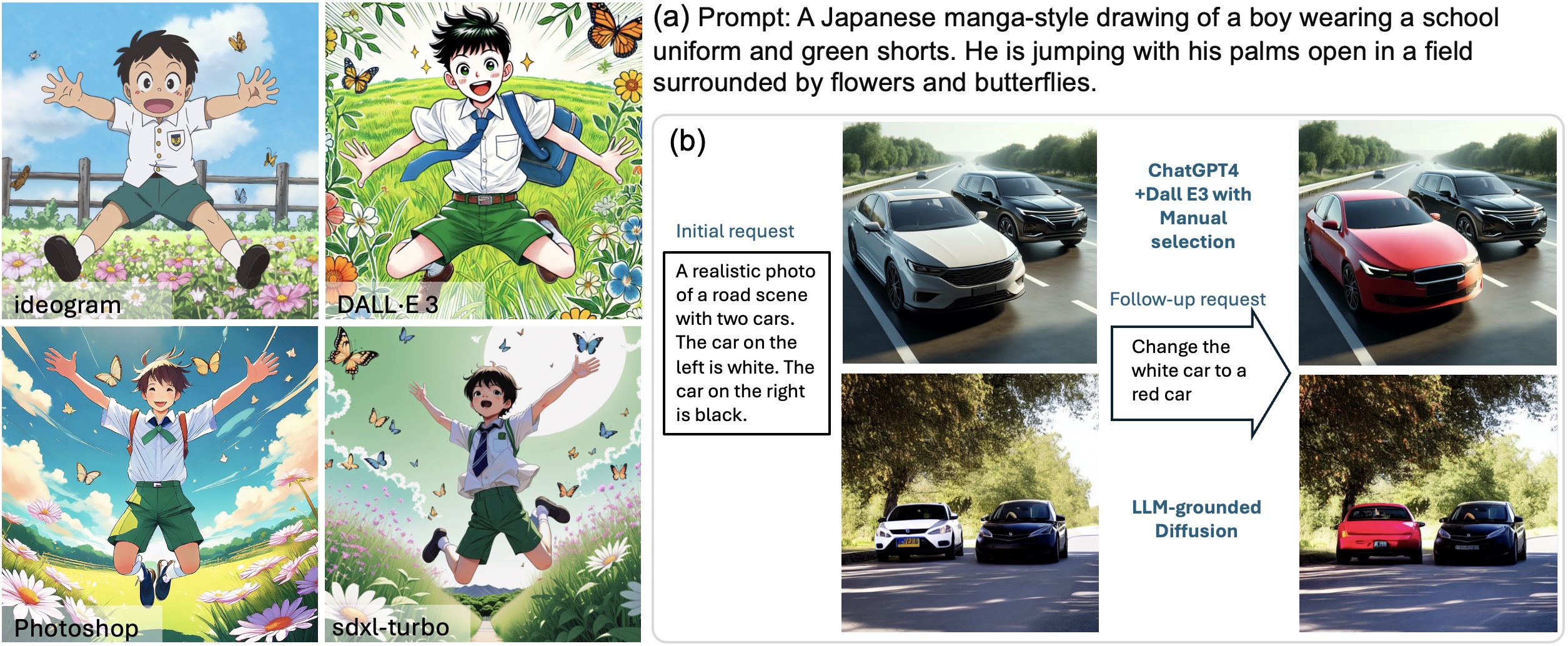}
    \caption{Text-to-image generation (generated on 27 November 2024). (a) text-to-image generation by Ideogram v1, DALL·E 3, Photoshop 2025, and sdxy-turbo by Nvidia. (b) The top-row images were generated by DALL·E in ChatGPT 4. The bottom-row images are generated by LLM-grounded Diffusion \cite{Lian:LLMG:2024}. }
    \label{fig:LLMGround}
\end{figure}

As described in Section \ref{ssec:DMs}, recent advances in AI technologies for image generation are based on Diffusion Models (DMs). Well-known and highly competitive text-to-image models include Stable Diffusion\footnote{\url{https://stability.ai/stable-image}}, Midjourney\footnote{\url{https://www.midjourney.com/home}}, DALL·E\footnote{\url{https://openai.com/dall-e-3}}, and Ideogram\footnote{\url{https://ideogram.ai/}}. Released in {  June} 2024, the latest version of Stable Diffusion (SD3), has been reported to outperform state-of-the-art text-to-image generation systems such as DALL·E 3 (released August 2023) \cite{esser:scaling:2024}, Midjourney v6 (released December 2023), and Ideogram v1 (released February 2024) in terms of typography and prompt adherence, based on human preference evaluations. These open-source tools are built on a Multimodal Diffusion Transformer (MM-DiT) architecture, which integrates attention from both text and images. LLM4GEN \cite{Liu_Ma_2025} fuses features from LLM and CLIP models to enhance the semantic understanding in text-to-image diffusion models, enabling them to better handle complex and dense prompts involving multiple objects. Examples of text-to-image generation are shown in Fig. \ref{fig:LLMGround} (a) comparing the performance of four models, i.e. Ideogram v1, DALL·E 3, Photoshop 2025, and sdxy-turbo by Nvidia. It is clear that hands are one of the most difficult features to generate, e.g., one hand has six fingers.

DALL·E 3, available on ChatGPT 4, also provides an inpainting tool, allowing the user to manually select the area to edit. However, as of April 2024, its performance is still limited. As illustrated in Fig. \ref{fig:LLMGround} (b), the selected area is the white car, and with the follow-up request to change the white car to the red car, DALL·E 3 generates correctly. However, if asked to replace it with a bicycle, it does not work. LLM-grounded Diffusion \cite{Lian:LLMG:2024} was the first to introduce a framework that allows multiple rounds of user requests without the need for manual selection on the image. This is achieved by generating layout-grounded images, first using stable diffusion and then masking the latent variables as priors for the next round of generation\footnote{Images in Fig. \ref{fig:LLMGround} (b) were generated using their demo: \url{https://huggingface.co/spaces/longlian/llm-grounded-diffusion}.}. Since then, text-driven image editing has seen significant improvements in quality, with most recent approaches adopting Diffusion Transformer architectures \cite{Feng_Ma_2025, Huang:Diff:2025}. 

Similar to DALL·E 3, Photoshop features a Generative Fill tool\footnote{\url{https://www.adobe.com/th_en/products/photoshop/generative-fill.html}} designed to generate new images or assist with photo editing. It accepts a text prompt and provides several generation choices. After defining the editing area, users can remove and add new objects (more inpainting tasks are discussed in Section \ref{ssec:inpaiting}), transfer to new styles, and expand content within images. Recently, Brooks et al. introduced InstructPix2Pix \cite{Brooks:InstructPix2Pix:2023},  a conditional diffusion model that generates image editing examples without predefined editing areas. By combining GPT-3 and Stable Diffusion, the model effectively captures and matches the semantic meaning of the content in both text and image. Sometimes, style and context are not easy to describe in words. Textual Inversion \cite{gal:Image:2023} personalizes large pre-trained text-to-image diffusion models based on specific objects and styles, using 3-5 images of a user-provided concept. ByteDance announced Hyper-SD \cite{ren:hypersd:2024} which proposed trajectory segmented consistency distillation and provides real-time high-resolution image generation from drawing with a control text prompt. 

%a visual language model combining both vision (image understanding) and language (text understanding) capabilities within a single model \cite{Alayrac:Flamingo:2022}.

% --------------------------------------------------
% previous paper presents only animation
\subsubsection{Video generation and animation} 
\label{sssec:videogen}

\begin{figure}
    \centering
    \includegraphics[width=\textwidth]{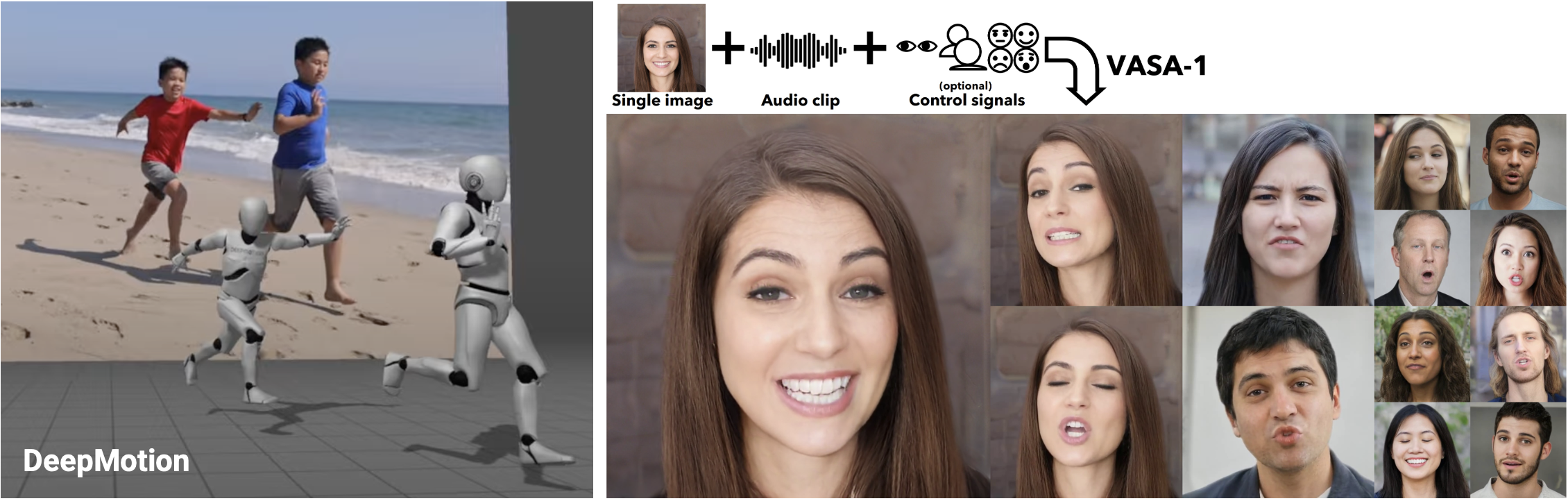}
    \caption{(Left) Video-to-3D post animation by DeepMotion. (Right) Image and audio to video by VASA-1 \cite{xu:VASA-1:2024}}
    \label{fig:Deepmotion_Vasa}
\end{figure}

{ Despite the success of text-to-image generation, text-to-video generation has not advanced at the same pace, growing more rapidly only in 2024 due to its computational expense and content complexity. Several major companies and private platforms have now released offerings, including Gemini 1.5 by Google, Make-A-Video by Meta, and Sora by OpenAI. Make-A-Video \cite{singer:Make:2023}, through a spatiotemporally factorized diffusion model, leverages joint text-image priors and super-resolution in space and time, though some results exhibit flickering artifacts\footnote{\url{https://makeavideo.studio/}}. Gen-2 by Runway\footnote{\url{https://research.runwayml.com/gen2}} supports both text- and image-to-video generation, producing smooth 4-second clips. In April 2024, Adobe Premiere Pro announced integration of generative AI tools for video extension with third-party models by OpenAI, Runway, and Pika Labs\footnote{\url{https://www.adobe.com/products/premiere/ai-video-editing.html}}, including contextual selection, inpainting for object removal, and object addition to videos via text prompts.

Text-to-video technologies, combined with AI voice, have been tested not only by artists and producers but also by a wider audience. Results from these experiments—such as automatically turning scripts into trailers and music videos—have been widely shared online\footnote{\url{https://twitter.com/minchoi/status/1775907105813217398}}. However, scene composition and transitions still require further editing to meet production needs\footnote{See an example by Curious Refuge at \url{https://www.youtube.com/watch?v=fJQbP34GoHQ}}.
In April 2024, Microsoft introduced VASA-1 \cite{xu:VASA-1:2024}, which converts a single image and speech audio clip into a realistic video of talking faces mimicking expressions and head movements (Fig. \ref{fig:Deepmotion_Vasa}, right). The resulting quality surpasses Google’s VLOGGER \cite{corona:vlogger:2024}, which uses a similar diffusion-based approach but additionally generates upper-body and hand motion. Recently, ByteDance proposed an audio-driven interactive head-generation model \cite{Zhu:INFP:2024} offering listening and speaking states during multi-turn conversations, based on a conditional diffusion transformer.

The main technologies underpinning text-to-video and image-to-video tasks are based on diffusion models (DMs) combined with 3D convolutions—or separate spatial and temporal convolutions—and attention modules \cite{wang:modelscope:2023}. Tune-A-Video \cite{wu:tune:2023} modifies the style of an input video using text prompts, leveraging pretrained text-to-image models and attention tuning for temporal consistency. Early methods often exhibited flickering, as observed in the CVPR2023 text-guided video editing competition. Dreamix \cite{molad:dreamix:2023} mitigates this but produces blurry videos. CogVideo \cite{hong:cogvideo:2023} employs VQ-VAE to convert frames into tokens fused with text embeddings to generate new videos. Phenaki \cite{villegas:phenaki:2023} uses transformers for variable-length outputs, though with lower quality than DMs. Comprehensive evaluations appear in \cite{Liu:FETV:2023}. More recent work applies spatiotemporal layers to model dynamics \cite{Gupta:Photorealistic:2024}, redesigning transformer blocks for latent video diffusion with restricted spatial and spatiotemporal attention. LaVie \cite{wang2025lavie} shows that simple temporal self-attention, combined with rotary positional encoding, effectively captures temporal correlations. Image-to-video generation is analogous to text-to-video but conditions diffusion models on images rather than text; hybrid approaches combine textual descriptions (for motion) and images (for scene layout) \cite{wu2025customcrafter}.
An increasing number of free and commercial tools are emerging, including Veo 3 by Google DeepMind\footnote{\url{https://deepmind.google/models/veo/}}, Kling AI\footnote{\url{https://www.klingai.com/}}, Pika 2.2\footnote{\url{https://pikartai.com/pika-2-2/}}, and Hailuo AI\footnote{\url{https://hailuoai.video/}}. Though not perfect, their generated videos appear remarkably realistic.}

Generating characters with human posture and motion from text prompts has also become popular. Make-An-Animation \cite{Azadi:Make:2023} trains on image-text datasets and fine-tunes on motion capture data, adding additional layers to model the temporal dimension. Animate Anyone by Alibaba Group \cite{Hu_2024_CVPR} inputs a real photo or anime of a person with a sequence of guided poses. The results are significantly better than existing techniques, including Disco \cite{wang:disco:2024} and Bidirectionally Deformable Motion Modulation (BDMM) \cite{Yu:Bidirectionally:2023}. They also suggest using Animate Anyone  with Outfit Anyone\footnote{\url{https://humanaigc.github.io/outfit-anyone/}} to produce a character with a reference outfit.

Viggle\footnote{\url{https://viggle.ai/}} claims to be the first video-3D foundation model embodying an actual understanding of physics. It combines a character and a text prompt about motion to generate character animation. Available AI tools for 3D on the market include DeepMotion\footnote{\url{https://www.deepmotion.com/}} that offers text-to-3D post animation and video-to-3D post animation, shown in Fig. \ref{fig:Deepmotion_Vasa} (left). The later function can track multiple people from real video and generate replicated characters with the same motions.

% --------------------------------------------------
\subsubsection{Augmented, virtual and mixed reality, and 3D content}

While the benefits of LLMs in Augmented Reality (AR) directly target educational purposes, enhance cognitive support, and facilitate communication \cite{XU2025103402}, mixed reality (MR) has once again become exciting since the release of the Apple Vision Pro in February 2024. This demonstrated the potential of MR experiences by merging real-world environments with computer-generated ones. Thanks to the rapid growth of AI-based 3D representation (see Section \ref{ssec:3Dreconstruct}), the generation of AR/VR/MR content has advanced significantly. Real-time rendering with immersive interaction has improved, and real scenes can now be generated avoiding uncanny valley effects. There has also been an attempt to use autoregressive and generative models to estimate lighting, achieving a visually coherent environment between virtual and physical spaces in AR \cite{zhao2024clear}.

Similar to other content generation tools, LLMs have been influenced on immersive technologies, including text-to-3D and image-to-3D. Exciting examples include
Holodeck \cite{yang:Holodeck:2024}, which automatically generates 3D embodied environments via text-prompt interactions with a large language model (GPT-4). 3D objects are gathered from Objaverse \cite{deitke:Objaverse:2023}, a dataset with 800K+ annotated 3D objects. RealFusion \cite{Melas:RealFusion:2023}, a single image to 3D object generator, merges 2D diffusion models with NeRF, improving Instant-NGP \cite{mueller:instant:2022}, which provides an API for VR controls. NeuralLift-360 \cite{Xu:NeuralLift:2023} also uses diffusion models to generate priors for novel view synthesis. Magic123 \cite{Qian:Magic123:2024} is the latest image-to-3D tool that uses  2D and 3D priors simultaneously  to produce high-quality high-resolution 3D geometry and textures. DreamGaussian \cite{tang:dreamgaussian:2024} offers text-to-3D and image-to-3D by adapting 3D Gaussian splatting (more in Section \ref{sssec:3DGS}) into generative settings using a diffusion prior. This generates photo-realistic 3D assets with explicit mesh and texture maps within only 2 minutes. DreamGaussian4D \cite{ren:dreamgaussian4d:2023} employs image-to-video diffusion and a 4D Gaussian Splatting representation to generate an image-to-4D model. The results are not very sharp, but they can be further edited with Blender. 

In July 2024, Shutterstock launched its Generative 3D service in commercial beta, powered by NVIDIA Edify, a multimodal generative AI architecture. This service enables creators to rapidly prototype 3D assets and generate 360-degree HDRi backgrounds to light scenes using text or image prompts. In conjunction with OpenUSD, the created scenes can be rendered into 2D images and used as input for AI-powered image generators, allowing for the production of precise, brand-accurate visuals.

%For 3D virtual environment, SIMA by Google DeepMind build a versatile AI agent, mastering video games by understanding natural language instructions.

% ===================================================
%\subsubsection{Content and captions} Move to text generation

% ===================================================
\subsection{Information analysis}

\subsubsection{Text categorization}

Applications of text categorization include detecting spam emails, automating customer support, monitoring social media for harmful content, etc. At its core, text categorization involves assigning predefined labels to text documents, which can be anything from a tweet to a lengthy article. LLMs are particularly well-suited for this task due to their ability to comprehend complex and nuanced language. One of the main advantages of using LLMs in text categorization is their transfer learning capability. Models can be pre-trained on a large amount of text and then fine-tuned on a smaller, task-specific dataset, with or without further post-processing techniques. For example, CARP \cite{sun:text:2023} applies kNN to integrate a diagnostic reasoning process for final decision. ChatGraph, proposed by Shi et al. \cite{shi:chatgraph:2023}, utilizes ChatGPT to refine text documents. It uses a knowledge graph, extracted using another specifically defined prompt, and finally, a linear model is trained on the text graph for classification. Multiple learners are also used to enhance the performance \cite{Hou:promptboosting:2023,AI2025125952}.

\subsubsection{Advertisements and film analysis}

Not only does AI assist in generating ideas and content, but it can also aid creators in effectively matching content to their audiences, particularly on an individual level \cite{feizi:Online:2023}. This effectively helps in advertising personalization—eMarketer\footnote{\url{https://www.emarketer.com/content/spotlight-marketing-personalization}} reported that nearly nine out of ten consumers are comfortable with their browsing history being utilized to create personalized ads. In contrast to outdated syntax-style searches, advanced LLM tools can comprehensively grasp user intent behind each search through conversation prompts, providing advertisers with a high level of granularity.

Current advances in generative AI would greatly benefit sentiment analysis, also known as opinion mining, where opinions are gathered from social media, articles, customer feedback, and corporate communication and are analyzed to understand the emotion of the owners. This is a potential tool for filmmakers and studios, enabling the creation of effective and targeted marketing campaigns. By analyzing viewer emotions and opinions, AI can provide valuable insights into audience preferences, aiding in the optimization of film marketing strategies. Sentiment analysis with modern generative AI produces more accurate results. Technically, LLMs learn complex patterns and relationships in text data for sentiment classification \cite{Mao:Biases:2023, krugmann:sentiment:2024}. SiEBERT \cite{Hartmann:More:2023} provides a pre-trained model with open-source scripts to be fine-tuned to further improve accuracy for novel applications. Cinema Multiverse Lounge \cite{Ryu:Cinema:2025}, a multi-agent conversational system, allows users to interact with LLM-driven agents, each embodying a distinct film-related target user.

\subsubsection{Content retrieval and recommendation services}

Generative retrieval (GR) was pioneered by Metzler et al. \cite{Metzler:Rethinking:2021}. Unlike traditional retrieval, which adheres to the ``index-retrieve-then-rank" paradigm, the GR paradigm employs a single model to obtain results from query input. The model generally involves deep-learning based transformers, generating output token-by-token. More recent work in \cite{li2024learning} introduces learning-to-rank training to enhance the performance system up to 30\%.
GR has several advantages including substituting the bulky external index with an internal index (i.e., model parameters), significantly reducing memory usage, and enabling optimization during end-to-end model training towards a universal objective for information retrieval tasks. Conversational question answering techniques have been integrated to enhance the document retrieval \cite{li:unigen:2024}. 

When retrieving visual content, recent work exploits generative models to enhance content-based model search \cite{Lu:content:2023}. These models decode the text, image, or video query into samples of possible outputs, which are then used to learn statistics for better matching between the query and output candidates. DMs are also employed for visual retrieval tasks, where they learn joint data distributions between text queries and video candidates \cite{Jin:DiffusionRet:2023}. A comprehensive survey on Generative Information Retrieval is available in \cite{Li:From:2025}.

%b\subsubsection{Recommendation services}

While the retrieval task involves users directly defining a specific query input, recommendation services operate by retrieving content based on previous usage patterns. Essentially, a recommendation engine is a system that suggests products, services, or information to users through data analysis. Research in \cite{CHUA:AI:2023} has reported a positive association between buyers' attitudes toward AI and their behavioral intention to accept AI-based recommendations, with potential for further growth. Notable examples include the recommendation framework developed by Google \cite{Rajput:recommender:2023}, which utilizes GR. This framework assigns Semantic IDs to each item and trains a retrieval model to predict the Semantic ID of an item that a given user may engage with. A report by Aggarwal et al.~\cite{aggarwal2025evolution} states that the recommendation accuracy of recommendation services has increased from 45.0\% to 91.5\% with the integration of generative AI.

\subsubsection{Intelligent assistants}

Intelligent assistants refer to software programs or applications that use AI and NLP to interact with users and provide helpful responses or perform tasks. These assistants can range from simple chatbots to sophisticated virtual agents capable of understanding and responding to complex queries. They're designed to assist users in various tasks, from answering questions and providing information to scheduling appointments and controlling smart home devices.

Current LLMs obviously enhance the performance of intelligent assistants, designed to understand complex inquiries and generate more natural conversational responses, such as Sasha \cite{King:Sasha:2024}. Generative AI can also be used to enhance the performance of human customer support agents, aiding in search and summarization, as discussed in the previous section. Brynjolfsson et al. \cite{brynjolfsson:generative:2023} examined the implementation of a generative AI tool designed to offer conversational guidance to customer support agents. Their research revealed that AI assistance significantly enhances problem resolution and customer satisfaction. Furthermore, they observed that AI recommendations prompt low-skill workers to adopt communication styles akin to those of high-skill workers. AI-based intelligent assistants may currently be more focused on educational purposes, but they can clearly help artists write more efficiently \cite{Lee:design:2024} or assist in customizing personal requirements \cite{sajja2024ai}. The performance of personalized assistants can be enhanced with domain-specific knowledge to provide more in-depth responses to users \cite{jiang2025domain}.
 
% ===================================================
\subsection{Content enhancement and post production workflows}

\subsubsection{Enhancement}

In our previous review paper \cite{Anantrasirichai:AI:2022}, we discussed AI technologies for contrast enhancement and colorization as separate topics, as methods were developed specifically for each task. However, in recent years, there has been a shift towards addressing more complex issues, such as those encountered in low-light environments and underwater scenarios. These real-world situations often involve a combination of challenges, including low contrast, color imbalance, and noise.

In low-light conditions, scenes often exhibit low contrast, leading to focusing difficulties or the need for long exposures, which can result in blurred images and videos. To address this, LEDNet \cite{Zhou:LEDNet:2022} has introduced a synthetic dataset for such scenarios and incorporated a learnable non-linear activation function within the network to enhance feature intensities. Meanwhile, SNR-Aware \cite{Xu:SNR:2022} estimates spatial-varying Signal-to-Noise Ratio (SNR) maps and proposes local and global learning branches using ResNet and transformer architectures, respectively. NeRCo \cite{Yang:Implicit:2023} addresses the low-light problem with INR, which unifies the diverse degradation factors of real-world scenes with a controllable fitting function.  Diffusion models (DMs) have also become popular choices for low-light image enhancement \cite{HOU:Global:2023,Yi:Diff:2023,Jiang:Low:2023}. Diff-Retinex \cite{Yi:Diff:2023} formulates the low-light image enhancement problem into Retinex decomposition, and employs multi-path generative diffusion networks to reconstruct the normal-light Retinex probability distribution. A recent state-of-the-art approach presented in \cite{Jiang:Low:2023} decomposes images into high and low frequencies using wavelet transform. High frequencies are enhanced using a transformer-based pipeline, while the low frequencies undergo a diffusion process. This method achieves nearly 2.8dB improvement over the state-of-the-art transformer-based approach, e.g. LLFormer \cite{Wang:Ultra:2023}, and significantly better than  INR-based method, NeRCo \cite{Yang:Implicit:2023}, on a real low-light image benchmarking dataset. The technique has been extended for video enhancement in \cite{lin2024lowlight}. The output of the enhancement typically depends on user preferences. This has been viewed as a one-to-many inverse problem, with attempts to solve it using Bayesian approaches. For example, a Bayesian Enhancement Model (BEM) \cite{huang2025bayesian} incorporates Bayesian Neural Networks (BNNs) to capture data uncertainty and produce diverse outputs. The method can be used with Transformers or Mamba as the architecture backbone. 

Regarding video enhancement, transformer and DMs are still in their early stages. STA-SUNet \cite{Lin:SPATIO:2024} has demonstrated that using transformers for low-light video enhancement outperforms CNN-based methods \cite{anantrasirichai:BVI:2024}.  The recent Mamba-based network \cite{huang2025bvi} also demonstrates promising results, outperforming STA-SUNet by more than 2 dB in PSNR. It is important to note that low-light enhancement is subjective. While most training datasets use normal lighting conditions as ground truth \cite{Lin:BVI-RLV:2024}, the enhanced images and videos may alter the mood and tone of the content. Therefore, the tools for creative industries should be adjustable, not only for entire images and videos but also adaptive to specific areas and content. For instance, CLE Diffusion \cite{Yin:CLE:2023} enables user-friendly editing of lighting with fine-grained regional controllability. 

Recent efforts have focused on enhancing User-Generated Content (UGC) videos—authentic recordings created by individuals rather than brands, often showcasing real experiences with products or services. The winning solution of the NTIRE 2025 Challenge on UGC Video Enhancement \cite{safonov2025ntire} implemented a pipeline of four sequential modules: color enhancement, denoising, BasicVSR++ restoration \cite{Chan:BasicVSR:2022}, and SwinIR \cite{Liang:SwinIR:2021}. This method achieved a 17\% higher subjective score than the second-place entry, which used a two-stage framework, highlighting a notable improvement in perceived visual quality.

% --------------------------------------------------------
\subsubsection{Style transfer}

Style transfer in AI art refers to a technique where the artistic style of one image (or video) is applied to another image (or video) while preserving the content of the latter. Style transfer has numerous applications in art, design, and image editing, allowing artists and designers to create unique and visually appealing compositions by blending different artistic styles with existing images (or videos). The applications also include image-to-image and sequence-to-sequence translations.

StyTr2 \cite{Deng:StyTr2:2022} is the first transformer-based method for style transfer, applying content as a query and style as a key of attention. InST \cite{Zhang:Inversion:2023} utilizes Stable Diffusion Models as the generative backbone and introduces an attention-based textual inversion module to learn the description of the content. StableVideo \cite{Chai:StableVideo:2023} uses a text prompt to describe the desired appearance of the output, transforming the input video to have a new look based on a diffusion model. For instance, a video of a white car driving in summer can be altered to show a red car driving in winter. A large pre-trained DM is employed in \cite{Chung_2024_CVPR}, where the style is injected to manipulate the self-attention of the decoder. To deal with the disharmonious color, they propose an adaptive instance normalization. A survey of style transfer using transformers and diffusion models can be found in \cite{ZHOU:Bridging:2025}. Implicit Neural Representations (INRs) are less commonly used in style transfer tasks due to the difficulty of modeling the cross-representation between style and content. Moon et al. \cite{Moon:generalizable:2023} combined INRs with vision transformers for generalizable style transfer; however, the results remain limited in quality. In contrast, the method proposed by Kim et al. \cite{Kim:Controllable:2024} uses multilayer perceptrons (MLPs) to map image coordinates to the colors of the stylized output, guided by features extracted from both the content and style inputs to allow controllability.

% --------------------------------------------------------
\subsubsection{Upscaling imagery: super-resolution (SR)}

Impressive super-resolution (SR) results from transformer and diffusion models have been published extensively in the past few years. Originally, SR methods were developed using multiple low-resolution (LS) images, as different features in each image are combined to construct an enhanced one. However, these methods are not practical, as in most cases only one LS image is available. Hence, more methods have been developed for single image super-resolution (SISR).

The first use of a transformer, called ESRT, was for capturing long-term dependencies, such as repeating patterns in buildings. This was done in the feature domain extracted by a lightweight CNN module \cite{Lu:Transformer:2022}, outperforming those that use only CNNs. Since then, most SISR methods have been based on transformers. The Hybrid Attention Transformer (HAT) \cite{Chen:Activating:2023} was introduced, which improves the SR quality over ESRT by more than 2dB when upscaling 2$\times$-4$\times$. However, the NTIRE 2023 Real-Time Super-Resolution Challenge \cite{Conde:Efficient:2023} showed that the winner, Bicubic++ \cite{Bilecen:Bicubic:2023}, uses only convolutional layers and achieves the fastest speed at 1.17ms in upscaling 720p to 4K images. This method is significantly faster than any of the participants in the NTIRE 2025 Challenge \cite{chen2025ntire}, where Transformer-based architectures continue to dominate as the mainstream approach.

For DMs, SR3 by Google \cite{Saharia:image:2023} has produced truly impressive results. It operates by learning to transform a standard normal distribution into an empirical data distribution through a sequence of refinement steps, interpolating in a cascaded manner—upscaling 4$\times$ at a time. Later, IDM \cite{Gao:Implicit:2023} combines INR with a U-Net denoising model in the reverse process of the DM. It is crucial to emphasize again that DMs are generative models. The SR results are generated based on the statistics we provide to the model during training (LR training samples). This is not for a restoration task, but rather for synthetic generation. A survey in SISR using DMs can be found in \cite{moser:diffusion:2024}.

For video SR, numerous methods have emerged as part of a unified enhancement framework, as discussed in the previous section. One of the pioneering works to incorporate transformers specifically for video SR tasks is the Trajectory-aware Transformer for Video Super-Resolution (TTVSR) \cite{Liu:Learning:2022}. Although the results are slightly inferior to those of BasicVSR++ \cite{Chan:BasicVSR:2022}, which employs CNN and was introduced around the same time, both methods significantly enhance detail and sharpness compared to previous approaches, albeit not in real time. To address this limitation, the Deformable Attention Pyramid \cite{Fuoli:Fast:2023} has been introduced, offering slightly lower quality but a speed-up of over 3$\times$. Recently, Adobe announced their VideoGigaGAN \cite{xu:videogigagan:2024}, which can perform 8$\times$ upsampling. This is achieved by adding flow estimation and temporal self-attention to the GigaGAN upsampler \cite{kang:gigagan:2023}, which is primarily used for image SR, and text-to-image synthesis. Cao et al. \cite{cao2025zero} introduce a zero-shot video super-resolution framework that leverages a pre-trained image diffusion model, and replaces the spatial self-attention layer with a novel short-long-range (SLR) temporal attention layer. Recently, SeedVR integrated text information (captions) into a Diffusion Transformer (DiT) model, achieving state-of-the-art performance in video super-resolution.

\begin{figure}
    \centering
    \includegraphics[width=\textwidth]{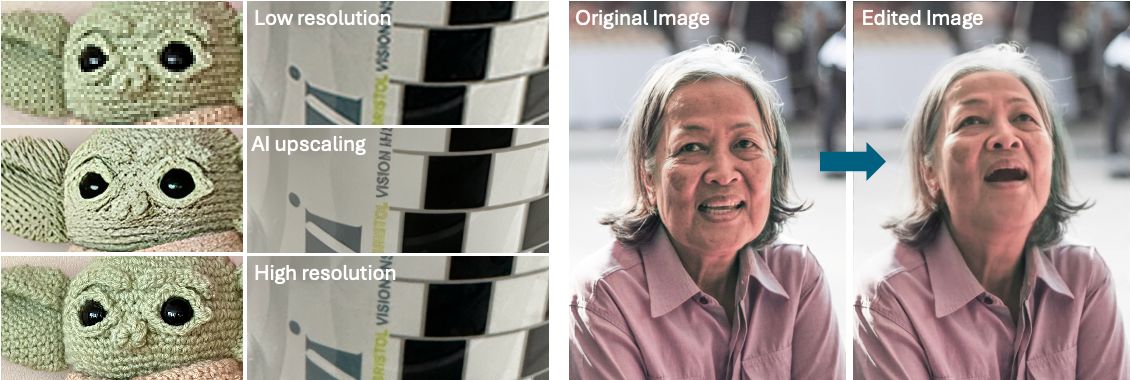}
    \caption{(Left) Examples of SR ($\times$4) using generative model. (Right) Real-time portrait editing with FacePoke.} 
    \label{fig:SR}
\end{figure}

Compared to traditional upscaling methods, generative AI can add details that did not exist in the original input image. These methods excel at generating high-quality natural images and structures, such as buildings, which are commonly included in training datasets. However, the process can be slow and may produce unpredictable results if the input image has very low resolution or contains content rarely seen in natural images. As shown in Fig. \ref{fig:SR} (left), generative AI fails to upscale the knitting texture areas, instead generating lines more commonly found in typical images. While AI methods produce sharper edges, they perform less effectively on text.

\subsubsection{Restoration}

In our previous review paper \cite{Anantrasirichai:AI:2022}, we categorized the work on restoration into several different types of distortions, including deblurring, denoising, dehazing, and mitigating atmospheric turbulence. Recent work, however, uses a unified network architecture to address these as inverse problems $y = hx + n$, where $x$ and $y$ are the ideal and observed data, respectively. $h$ is a degradation function, such as blur, and $n$ is additive noise. Often, the super-resolution task is also considered as an inverse problem, meaning $h$ includes the downsampling process. Note that although designed as a single network, the model is trained with each distorted dataset separately. 

The pioneering transformer-based method for image restoration, SwinIR \cite{Liang:SwinIR:2021}, employs several concatenated Swin Transformer blocks \cite{Liu:Swin:2021}. SwinIR surpasses state-of-the-art CNN-based methods proposed up to the year 2021 in super-resolution and denoising tasks. The model is smaller and reconstructs fine details more effectively.
Other two popular approaches that emerged in the same timeframe are Uformer \cite{Wang:Uformer:2022} and Restormer \cite{Zamir:Restormer:2022}. Both incorporate Transformer blocks into hierarchical encoder-decoder networks, employing skip connections similar to those in U-Net. Their objective was to restore noisy images, sharpen blurry images, and remove rain. The networks focused on predicting the residual $R$ and obtaining the restored image $\hat{x}$ through $\hat{x} = y + R$. While their performance is very similar, Restormer has half the parameters of Uformer. More recently, GRL by Li et al. \cite{li:GRL:2023} exploits a hierarchy of features in a global, regional, and local range using different ways to compute self-attentions as an image often show similarity within itself in different scales and areas, which outperforms SwinIR and Restormer. Additionally, Fei et al. introduced the Generative Diffusion Prior \cite{Fei:Generative:2023} for unsupervised learning, aiming to model posterior distributions for image restoration and enhancement. VmambaIR \cite{Shi:VmambaIR:2025} incorporates Mamba blocks into the U-Net architecture, achieving superior performance compared to SwinIR and Restormer in both visual quality and model size.

For video restoration, the general framework comprises frame alignment, feature fusion and reconstruction. The process could be similar to image restoration but input multiple frames and run through the sequences in a sliding window manner to exploit the temporal information within a number of consecutive frames. 
Recently, Video Restoration Transformer (VRT) \cite{Liang:VRT:2024} and its improved version with a recurrent process (RVRT) \cite{liang:RVRT:2022}, have emerged as the state of the arts for video super-resolution, deblurring, denoising, and frame interpolation. This method introduces temporal reciprocal self-attention in the transformer architecture and parallel warping using MLP. These innovations enable parallel computation and outperform the previous state-of-the-art methods by up to 2.16dB on benchmark datasets. FMA-Net \cite{Youk:FMA:2024} proposed multi-attention for joint video super-resolution and deblurring, achieving fast runtime with nearly 40\% improvement over RVRT, and the restored quality was reported better by up to 3\%. 

For audio restoration, most software discussed in Section \ref{sssec:musicgen} offers tools for enhancing audio quality, such as eliminating background noise, echo, microphone rumble, and occasionally room reverberation, which have been well-established even before the advent of deep learning. There have been efforts to utilize AI for learning global contextual information to aid in the removal of unwanted sounds, leading to better final quality \cite{Yu:DBT:2022}. The latest advancements in this domain are primarily focused on addressing issues where significant portions of the audio data are missing. For instance, Moliner et al. \cite{Moliner:solving:2023} tackle problems such as audio bandwidth extension, inpainting, and declipping by treating them as inverse problems using a diffusion model. For a comprehensive survey on the use of diffusion models in restoration tasks, refer to \cite{10902142}. 

The following methods have been proposed for specific problems, but ideally, they should be adaptable for other tasks, even though they may not perform as well as they do for the original task.
 
i) \textbf{Deblurring}: A lightweight deep CNN model was recently proposed in \cite{Pan:Deep:2023}, where a new discriminative temporal feature fusion has been introduced to select the most useful spatial and temporal features from adjacent frames. Feature propagation along the video is done in the wavelet domain. The deblurring performance is comparable to RVRT \cite{liang:RVRT:2022}, but it is 5 times faster. DaBiT \cite{Morris:DaBiT:2024} mitigates focal blur content with depth information and applies SR for further enhancing fine details.
Note that not only in software, but AI technologies have also been integrated into hardware. This includes autofocus, which is crucial for capturing sharp images of subjects, especially in dynamic environments where manual adjustments are impractical due to rapid movement. AI-driven autofocus methods have emerged, often tailored for specific camera hardware. For instance, Choi et al. proposed an autofocus model optimized for dual-pixel Canon cameras \cite{Choi:Exploring:2023}. Additionally, Yang et al. investigated the correlation between language input and blur map estimation, utilizing semantic cues to enhance autofocus performance \cite{yang:ldp:2023}. Remarkably, their model achieves comparable results to previous state-of-the-art methods while being more lightweight \cite{Yang:K3DN:2023}. Autofocus could be used in conjunction with real-time object tracking (see Section \ref{sssec:tracking}) to produce desirable sharpness for moving objects in the video. Recently, Feng et al. \cite{feng2025residual} proposed a novel residual diffusion deblurring framework that integrates a conditional diffusion model guided by a defocus map and incorporates residual learning into the single-image defocus deblurring process.

ii) \textbf{Denoising}: SUNet \cite{Fan:SUNet:2022} applies Swin transformer blocks combined in a UNet-like architecture. Denoising with diffusion models (DMs) \cite{yang:realworld:2023} has been proposed by diffusing with estimated noise that is closer to real-world noise rather than Gaussian noise, achieving better performance than SwinIR \cite{Liang:SwinIR:2021} and Uformer \cite{Wang:Uformer:2022}. INR with complex Gabor wavelets as activation functions show promising denoising results \cite{Saragadam:wire:2023}. The NTIRE 2025 Image Denoising Challenge \cite{sun2025tenth} revealed that the top-performing methods combined transformer-based and convolutional network architectures. Similarly, recent advances in video denoising also adopt a hybrid approach that integrates both architectures \cite{Jin:Masked:2025,  Yue:RViDeformer:2025}.

iii) \textbf{Dehazing}: Vision transformers for single image dehazing were proposed in DehazeFormer \cite{Song:vision:2023}. Similar to SUNet, it is a UNet-like architecture, but introduces Rescale Layer Normalization for better suit on improving contrast. The Fast Fourier Transform (FFT) has been employed in \cite{fang2025guided} due to the phase spectrum conveying more structural detail than the amplitude spectrum and demonstrating greater robustness to contrast distortion and noise. Then cross-attention between the RGB and YCbCr color spaces is applied. This approach achieves nearly 5 dB higher PSNR than DehazeFormer on a real-world smoke dataset.  For video dehazing, Xu et al. \cite{Xu:Video:2023} introduced a recurrent multi-range scene radiance recovery module with the space-time deformable attention. They also employ physics prior to inform haze attenuation. This method outperforms DehazeFormer by approximately 1dB.

iv) \textbf{Mitigating atmospheric turbulence}: Similar to dehazing, physics-inspired models have been widely developed to remove turbulence distortion \cite{Jaiswal:Physics:2023,Jiang:NeRT:2023}, while complex-valued CNNs have been proposed to exploit phase information \cite{Atmospheric:2023}. There was also an attempt to use instance normalization (INR) to solve this problem, offering tile and blur correction \cite{Jiang:NeRT:2023}. However, diffusion models outperform on a single image \cite{Nair:AT-DDPM:2023}, and transformer-based methods remain state-of-the-art for restoring videos \cite{Zhang:Image:2024, zou2024deturb}. Mamba architecture, employed in \cite{Hill2025MAMAT}, outperforms Transformers and improves object detection performance. A recent review can be found in \cite{Hill2025}.

\subsubsection{Inpainting}
\label{ssec:inpaiting}

Visual inpainting is the process of filling in lost or damaged parts of an image or video. CNNs and GANs have already achieved impressive results (see our previous review paper \cite{Anantrasirichai:AI:2022}). Recent work has focused more on editing rather than simply filling in the missing areas. This means users can now mask large areas of an image, and AI tools generate multiple results for users to choose from, a technique known as pluralistic inpainting \cite{zheng:pluralistic:2019}. Some notable methods include the following: Mask-Aware Transformer (MAT) \cite{Li:MAT:2022} offers several outputs to fill a large missing area, consisting of a convolutional head, a transformer body, and a convolutional tail for reconstruction, along with a Conv-U-Net for refinement. PUT \cite{Liu:Reduce:2022} proposes a patch-based vector VQ-VAE and unquantized Transformer to minimize information loss. Spa-former \cite{HUANG:Sparse:2024} employs a UNet-like architecture, where each level performs transformer with sparse self-attention to remove coefficients with low or no correlation, leading to memory reduction, while improving result quality by up to 5\% compared to PUT.

Video inpainting presents greater complexity compared to image inpainting, despite the abundance of information available in an image sequence. The process typically involves tracking masks across frames, estimating optical flow, and ensuring temporal consistency.  The current state-of-the-art methods include DLFormer \cite{Ren:DLFormer:2022} and  ProPainter \cite{Zhou:ProPainter:2023}. DLFormer conducts inpainting in latent space and utilizes discrete codes for video representation. On the other hand, ProPainter employs flow-based deformable alignment to enhance robustness to occlusion and inaccurate flow completion. The method excels in filling complete and rich textures, achieving a speed of 12 fps for full HD video. Video inpainting is also used for dubbing. DINet \cite{Zhang:DINet:2023} replaces the mouth area to synchronize with a new language being spoken.

A comprehensive survey of learning-based image and video inpainting, covering approaches such as CNNs, VAEs, GANs, transformers, and diffusion models, can be found in \cite{quan:deep:2024}. Additionally, Elharrouss et al. \cite{elharrouss2025transformer} provide an in-depth review of the current challenges and future directions specific to transformer-based inpainting techniques.

%----------------------------------------------------
\subsubsection{Image Fusion}
\label{sssec:fusion}

Image fusion is the process of  merging multiple images from either the same source (such as varying focal points or exposures) or different modalities (e.g. visible and infrared cameras) into a single image. This process integrates complementary information from the various images to enhance overall quality, improve interpretation, and increase the usability of the final image.

Transformers and CNNs have been combined to extract global and local information, respectively. Most methods use CNNs for feature extraction, with transformers operating in the latent space \cite{Ma:SwinFusion:2022, Rao:TGFuse:2023}. Notable methods include SwinFusion \cite{Ma:SwinFusion:2022}, which utilizes a self-attention-based intra-domain fusion unit and a cross-attention-based inter-domain fusion unit to achieve multi-modal and digital photography image fusion. Transformer-based image fusion has also been applied to downstream tasks like segmentation \cite{Liu:Multi:2023}, achieving superior results by leveraging the additional information. Self-attention blocks are employed to enhance intra-feature representations, while the cross-attention mechanism integrates inter-feature information to improve the quality of the fused output \cite{LI2024102147}.

DDFM, the first diffusion model-based image fusion method, estimates noise in the reverse process by combining multiple inputs \cite{Zhao:DDFM:2023}. The expectation-maximization (EM) algorithm is integrated to estimate the noise distribution parameters, resulting in sharper images compared to traditional DDPM. For an in-depth review, the reader is referred to recent work in \cite{Karim:Current:2023, Zhang:Visible:2023}.

%----------------------------------------------------
\subsubsection{Editing and Visual Special Effects (VFX)}

Editing or modifying specific areas of an image is much easier with DM technologies, particularly for headshot photos, such as targeting the eyes and mouth on the face \cite{guo2024liveportrait}. This capability has been extended to video generation (see Section \ref{sssec:videogen}). Fig. \ref{fig:SR} shows an example of the online tool, FacePoke\footnote{\url{https://huggingface.co/spaces/jbilcke-hf/FacePoke}}, which allows users to move the head and modify the shapes of the eyes and mouth in real time. Motion-I2V \cite{Shi:Motion-I2V:2024} provides motion blur and motion drag tools to control specific areas of an image to add motion. The method is based on a diffusion-based motion field predictor and motion-augmented temporal attention.

VFX aims to create and/or manipulate imagery outside the context of a live-action shot in filmmaking and video production. When adding objects, scenes, and effects into traditional photographic videos, generative AI has obviously become an important tool, but some manual operations are still required. For example, in After Effects (EA)\footnote{\url{https://www.adobe.com/uk/products/aftereffects.html}}, the user selects the area where the object will be added and uses text prompts to describe such object. Subsequently, with the current EA version, the user will need to apply motion tracking so the generated object is moved accordingly.

AI technologies can upscale, enhance, and restore low-quality or old footage. For example, standard definition videos can be converted to high definition or even 4K quality without traditional manual remastering processes. This is particularly useful for remastering old movies or enhancing visual details in scenes. Generative AI has also simplified and accelerated automated processes, such as rotoscoping \cite{Tous:Lester:2024}, an animation technique where animators trace over motion picture footage frame by frame to create realistic action. AI models can accurately detect and segment objects and characters in video frames, significantly speeding up the post-production process. Additionally, AI can assist the rapid creation of 3D models from 2D images generating realistic animations with minimal input data, facilitating complex human motions and synchronized facial expressions to voiceovers. One restriction is that current technologies still cannot yet generate full 4K accurate visual effects.

%However, video generation is obviously one of the fastest grow technologies in 2024 (refer to section \ref{sssec:videogen}).

% ===================================================
\subsection{Information Extraction and Understanding}
\label{ssec:infoextract}

\begin{figure}
    \centering
    \includegraphics[width=\textwidth]{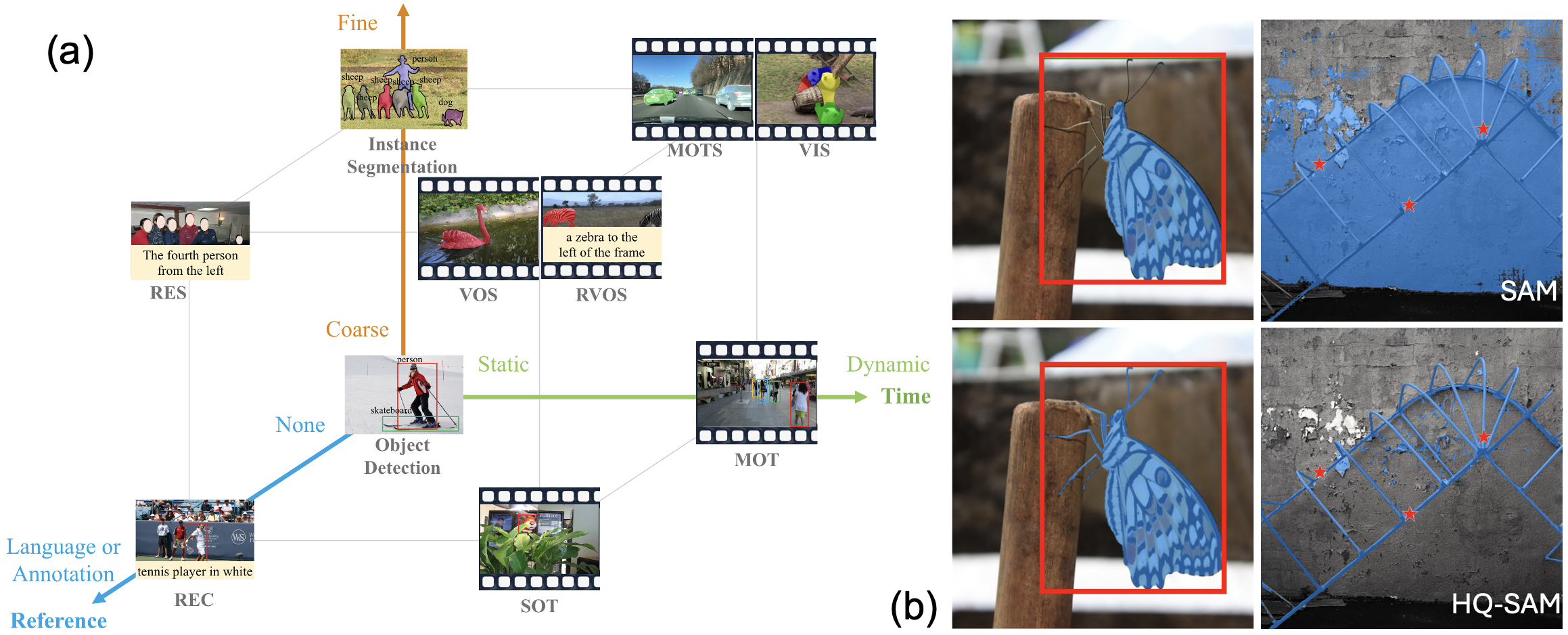}
    \caption{(a) Tasks in Object-centric understanding defined by Yan et al. \cite{Yan:Universal:2023} (REC:,  Referring Expression Comprehension, RES: Referring Expression Segmentation, VOS:  Video Object Segmentation, RVOS: Referring Video Object Segmentation, MOT: Multiple Object Tracking, MOTS: Multi-Object Tracking and Segmentation, VIS: Video Instance Segmentation, SOT: Single Object Tracking. (b) Current high-quality segmentation \cite{Ke:SAM-HQ:2023}.}
    \label{fig:segmentation}
\end{figure}

AI plays a crucial role in automating and optimizing the process of information extraction and understanding, enabling organizations to derive actionable insights from large and diverse data. Yan et al. \cite{Yan:Universal:2023} have categorized information extraction tasks based on the Format-Time-Reference space, as illustrated in Fig. \ref{fig:segmentation} (a), where object detection and video object segmentation (VOS) are considered to be the simplest and the most complex tasks, respectively. Recent advancements in this field draw significant inspiration from LLMs. These advancements include the utilization of prompts as conditional inputs for acquiring information. Moreover, following the pipeline approach used in LLMs, there is a growing trend towards leveraging very large datasets to pre-train models before fine-tuning them for specific downstream tasks. For instance, Meta AI \cite{Oquab:DINOv2:2024} has introduced DINOv2, aimed at enriching information about visual content through self-supervised learning. This model was trained with 142 million carefully selected images, employing the ViT architecture. Google have introduced VideoPrism \cite{zhao:videoprism:2024}, a tool for scene understanding including classification, localization, retrieval, captioning, and question answering (QA). The model was trained on an extensive and diverse dataset consisting of 36 million high-quality video-text pairs and 582 million video clips accompanied by noisy or machine-generated parallel text.

%Dataset \cite{Athar:BURST:2023}

%----------------------------------------------------
\subsubsection{Segmentation}
\label{sssec:seg}

The need for segmentation has grown dramatically in the past few years, given its central role in visual perception. Many segmentation methods now integrate an input prompt for users to define their preferred output appearances, such as pixel-wise segmentation, bounding boxes around objects, or segmented areas of interest. Most of these methods utilize transformer architectures \cite{Bowen:Marked:2022}. Among them, Segment Anything (SAM) by Meta AI \cite{Kirillov:SAM:2023} stands out as a pioneer in promptable segmentation approaches. This method computes masks in real-time and has been trained with over 1 billion masks across 11 million images, facilitating transferability from zero-shot to new image distributions and tasks. HQ-SAM \cite{Ke:SAM-HQ:2023} enhances SAM by incorporating global-local feature fusion, leading to high-quality mask predictions. SegGPT \cite{Wang:SegGPT:2023} proposed context ensemble strategies and allows users to tune a prompt for a specific dataset, scene, or even a person, while SEEM \cite{Zou:Segment:2023} provides a completely promptable and interactive segmentation interface. More recently, SAM2 \cite{ravi2024sam2} introduced support for real-time video segmentation. It is a unified model trained on a larger dataset than SAM. Interactive tools enable users to mark areas of interest and specify regions to exclude from the segmentation map. Zhou et al. propose an audio-visual segmentation (AVS) to generate pixel-level segmentation masks for sounding objects in audible videos. DVIS++ by \cite{Zhang:DVISp:2025} introduces a universal video segmentation framework capable of producing instance, semantic, and panoptic segmentation outputs. This transformer-based architecture comprises a segmentor, tracker, and refinement module, achieving state-of-the-art performance across several video segmentation benchmarks.

With DMs tehchnologies, Baranchuk et al. \cite{Baranchuk:label:2022} have investigated semantic representation, and found DMs outperform other few-shot learning approaches. DiffuMask \cite{Wu:DiffuMask:2023} automatically generate image and pixel-level semantic annotation using pre-trained Stable Diffusion with input as a text prompt. It has been proven that using these synthetic data improve segmentation accuracy. Currently, the state-of-the-art panoptic segmentation is the method developed by Nvdia, which is based on text-to-image DMs \cite{Xu:Open:2023}, outperforming the previous methods by up to 7.6\%.

Applying INRs to segmentation is more popular in the medical domain, as the specific signals used, such as computed tomography (CT) and magnetic resonance imaging (MRI), can be formulated as continuous functions. In creative technologies, unsupervised domain adaptation (UDA) and INRs are used for continuous rectification function modeling in \cite{Gong:Continuous:2023}, achieving superior segmentation results in night vision. Recently, this work has been integrated with a non-local means block in \cite{Lin:Feature:2024}  showning a significant improvement for instant segmentation in low-light scenes.

3D segmentation is also crucial for scene manipulation. In radiance fields, earlier segmentation methods required additional modules such as using k-means clustering to separate objects from the background \cite{Goel:Interactive:2023}. However, the recent SA3D approach \cite{Cen:Segment:2023} segments 3D objects using NeRFs as the structural prior. SA3D operates by taking a trained NeRF and a set of prompts from a single view, then performing an iterative procedure. This involves rendering novel 2D views, self-prompting SAM for 2D segmentation, and projecting the segmentation back onto 3D mask grids. A comprehensive survey of 3D segmentation in computer vision can be found in \cite{HE2025102722}.

%----------------------------------------------------
\subsubsection{Detection and recognition}
\label{sssec:recog}

Introduced in 2020, DETR by Facebook AI \cite{Carion:DERT:2020} was one of the first to adopt a transformer architecture for object detection. The approach achieves comparable results to an optimized Faster R-CNN \cite{Ren:Faster:2027}, introduced in 2015. Deformable convolution has alson been used, (Deformable DETR \cite {Zhu:Deformable:2021}), resulting in training faster with approximately 5\% accuracy improvement.
A survey until 2022  \cite{Zou:object:2023} reported that Deformable DETR and Swin Transformers \cite{Liu:Swin:2021} outperform pure CNN-based YOLOv4 \cite{bochkovskiy2020yolov4}. 
SwinV2 improves the first version by replacing original dot product attention with scaled cosine attention, improving accuracy by approximately 5\%. Later, RT-DETR \cite{lv2:detrs:2024} improved inference speed by decoupling the intra-scale interaction and cross-scale fusion of features with different scales. RT-DETR is 25\% faster than YOLOv8\footnote{\url{https://github.com/ultralytics/ultralytics}} with 6\% improvement on MS COCO Object Detection dataset. Recently, YOLOv10 \cite{wang:yolov10:2024} has been released. YOLOv10 further improves the speed of detection approximately by 30\% over RT-DETR with the same accuracy. A review of transformer-based methods for object detection can be found in \cite{Li:Transformer:2023, KHEDDAR2025103347}. Recently, YOLO12 \cite{tian2025yolov12} introduced an attention-centric architecture, achieving a 2.1\% and 1.2\% mAP improvement over YOLOv10-N and YOLOv11-N respectively, with only a slight decrease in speed.

To detect 3D objects, the transformer-based method MonoDTR \cite{Huang:MonoDTR:2022} incorporates depth estimation from a single 2D image \cite{Yang:depthanything:2024} to predict 3D bounding boxes. More 3D object detection methods have been developed for autonomous driving \cite{10637966}; however, these approaches can also be adapted for AR and VR applications \cite{im2025gate3d}.

While DMs are primarily used to generate synthetic datasets \cite{Wu:datasetDM:2023, Fang:Data:2024}, they have also been demonstrated to function as zero-shot classifiers by Li et al. \cite{Li:Your:2023}. DMs are also of interest for detection tasks, Although feature extractors are still predominantly based on CNNs, such as ResNet, or Transformers (like Swin). DiffusionDet \cite{Chen:DiffusionDet:2023} formulates object detection as a denoising diffusion process from noisy boxes to object boxes, reporting performance that surpasses DETR. DMs have also been employed for anomaly detection \cite{Zhang:DiffAD:2025, WU2025102965}, functioning similarly to zero-shot classifiers.

%----------------------------------------------------
\subsubsection{Tracking}
\label{sssec:tracking}

Object tracking stands out as one of the tasks that greatly benefits from transformers since \textit{attention} is needed in both space and time. An experimental survey cited in \cite{Kugarajeevan:Transformers:2023} reveals that transformer-based methods consistently rank at the top of the leaderboard across various datasets. In the Visual Object Tracking (VOT) challenges of 2023\footnote{\url{https://eu.aihub.ml/competitions/201\#results}}, all of the top-10 employed transformer-based methodologies. The highest-performing approach achieved a 10\% improvement in tracking quality compared to the winner in 2020. The current state-of-the-art for single-object tracking\footnote{\url{https://paperswithcode.com/sota/visual-object-tracking-on-lasot}}, however, is based on cross-attention and Mamba \cite{kang2025exploring}.

The first three tracking-by-attention approaches are TrackFormer \cite{Meinhardt:TrackFormer:2022}, MixFormer \cite{cui:mixformer:2022}, and ToMP \cite{Mayer:Transforming:2022}. TrackFormer extracts visual features using a CNN-based encoder, which are then tracked using a vanilla transformer \cite{Vaswani:attention:2017} in a frame sequence, while MixFormer introduces cross-attention between the target and search regions. ToMP tracks the objects using prediction aspects. Many more methods have been proposed, including SeqTrack \cite{Chen:SeqTrack:2023} and Track Anything Model (TAM) \cite{yang:track:2023}. SeqTrack extracts visual features with a bidirectional transformer, while the decoder generates a sequence of bounding box values autoregressively with a causal transformer. TAM combines SAM \cite{Kirillov:SAM:2023} and XMem \cite{cheng:xmem:2022}, offering tracking and segmentation performance on the human-selected target. However, the masked area is still not very sharp, and there is a subtle degree of temporal inconsistency. MOTRv2 \cite{Zhang:MOTRv2:2023} combines YOLOX \cite{ge:yolox:2021} for object recognition and MOTR \cite{zeng:motr:2022} for tracking, outperforming TrackFormer by 20\%. Additionally, some methods have been specifically proposed for challenging environments, such as low light \cite{Yi:Comprehensive:2024} and small objects, as seen in AnyFlow \cite{Jung:AnyFlow:2023}. The latter exploits INR to upsample a continuous coordinate-based flow map, similar to SISR technique proposed in \cite{Chen:Learning:2021}.

Similarly to detection tasks, DMs for tracking tasks are used as downstream processes by concatenating the diffusion head to the feature extraction backbone. However, a spatial-temporal fusion module has been added to the diffusion head to exploit temporal video features\cite{Luo:DiffusionTrack:2024}. DiffusionTrack \cite{Xie:DiffusionTrack:2024} localizes the target in a progressive diffusion manner, which is claimed to better handle challenging scenarios. The method in \cite{Zhang:DiffusionTracker:2024} exploits spatial-temporal weighting to suppress the probability of the tracker changing the target to the distractors. It, however, reports underperformance compared to  MixFormer.

% ===================================================
\subsection{3D Reconstruction and Rendering}
\label{ssec:3Dreconstruct}

Bridging the gap between digital and physical realms, 3D reconstruction and rendering are integral to various creative technologies.  In film and animation, they enable the creation of detailed digital models that blend seamlessly with live-action footage. Video games and digital twins leverage these technologies for dynamic environmental rendering. VR and AR use 3D reconstruction to create immersive and interactive experiences, with AR integrating digital content into real-world contexts. With recent AI technologies, 3D reconstruction and rendering have become faster and closer to reality. In particular, neural radiance fields and Gaussian Splatting enable artists and film producers to create shots that cannot be done in the real shooting environments.

% -------------------------------
\subsubsection{Depth Estimation}
\label{sssec:depth}

Accurate depth information (alongside texture data) is typically required to construct 3D models. Depth sensors, such as lidar (Light Detection and Ranging) and structured-light 3D scanners, can be used for this purpose, but their applications are often limited by distance and cost. Consequently, vision-based sensors have become widely used. These sensors utilize two or more cameras to simulate human binocular vision or employ a single camera to capture images from different locations.

As deep learning can capture monocular cues such as object size, texture gradients, and perspective, depth estimation from a single image can produce accurate results. There have been attempts to use transformers, such as \cite{Zhang:Lite:2023} and \cite{Chen:Vision:2023}, and diffusion models, such as \cite{Ji:DDP:2023} and \cite{Ke:Repurposing:2024}. Amongst these, Depth Anything v2 \cite{Yang:depthanythingv2:2024} has become a state-of-the-art monocular depth estimation method. It is built on the previous version \cite{Yang:depthanything:2024}, jointly trained on large-scale labeled and unlabeled images and uses semantic priors from pretrained encoders. Depth Anything v2 significantly outperforms V1 in fine-grained details and robustness by using synthetic images and pseudo-labeled real images, as well as by extracting intermediate features from DINOv2 \cite{Oquab:DINOv2:2024}, which is trained with vision transformers. One of the notable capabilities of Depth Anything v2 is its ability to predict the depth of transparent and reflective surfaces.

% -------------------------------
\subsubsection{Neural Radiance Fields}
\label{sssec:nerf}

Neural Radiance Fields (NeRFs), introduced in \cite{Mildenhall:NeRF:2020}, have demonstrated the ability to learn a 3D scene from a smaller number of images captured from various viewpoints, as opposed to photogrammetry. They excel in neural rendering, particularly in view-dependent novel view synthesis, and have effectively tackled several challenges associated with automated 3D capture \cite{xie2022neural}, such as accurately representing the reflectance properties of the scene. NeRFs offer high-resolution photo-realistic novel views and flexibility in postprocessing. They have hence gained significant attention in cinematography \cite{Azzarelli2024}, as they offer reduced time and cost, particularly for outdoor shooting.

\begin{figure}
    \centering
    \includegraphics[width=\textwidth]{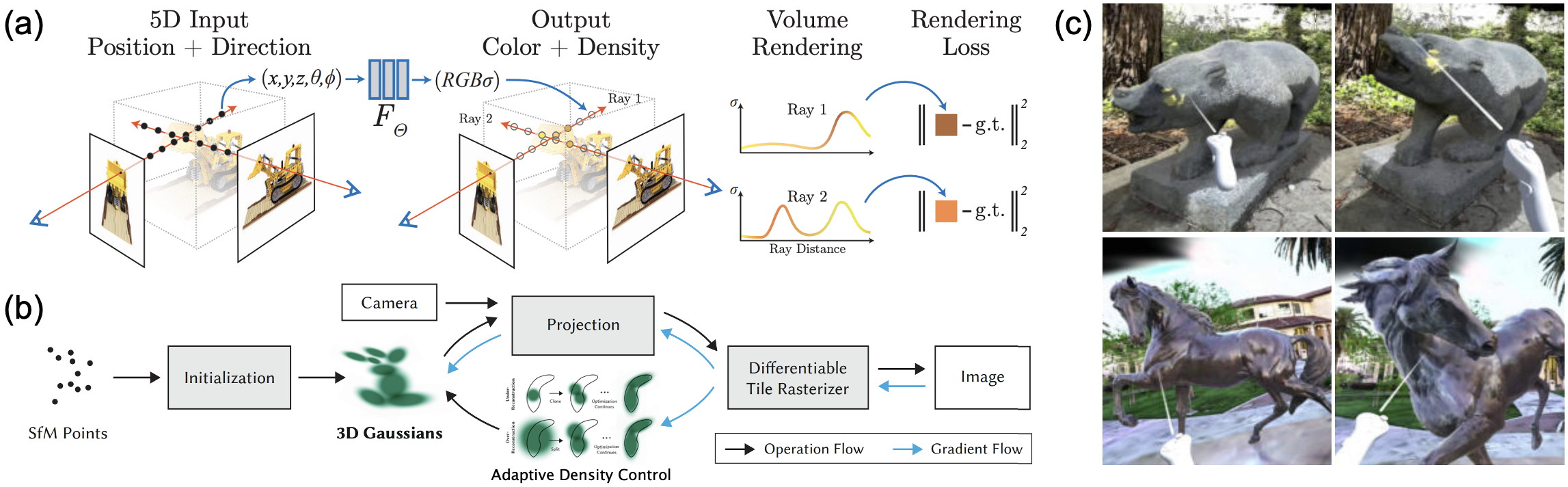}
    \caption{3D representation. (a) Neural Radiance Fields (NeRFs) \cite{Mildenhall:NeRF:2020}. (b) Gaussian Splatting \cite{kerbl:3Dgaussians:2023}. (c) Example scenes of VR-GS system for 3D content interaction in VR \cite{jiang:vrgs:2024}. }
    \label{fig:3Drepresentation}
\end{figure}

In the NeRF process (see Fig. \ref{fig:3Drepresentation} (a)), the camera positions and orientations are typically estimated from a series of 2D images using techniques like feature-mapping and Structure-from-Motion (SfM), as demonstrated in \cite{schoenberger:sfm:2016}. Leveraging INR, each image (or camera pose) is mapped into camera rays that traverse the scene, generating 3D points with directional radiance (towards the camera). These points are then processed by an MLP to predict volume density and emitted radiance. Subsequently, volume rendering techniques are employed to generate an image, which is compared with the original via loss calculation. The MLP iteratively refines the model by minimizing this loss.

Since their introduction, there have been many variants of NeRFs aimed at improving their performance. Mip-NeRF360 \cite{Barron:Mip-NeRF360:2022} proposed an unbounded anti-aliased technique achieving full 360 degree content. Google Research
\cite{Mildenhall:NeRFDark:2022} trains NeRF from noisy RAW images captured in the dark scene, allowing changing viewpoint, focus, exposure, and tone mapping simultaneously. With segmentation techniques significantly advanced (see Section \ref{sssec:seg}), there have been integrations utilizing semantic segmentation to enhance 3D representation \cite{Guo:neural:2022}. DSEM-NeR \cite{LIU:DSEM:2025} integrates the pretrained CLIP model to extract multimodal features—including color, depth, and semantics—from multi-view 2D images, thereby enhancing the reconstruction quality of complex scenes.

While the rendering quality of NeRF is very good, training and rendering times remain extremely high. The Instant-NGP tool developed by Nvidia \cite{mueller:instant:2022} enables real-time training of NeRFs by bypassing sampling in empty spaces and dense areas, and by incorporating multi-resolution hash encoding techniques. These advancements substantially reduce the computational burden associated with representing high-resolution image features -- training times have been reduced from hours to just a few seconds. Moreover, it offers VR controls for immersive 3D rendering experiences using OpenXR\footnote{An open-source, royalty-free standard for access to virtual reality and augmented reality platforms and devices. \url{https://www.khronos.org/openxr/}}. This allows users to navigate scenes, manipulate objects, and interact with the environment directly through VR headsets. Diffusion models are integrated to regularize NeRF reconstructions \cite{wynn:diffusionerf:2023}, resulting in smoother depth continuity and clearer edges where depth discontinuities occur.

The initial application of NeRFs to dynamic scenes was undertaken by Pumarola et al. \cite{pumarola:DNeRF:2020}, known as D-NeRF.
However, the current leading method for generating high-quality novel views of real dynamic scenes is TiNeuVox \cite{Fang:Fast:2022}.  It enhances temporal information by interpolating voxel features before feeding them into the radiance network to estimate density and color, similar to ordinary NeRF. DynVideo-E~\cite{Liu_2024_CVPR} adds an MLP to predict motion fields but focuses on human-centric content. PaReNeRF~\cite{Tang_2024_CVPR} addresses large-scale dynamic scenes using patch-based sampling. The main drawback of these methods is the large model size and/or long training time. Therefore, $K$-planes \cite{Fridovich:kplanes:2023} propose a simple planar factorization for volumetric rendering, achieving low memory usage (1000$\times$ compression over a full 4D grid). Wavelet transforms are employed in \cite{azzarelli:waveplanes:2023} to further reduce model size. KFD-NeRF~\cite{zhan2024kfd} incorporates a Kalman filter-guided deformation field for more accurate motion estimation.

% -------------------------------
\subsubsection{3D Gaussian Splatting}
\label{sssec:3DGS}

The main issue with NeRFs as a method to generate high-quality novel views is training time, which can exceed a day for high-resolution content on a single RTX 3090 GPU~\cite{9879447}. 3D Gaussian Splatting (3D-GS) \cite{kerbl:3Dgaussians:2023} has been introduced to address this, using anisotropic 3D Gaussians to form a high-quality, unstructured representation of radiance fields. The process estimates a sparse point cloud through SfM. Each point possesses 3D Gaussian properties, such as position, covariance matrix, opacity, and spherical harmonics coefficients representing colors. The optimization of these parameters is interleaved with steps that control the density of the Gaussians to better represent the scene, as shown in Fig. \ref{fig:3Drepresentation} (b). A survey of 3D-GS can be found in \cite{10521791}.

In contrast to traditional NeRFs based on implicit scene representations, 3D-GS provides an explicit representation that can be seamlessly integrated with post-processing manipulations, such as animating and editing. VR-GS \cite{jiang:vrgs:2024} offers intuitive and interactive physics-based game-play with deformable virtual objects and realistic environments represented with 3D-GS. The example scenes are shown in Fig.~\ref{fig:3Drepresentation} (c). Physics-inspired approaches are also integrated to improve 3D modeling in different media, such as 3D underwater scenes \cite{Wang2025}.

For dynamic scenes, 4D Gaussian Splatting (4D-GS) \cite{wu:4dgaussians:2024} introduces a Gaussian deformation field for motion and shape. It exploits a multi-resolution encoding method, achieving real-time rendering of up to 82 fps at a resolution of 800$\times$800 pixels on an RTX 3090 GPU. Instead of developing in 4D, CoGS \cite{Yu:CoGS:2024} exploits 3D-GS by integrating control mechanisms in separate regions to learn individual temporal dimensions. SC-GS \cite{Huang:SCGS:2024} extracts sparse control points and uses an MLP to predict time-varying 6 DoF transformations. While the results show better visual quality than 4D-GS and CoGS, the performance heavily relies on camera pose estimation. Kong et al. \cite{kong2025efficient} represent dynamic scenes using sparse, time-variant attribute modeling with a deformable MLP, while efficiently filtering out anchors corresponding to static regions. Their model achieves fast rendering speeds of over 110 FPS at a resolution of 960$\times$540—nearly 10 times faster than SC-GS—and delivers a 1 dB improvement in PSNR.

LUMA AI\footnote{\url{https://lumalabs.ai/interactive-scenes}} and 
Polycam\footnote{\url{https://poly.cam/captures}} offer free tools for Gaussian splatting and photogrammetry creation for non-commercial use. The 3D objects created can be experienced with VR headsets for more immersive 3D and further used or developed in other applications. However, these tools have limitations in handling dynamic scenes due to occlusions, sparse observations per timestamp, and object reappearances over time. Rendering dynamic avatars can produce higher quality results by incorporating additional information. For example, EVA \cite{junkawitsch2025eva} disentangles the 3D Gaussian appearance into skeletal motion, facial expressions, body movements, and skin. These components are then splatted to render the final photorealistic image.

% -------------------------------
\subsubsection{Digital Twins}
\label{sssec:digital_twins}

A digital twin is a virtual replica of a physical object, system, or process, continuously updated with real-time data for purposes such as simulation, testing, monitoring, and maintenance. This technology is increasingly adopted across various applications within the creative industries. For example, in product design and branding, it enables immediate observation of how a design performs in various contexts, facilitating the development of user-friendly products. Unilever reported that integrating digital product twins with 3D technologies, such as NVIDIA Omniverse\footnote{\url{https://www.nvidia.com/en-gb/omniverse/}}, enabled the creation of product imagery twice as fast and 50\% more cost-effective\footnote{\url{https://www.unilever.com/news/press-and-media/press-releases/2025/unilever-reinvents-product-shoots-with-digital-twins-and-ai/}}. Digital twins also allow consumers to explore products or spaces virtually, simulating real-world interactions. 

Accenture plc, a global professional services company, collaborated with Walt Disney Studios to develop digital twin technologies aimed at transforming the filmmaking process\footnote{\url{https://www.accenture.com/mx-es/case-studies/communications-media/empowering-film-creatives-digital-twins}}. Their goal is to generate remotely accessible 3D models, enabling virtual exploration of potential shooting locations without requiring physical visits. The Virtual StudioLAB provides a digital replica created using 360-degree imagery and 3D modeling. These innovations have streamlined pre-production workflows for major productions from Marvel Studios and 20th Century Studios.

Digital representations such as avatars, proxies, and digital twins are increasingly being explored in artistic contexts, particularly in relation to identity, presence, and embodiment in virtual environments. The Tate Modern’s film programme Avatars, Proxies and Digital Twins (Feb–May 2025) investigated these themes through curated audiovisual works, offering critical reflections on digital personhood. By engaging with diverse narrative forms, the programme highlighted the sociocultural implications of digital self-representation, prompting discourse on authenticity, agency, and the role of immersive media in shaping future human–machine interaction.

% ===================================================
\subsection{Data Compression}
\label{ssec:compression}

Data compression plays an important role in the delivery of creative content to audiences, effectively reducing memory and bandwidth requirements during signal storage and transmission \cite{Bull:intelligent:2021}. Although coding methods based on conventional signal processing theories are still widely employed in most standards and application scenarios, learning-based solutions have emerged in research, showing great potential to achieve competitive performance in recent years. This subsection provides a brief overview of the recent advances in image, video, and audio compression, in particular focusing on the approaches proposed after 2021.

\subsubsection{Image Compression}

Since the first neural image codec \cite{balle2016density} was proposed in 2016, numerous learning-based image compression methods have been developed, with significant performance improvements reported \cite{balle2018variational,cheng2020learned}. Driven by the latest advances in neural network architectures, neural image codecs now outperform standard image codecs. Instead of using CNNs as the basic network structure, transformer-based architectures have become popular, offering the potential for  better compression efficiency.  Notable examples include SwinT-ChARM \cite{zhu2022transformer}, STF \cite{zou2022devil} and LIC-TCM \cite{liu2023learned}. SwinT-ChARM \cite{zhu2022transformer} employs Swin transformers for non-linear transforms and outperforms the latest standard image codec, the Versatile Video Coding (VVC) Test Model (VTM, All Intra). STF \cite{zou2022devil} is based on a symmetrical transformer framework containing absolute transformer blocks in both the down-sampling encoder and the up-sampling decoder, which also shows improved rate-quality performance over VTM. LIC-TCM \cite{liu2023learned} exploits the local modeling ability of CNN and the non-local modeling performance of transformers, and proposes a parallel transformer-CNN mixture block. This new network structure, together with a channel-wise entropy model based on attention modules using Swin transformers, contributes to the superior performance of STF, with more than 10\% bitrate savings over VTM. 

An alternative approach to learned image coding is based on advanced generative models. Early works \cite{agustsson2019generative,mentzer2020high} employed GANs to generate more photo-realistic results with improved visual quality. Although these models fail to outperform conventional, CNN-based or transformer-based approaches, when distortion-based quality metrics, e.g., PSNR, are used for performance evaluation, they have been reported to perform well when perceptual quality models, such as MS-SSIM \cite{Bovik_MSSSIM} and VMAF \cite{VMAFblog}, or subjective tests are employed to measure perceived video quality. More recently, diffusion models have been applied in image compression to allow realistic reconstruction at ultra-low bitrates \cite{careil2023towards} achieving competitive performance compared to GAN-based models \cite{yang2024lossy}. However, it should be noted that some of these generative models aim to generate (or synthesize) images with ``perfect realism'' rather than reconstruct results that are most similar to the original content. Notable work in this category includes image codecs using score-based generative models \cite{hoogeboom2023high} and the diffusion-based residual augmentation codec (DIRAC) \cite{ghouse2023residual}. Moreover, another type of generative model based on INR has been employed for image compression; this learns a mapping between the spatial coordinates and the respective pixel values for the input image. The learned INR model is then compressed through parameter quantization and model compression to minimize the required bitrate. Notable INR-based image codecs include COIN/COIN++ \cite{dupont2021coin,dupontcoin++} and \cite{strumpler2022implicit} that combine SIREN networks \cite{sitzmann2020implicit} with positional encoding.   

In order to evaluate and compare neural image codecs under fair test conditions, public grand challenges have been increasingly run,  typically associated with international conferences. One of the most well-known of these is the Challenge on Learned Image Compression (CLIC) \cite{clic}. In its latest competition, the best performing learned image codec \cite{li2024semantic}, which is based on a GAN-enhanced Vector Quantized Variational AutoEncoder (VQ-VAE) framework, offered up to 0.6dB PSNR gain over VTM (version 22.2, All Intra) at similar bitrates; this codec is based on an autoencoder architecture with latent refinement and perceptual losses. 

To support the deployment of neural image codecs, the International Organization for Standardization (ISO)/International Electrotechnical Commission(IEC) has developed a royalty-free learned image coding standard, denoted as JPEG AI \cite{ascenso2023jpeg}, which aims to offer significant performance improvement over existing standards for both human and machine vision tasks. The Call for Proposals of JPEG AI was published in 2022, while the Working Draft and the Committee Draft outlining its core coding system were released in 2023 \cite{JPEGAIN100634}, with its first version published in October 2024 \cite{JPEGAIN100634}. JPEG AI follows the same framework (the auto-encoder structure) as most existing neural image codecs, and its test model JPEG AI VM (version 4.3) has been reported to achieve up to 28.5\% coding gains over VVC VTM (All Intra mode) \cite{JPEGAIM101081}.

\subsubsection{Video Compression}

Compared to image coding, the compression of video content is a much more challenging task, particularly for immersive video formats and diverse content types. Although video coding standards including H.264/AVC (Advanced Video Coding), H.265/HEVC (High Efficiency Video Coding) and H.266/VVC (Versatile Video Coding) are still predominant in real-world applications, learning-based video coding has advanced dramatically in the past five years, with new deep learning enhanced conventional coding tools and end-to-end optimized neural video coding frameworks proposed.    

i) \textbf{The enhancement of conventional coding tools} focuses on employing deep learning techniques to improve the performance of one or more coding modules in a standard-compliant codec. These modules include intra prediction \cite{li2021deepqtmt}, inter prediction \cite{jin2021deep}, in-loop filtering \cite{feng2024low}, post-decode filtering \cite{zhang2023wcdann} and resolution re-sampling \cite{wang2023compression}. To facilitate efficient integration, the MPEG Joint Video Experts Team (JVET)  built a test model in 2022 based on VTM 11, named Neural Network-based Video Coding (NNVC) \cite{li2023designs}, with its latest version NNVC-7.1 containing two major learning-based coding tools, neural-network based intra prediction and in-loop filtering, which has achieved an up to 13\% coding gain over VTM 11 (Random Access mode) \cite{JVET-AG0014}. However, this learning-based codec requires much higher computational complexity (up to 477 kMACs/pixel) and high-spec GPU support compared to conventional codecs. Meanwhile, members of the Alliance of Open Media (AOM) have also developed multiple CNN-based coding tools for the next generation of video coding standard beyond AV1. The latest proposals focus on the trade-off between performance and complexity, with one of them based on in-loop filtering and super-resolution, which achieves an average BD-rate saving of 3.9\% (in PSNR) over AVM, the test model of AV2, but only requires a much lower computational complexity (below 1.5kMACs/pixel) \cite{joshi2023switchable}. More recently, research has been conducted to further improve the performance of these learning-based coding tools utilizing more advanced network architectures, including ViTs \cite{kathariya2023joint}, and diffusion models \cite{li2024extreme}. There are also investigations on applying preprocessing before compression \cite{chadha2021deep,tan2024joint}, where the training of the deep preprocessors is based on proxy video codecs and/or rate-distortion loss functions to simulate the behavior of conventional video coding algorithms.

ii) \textbf{End-to-end optimized neural video codecs.} Alongside the enhancement of coding tools in conventional video codecs, more recent research activities have focused on using neural networks to implement the whole coding workflow, enabling data-driven end-to-end optimization. The performance of these neural video codecs has advanced significantly in the last five years, since the first attempt, DVC \cite{lu2019dvc}, was published. DVC  matched the performance of a fast implementation of H.264 (x264). However currently, learned video coding algorithms (e.g., DCVC-FM \cite{li2024neural} and DCVC-LCG \cite{Qi2024longterm}) are able to compete or even outperform the state-of-the-art standard codecs, such as VVC VTM under certain coding configurations. These learning-based methods often focus on enhancement from different perspectives, including feature space conditional coding (e.g., FVC~\cite{hu2021fvc} and DCVC \cite{li2021deep}), instance adaptation ~\cite{khani2021efficient,yang2024parameter}, and motion estimation (e.g., DCVC-DC~\cite{li2023neural}). New architectures have also been proposed such as CANF-VC~\cite{ho2022canf} based on a video generative model, MTMT \cite{xiang2022mimt} using a masked image modeling transformer-based entropy model and VCT \cite{mentzer2022vct} based on a video compression transformer. It is noted that, although promising coding performance has been achieved by the aforementioned contributions, these neural video codecs (in particular those based on autoencoder backbones) are typically associated with high computational complexity (especially in the decoder), which constrains their deployment for practical applications. To address this issue, researchers are now focused on complexity reduction while maintaining coding performance through model pruning and knowledge distillation~\cite{guo2023evc,peng2024accelerating}. 

It should be noted that the neural codecs mentioned above are typically trained offline with diverse video content \cite{nawala2024bvi}, and deployed online for inference. In this case, model generalization becomes important, and this is why these codecs often have a large model capacity, resulting in large model sizes and slow inference runtime. Inspired by recent advances in implicit neural representations (INR), a new type of video codec has emerged that employs INR models to ``represent'' the video  by learning a coordinate-based mapping and compressing the network parameters for transmission. This approach converts a video coding problem into a model compression task, which allows the use of a much smaller network to ``overfit'' the input video, with the real potential for fast decoding. Existing implicit neural video representation (NeRV) models can be classified into index-based and content-based methods. The former takes frame~\cite{chen2021nerv}, patch~\cite{bai2023ps} or disentangled spatial/grid coordinates~\cite{li2022nerv} as model input, while content-based approaches~\cite{kwan2024hinerv,kim2024c3,leguay2024cool} have content-specific embedding as inputs. Currently, one of the best INR-based video codecs \cite{kwan2024nvrc} has already achieved a performance similar to that of VVC VTM (RA), but with a much lower decoding complexity compared to autoencoder-based neural codecs. Some of these models have also been applied to volumetric video content~\cite{ruan2024point,kwan2024immersive}, demonstrating their potential to compete with standard and other learning-based methods. However, it should be noted that the training of most NeRV models is based on an entire video sequence or even datasets; this results in a high system delay and does not meet the requirement of many low latency video streaming or real-time applications. To address this limitation, significant advances  have been made~\cite{gao2024pnvc} towards more practical INR-based video compression (such as the Low Delay and Random Access modes in VVC VTM \cite{bossen2023vtmctc}) by combining pre-training and online model overfitting.

% \begin{figure}
%         \centering
%         \includegraphics[width=\linewidth]{figures/codec_comparison.pdf}
%         \caption{Radar plots illustrating the performance of conventional and
% neural video codecs, in terms of coding efficiency (BD-rate
% measured by PSNR and MS-SSIM on UVG ~\cite{mercat2020uvg} and MCL-JVC ~\cite{wang2016mcl}
% datasets, against HM 18.0, Low-Delay), relative decoding speeds (FPS) and coding latency. The results are from \cite{gao2024pnvc}.}
%         \label{fig:codec-comparison}
%     \end{figure}
    
Similarly to image compression, international grand challenges are used to compare neural video compression methods, with notable venues including the NN-based Video Coding Grand Challenge associated with The IEEE International Symposium on Circuits and Systems (ISCAS) and the Challenge on Learned Image Compression (CLIC, video coding track) with IEEE/CVF CVPR and Data Compression Conference (in 2024). The best performer in ISCAS 2024 NN-based Video Coding Grand Challenge offers an overall 55\% BD-rate saving over HEVC Test Model HM 
 \cite{iscas2024}, while the winner of the CLIC (video coding track) in 2024, a neural-network enhanced ECM codec \cite{zhao2024neural} with a CNN-based in-loop filter, shows a more than 2dB (in PSNR) gain compared to VTM (RA) at the same bitrates. 

\subsubsection{Audio Compression}

Similarly to images and videos, learning-based solutions have also been researched to compress audio signals, and most neural audio codecs are based on VQ-VAE \cite{Oord:Neural:2017}. SoundStream \cite{zeghidour2021soundstream} is one such model, which can encode audio content at various bitrates. It is based on a residual vector quantizer (RVQ) which trades off rate, distortion and complexity. This work has been further enhanced with a multi-scale spectrogram adversary and a loss balancer mechanism, resulting in improved rate-distortion performance. A more advanced universal model has been further developed \cite{kumar2024high} based on improved adversarial and reconstruction losses, which can compress different types of audio. RVQ has also been extended from a single scale to multiple scales \cite{siuzdak2024snac}, which performs hierarchical quantization at variable frame rates. 

More recently, researchers have started to exploit the use of LLMs for audio compression, leveraging the audio generation/synthesis abilities of generative models. UniAudio 1.5 \cite{yang2024uniaudio} is one of such attempts, which converts an audio into the textural space, which can be represented by a pre-trained LLM that shares a similar backbone of UniAudio \cite{yang2023uniaudio}, a universal audio foundation model. LFSC is another neural audio codec based on LLMs, which achieved fast LLM training and inference through finite scalar quantization and adversarial training. 

% ===================================================
\subsection{Visual Quality Assessment}
\label{ssec:asssessment}

Assessing the quality of visual signals remains an important and challenging task for many image and video processing applications. While subjective tests involving human participants remain the gold standard, objective quality models are frequently used  because of their time and cost efficiency. These quality assessment methods are typically used to evaluate the performance of different visual processing approaches, and they can also be converted to loss functions, which are employed for optimizing learning-based processing models.  

In recent years, quality assessment methods have  been enhanced using deep learning techniques. The resulting learning-based quality models can quickly adapt to a specific type of content, leading to better performance compared to conventional, hand-crafted quality metrics. This section provides a brief summary of existing work in this research area, and highlights the main challenges which should be addressed in the near future. A more comprehensive overview of the image and video quality assessment literature can be found in \cite{zhai2020perceptual,zheng2024video,zhang2024quality}.

\subsubsection{Quality assessment models}

Image and video quality assessment methods can be classified into two primary categories according to the availability of the corresponding reference (un processed) content. These are referred to as full-reference and no-reference models\footnote{Reduced-reference quality metrics do exist in the literature, but research in this field has been less active in recent years.}. Prior to the AI era, conventional visual quality methods often exploited characteristics of the human vision system capturing information related to structural similarity (such as in SSIM and its variants \cite{Bovik_SSIM,wang2003multiscale,rehman2015display}), distortion \cite{chandler2007vsnr,larson2010most,STMAD}, and artifacts \cite{ou2010perceptual,zhu2014no,zhang2015perception}. In many cases, the extracted features are further processed by models that simulate texture masking \cite{helmholtz1896handbook}, contrast sensitivity \cite{kelly1977visual}, and saliency \cite{itti2001computational}. These hand-crafted quality models have also been combined with features within a regression-based framework in order to achieve more accurate prediction performance - VMAF is one such example \cite{VMAFblog}. When neural networks are used for feature extraction, they are trained to capture information which can directly contribute to quality prediction through an end-to-end optimization strategy. Initially, convolutional neural networks were used for this, with notable examples such as DeepQA \cite{kim2017deep}, LPIPS \cite{zhang2018unreasonable} and CONTRIQUE \cite{madhusudana2022image} for image quality assessment, and TLVQA \cite{korhonen2019two}, C3DVQA \cite{xu2020c3dvqa} and DeepVQA \cite{kim2018deep} for video quality assessment. Recent works have been reported to achieve better performance when Vision Transformers (ViTs) (or similar variants) are employed due to the effectiveness of their self-attention mechanism. Important works in this class include IQT \cite{cheon2021perceptual}, TRes \cite{golestaneh2022no}, SaTQA \cite{shi2024transformer}, FastVQA \cite{wu2022fast} and RankDVQA \cite{feng2024rankdvqa}. The former has been further extended as DOVER \cite{wu2023exploringvideo} and COVER \cite{he2024cover} when aesthetic and/or semantic aspects in the content are taken into account. 

More recently, inspired by the success of large language models (LLMs) \cite{openai:gpt4:2023,touvron2023llama} in other machine learning tasks, these have been utilized in image and video quality assessment,  demonstrating significant potential to achieve better model generalization. Q-Bench \cite{wu2024qbench} is one of the first attempts that employs multimodal large language models to predict the perceptual quality of images based on prompt-driven evaluation. It queries the LLMs to provide information related to the final quality rating of the input image and the quality description. This has been further extended for video quality assessment tasks in Q-Align \cite{wu2024qalign}. Other notable works include X-iqe \cite{chen2023x} that performs the quality prompt in a multi-iteration manner focusing on both image fidelity and aesthetics. Prompt-based approaches have also been proposed for differentiating the quality difference between multiple images, such as 2AFC-LMMs \cite{zhu20242afc} based on a two-alternative forced choice prompt and MAP (maximum a posteriori) estimation. Moreover, recent research works also focus on using pre-trained vision-language models, such as CLIP \cite{radford2021learning}, which align better image and text modalities. Important examples in this class for image quality assessment include ZEN-IQA \cite{miyata2024zen}, QA-CLIP \cite{pan2023quality} and PromptIQA \cite{chen2025promptiqa}. Similar works have also been proposed for video quality assessment, such as BVQI \cite{wu2023exploring,wu2023towards} and COVER \cite{he2024cover}.

% \subsubsection{Databases and training methodology}

To support the training and validation of learning-based quality assessment models, image or video databases containing ground-truth subjective quality scores are typically employed. Commonly used image quality databases include LIVE \cite{sheikh2006astatistical}, CSIQ \cite{larson2010most}, TID2013 \cite{ponomarenko2013color}, PieAPP and PIPAL, while video quality databases such as LIVE-VQA~\cite{seshadrinathan2010study}, KoNViD-1K~\cite{hosu2017konstanz}, YouTube UGC~\cite{wang2019youtube} and LIVE-VQC~\cite{sinno2018large} are typically employed for benchmarking in the literature. There are also databases developed that investigate the impact of specific video formats and/or artifacts, such as LIVE-YT-HFR  \cite{madhusudana2021subjective} focusing on frame rates, VSR-QAD \cite{zhou2024database} on spatial resolution (or super-resolution artifacts), BAND-2k \cite{chen2024band2k} on banding artifacts and Maxwell \cite{wu2023towards}/BVI-Artifact \cite{feng2024bvi} containing multiple artifacts commonly produced in video streaming. Based on these databases, many learning-based quality assessment models are trained to minimize the difference (L1 or L2 norm) between predicted quality indices and subjective scores. However, due to the limited number of ground-truth quality labels associated with these databases and the resource requirements associated with  collecting subjective data using human participants in psychophysical experiments, this type of training methodology cannot offer satisfactory performance, in particular when the model capacity is large. Moreover, since the experimental settings and conditions used for quality labeling are different in these databases, intra-database cross-validation is always required due to the limited model generalization and potential overfitting problems. 

To address these issues, various proxy quality metrics have been used to label images and videos, which avoid expensive subjective tests and enable the generation of a large amounts of training material with pseudo-ground-truth quality annotations. To further improve the reliability of quality labels, instead of learning the absolute values of the quality labels, ranking-inspired training strategies have been developed, which focus on improving the monotonicity characteristics of quality. Important examples based on these weakly supervised training methodologies include RankIQA \cite{liu2017rankiqa} and UNIQUE \cite{zhang2021uncertainty} for the image quality assessment task, and VFIPS \cite{hou2022perceptual} and RankDVQA \cite{feng2024rankdvqa} for video quality assessment. Moreover, different self-supervised learning approaches have also been employed, which transform quality labeling to an auxiliary task. For example, CONTRIQUE \cite{madhusudana2022image} learns relevant features from an unannotated image database based on the prediction of distortion types and degrees through contrastive learning. This method has been further applied to video quality assessment, resulting in a contrastive video quality estimator, CONVIQT \cite{madhusudana2023conviqt}. More recently, quality-aware contrastive loss has been designed in \cite{zhao2023quality,peng2024rmt} to stabilize the learning process.

\subsubsection{Performance and main challenges}

Due to the lack of standard test conditions and limited model generalization within many existing image and video quality assessment models, deep compression methods are typically trained and benchmarked using different databases in conjunction with intra-database cross-validation. This can result in inconsistent evaluation results and conclusions. To enable a fair and meaningful comparison, various challenges and contests have been held for visual quality assessment. The Sixth Challenge on Learned Image Compression (CLIC) \cite{clic} associated with the Data Compression Conference 2024 is one of the latest examples which includes two quality assessment tracks for image and video compression. The best performer in the video quality assessment track achieves a Spearman Ranking Correlation Coefficient value of 0.825 \cite{feng2024rankdvqa}, which is based on a ranking-inspired training methodology. Other notable challenges include the IEEE/CVF WACV 2023 HDR VQA Grand Challenge and the Video Super-Resolution Quality Assessment Challenge in ECCV 2024, which focus on high dynamic range and super-resolved content, respectively. 

Although significant progress has been made in the past few years in visual quality assessment, including new models and training methodologies, challenges remain, including limited model generalization and high computational complexity. Another important use of quality metrics is as embedded loss functions for image and video processing optimization. This requires further capability and robustness, alongside complexity reduction, all topics to be addressed in future work.

% ===================================================
{
\subsection{Summary of AI technologies for creative industries}
\label{ssec:summary}

This section consolidates the preceding discussion by providing a comparative overview of the main classes of AI models shaping contemporary creative practice. While earlier sections examined individual technologies in detail, the following summary highlights how these models—ranging from large language models (LLMs) and diffusion models (DMs) to Neural Radiance Fields (NeRFs) and Implicit Neural Representations (INRs)—differ in application domains, key advantages, and persistent limitations. This synthesis enables readers from both creative and technical backgrounds to discern where each model type contributes most effectively, where challenges remain, and how these approaches collectively reshape workflows across the creative industries. Table \ref{tab:tech_comparison} summarizes these relationships and serves as a conceptual reference for future research, mapping core model classes to their applications, strengths, and constraints to guide the evaluation and development of emerging AI methods in the creative industries.
}

\begin{table*}[t]
\centering
\caption{ Comparative summary of key AI technologies for creative industries}
\label{tab:tech_comparison}
\footnotesize
\begin{tabular}{>{\raggedright\arraybackslash}p{2cm}|>{\raggedright\arraybackslash}p{1.8cm}|>{\raggedright\arraybackslash}p{2.6cm}|>{\raggedright\arraybackslash}p{2.8cm}|>{\raggedright\arraybackslash}p{2.6cm}|>{\raggedright\arraybackslash}p{1.6cm}}
\toprule
\textbf{Technology} & \textbf{Core Mechanism} & \textbf{Creative Applications} & \textbf{Key Strengths} & \textbf{Main Limitations} & \textbf{Accessibility} \\
\midrule

\textbf{Transformers/ Attention} & 
Self-attention mechanisms capture global dependencies & 
Image/video restoration, super-resolution, segmentation, object detection, tracking & 
Parallel processing, long-range context, scalable, state-of-the-art performance & 
High memory usage, quadratic complexity, requires large training datasets & 
High: SwinIR, SAM, Restormer (open-source) \\
\hline

\textbf{Large Language Models (LLMs)} & 
Language understanding and generation & 
Text generation, dialogue systems, screenwriting, storyboarding, code assistance, script analysis & 
Highly flexible across domains; capable of reasoning and context understanding; supports creative ideation and natural language interaction & 
Hallucinations and factual errors; limited interpretability; potential bias from training data; copyright and authorship ambiguity & 
Moderate: GPT-4, Claude (APIs); LLaMA, Qwen (open weights) \\
\hline

\textbf{Diffusion Models (DMs)} & 
Iterative denoising from Gaussian noise & 
Image synthesis, Text-to-image/video generation, animation, VFX, design prototyping, restoration, style transfer, inpainting, restoration & 
High-quality diverse outputs; stable training; handles complex distributions; photorealistic; fine-grained control through prompts or conditioning inputs & 
Computationally expensive; slow sampling; temporal flicker in videos; difficulty ensuring semantic or style consistency & 
High: Stable Diffusion (open); DALL·E~3, Sora (APIs) \\
\hline

\textbf{Neural Radiance Fields (NeRFs)} & 
Volumetric scene representation via MLPs & 
3D scene reconstruction, virtual production, spatial storytelling, AR/VR content creation & 
Photorealistic rendering from sparse views; compact scene representation; supports dynamic viewpoint changes & 
Limited to static or semi-static scenes; slow training; poor performance in textureless regions; large memory footprint & 
Moderate: Instant-NGP, Mip-NeRF (open); requires CV expertise \\
\hline

\textbf{Implicit Neural Representations (INRs)} & 
Coordinate-based continuous functions & 
Video/image compression; super-resolution; 3D reconstruction; dynamic scene modeling & 
Continuous scene representation; parameter-efficient encoding; smooth interpolation and compact storage & 
Challenging to generalize across scenes; training instability; limited semantic control & 
Moderate: NeRV, COIN (open); technical expertise required \\
\hline

\textbf{Gaussian Splatting} & 
Explicit 3D Gaussians for scene representation & 
Real-time novel view synthesis; 3D scene reconstruction; VR/AR;interactive rendering & 
Fast training (minutes) and rendering (real-time); explicit representation; easy manipulation & 
Lower quality than NeRF for complex scenes, requires SfM initialization, memory intensive & 
High: 3D-GS, 4D-GS (open); LUMA~AI, Polycam (free tools) \\
\bottomrule

\end{tabular}
{\color{blue}* Product and platform data (e.g., Sora, Gemini, Stable Diffusion 3) are accurate as of mid-2025 and may evolve rapidly.}
\end{table*}

% =======================================
\section{Closing Thoughts: The Future of AI in Creative Applications}
%Concluding Remarks and Future of AI}
\label{sec:discussion}

This paper has presented a comprehensive review of current AI technologies and their creative industries applications that have emerged in recent years. Generative methods have driven a rapid growth in AI usage, particularly in the creative sector, significantly advancing the state of the art across various applications such as content creation, information extraction and analysis, content enhancement and data compression.

Through these applications, generative AI has not only broadened creative possibilities, but has also reduced the manual effort and time traditionally associated with the production pipeline, allowing for greater creative experimentation and more rapid and agile production cycles. As this technology advances, it promises to unlock even more sophisticated capabilities. However, creative technologists, artists and other users must adapt,  learn to use, and build these tools effectively and safely.

\subsection{Challenges for AI in the Creative Sector}

{\color{red} Artists are already exploring how to bridge structured nature of current creative AI and more traditional (analog) workflows. This includes using multiple methods, models, or tools to create new works. For example, artists can use tools to iterate on existing or past works; upload and fuse analog works or output from different Generative AI tools to further intervene in the AI generation process; and composite outputs to reconstruct or fill in missing or damaged parts of a work. However,} one of the primary challenges for artists engaging with modern generative AI and LLMs is the lack of consistent, controllable outputs. These models operate via stochastic sampling from high-dimensional latent spaces, meaning that identical prompts can yield different results across runs. This unpredictability can make it challenging for artists to achieve, or iterate toward, a precise creative vision. Although prompt engineering has emerged as a technique to guide model behavior, it requires technical knowledge and iterative refinement, which may not align with the intuitive or exploratory approaches common in artistic practice. 

Moreover, there can be a fundamental tension between the structured nature of current AI pipelines and current production, often improvisational, workflows used in creative disciplines. Many generative tools were originally designed for tasks like software development, content automation, or optimization \cite{zhong:LDB:2024}, and are less well suited for open-ended, exploratory creation. Artists typically work in cycles of ideation, experimentation, and revision—processes that demand fluid, real-time interaction and control, which existing AI systems struggle to support. These limitations point to a gap in current AI design: a need for systems that not only generate high-quality content but also adapt to the iterative, interpretive nature of artistic production. One possible approach to address these challenges is a reinforcement  of top-down creative workflows, where artists define high-level concepts, themes, or goals via text prompts before refining specific outputs. This approach helps align AI-generated results with artistic intent, offering a degree of control over inherently stochastic systems.

{ A further issue concerns creative authorship and ownership. As AI systems increasingly contribute to the ideation and execution of creative work, the line between human and machine authorship becomes blurred. Determining where creative credit lies—whether with the prompt designer, the model developer, or the AI system itself—poses significant legal and ethical challenges. Moreover, generative systems often reproduce stylistic elements from training data, raising questions about originality and cultural appropriation in AI-assisted creation.}

Speaking at the World Government Summit in Dubai in 2024,\footnote{\url{https://blogs.nvidia.com/blog/world-governments-summit/}} NVIDIA CEO Jensen Huang argued that, with rapid advancements in AI, learning to code may become less essential for newcomers to the tech sector. He envisioned a future where traditional programming could be replaced by more intuitive AI-driven tools, thereby automating complex tasks and enhancing productivity—particularly for artists without coding expertise. While this perspective remains debated, it highlights the potential for AI to become more accessible within creative fields, not just in coding but across areas such as VFX and virtual production. But to achieve this, AI-assisted coding tools must be better integrated into creative workflows. Creators must exploit techniques such as fine-tuning pre-trained models, few-shot learning, or domain adaptation—methods that are powerful yet typically inaccessible without machine learning expertise.

There are also broader concerns that persist regarding the long-term impact of AI on the creative industries, { economies, and labor markets. As automation accelerates, socioeconomic disparities may widen between those who can afford access to powerful generative systems and those who cannot. Freelancers and smaller studios risk being marginalized by large organizations with access to proprietary datasets and substantial compute resources.} The potential emergence of artificial general intelligence (AGI). Envisioned by organizations like OpenAI, DeepMind, and Anthropic, AGI could surpass human cognitive abilities, raising ethical and existential questions about the role of human agency in artistic expression. { Ensuring equitable access to AI tools and fair distribution of creative value will therefore be crucial to sustaining diversity, inclusion, and innovation within the sector.}

\subsection{Ethical Issues, Fakes and Bias}

%\url{https://www.artnews.com/art-news/news/new-data-poisoning-tool-enables-artists-to-fight-back-against-image-generating-ai-companies-1234684663/}

Advancements in generative AI, exemplified by models like Sora and Gemini 1.5 Pro, provoke ethical concerns and have societal implications. While their applications, with appropriate permission, can be beneficial and entertaining, these models, because they are capable of generating highly realistic content, escalate the risk of misuse through malicious deepfakes and misinformation. We are now in a situation where AI results transcend the uncanny valley, further complicating matters and challenging perceptions of authenticity. For example, the artist Miles Astray demonstrated that even authentic photographs could be mistaken for AI-generated images. His real photograph `F L A M I N G O N E' won both the jury’s award and the people’s choice award in the AI category of the 1839 Awards. His aim was to highlight the ethical dilemmas inherent in AI, suggesting that the benefits of discussing AI's ethical implications could surpass the ethical concerns related to viewer deception\footnote{\url{https://www.milesastray.com/news/newsflash-reclaiming-the-brain}}.

While democratizing AI tools no doubt presents opportunities to transform creative processes and workflows, it also necessitates robust regulatory frameworks to safeguard privacy and ownership. For example, deepfake technologies stimulate  significant concerns about the spread of misinformation and other malicious uses. Efforts to detect and identify increasingly realistic deepfakes are thus as important as the generative methods used to produce them. These must however be accompanied by increased media literacy, and policies that address the ethical and legal implications.

Diversity and representation is a key issue when using AI tools. Unified Concept Editing \cite{Gandikota:Unified:2024} has been proposed as a basis for image generation in digital mediums. This aims to ensure the production of safe content with diverse representation, reducing gender and racial biases. Hallucination in generative AI (the production of outputs that are not faithful representations of reality but instead contain imagined or unrealistic elements) is a further cause of concern. These undermine trust in AI processes and can be due to limitations in the training data, biases in the model architecture or imperfections in the optimization process. Hallucinations associated with LLMs are one of the issues highlighted by the UK Government  \cite{UK:Large:2024}, alongside bias, regurgitation of private data, difficulties with multi-step tasks and challenges in interpreting black-box processes.

Governments across the world  are increasingly expressing concerns about the challenges and uncertainties that generative AI technologies pose to rights holders and human creativity \cite{Jeary2024}. Generative AI presents substantial legal challenges, including the copyright status of AI-generated work and the intellectual property and copyright implications of the datasets used in training AI models. Viewpoints on this issue do however differ. For example, the track ``Heart on My Sleeve," penned by an (as yet unidentified) human author, featured AI-generated vocals that replicated the voices of Drake and The Weeknd. Released independently on April 4, 2023, it was accessible via streaming platforms including Apple Music, Spotify, and YouTube. The song quickly became viral, accumulating over 20 million views across all platforms\footnote{\url{https://www.nbcnews.com/pop-culture/viral-ai-powered-drake-weeknd-song-removed-streaming-services-rcna80098}}, prior to its removal by Universal Music Group, Drake's recording label. In contrast, Canadian artist Grimes has extended an invitation to musicians to emulate her voice via AI for the creation of new musical pieces, stipulating that the lyrics should not be harmful. She has advocated for the democratization of art and the abolition of copyright\footnote{\url{https://www.bbc.com/news/entertainment-arts-65385382}}. Additionally, Grimes has employed AI to design visual content for her LED backdrop at Coachella in 2024.

Finally, the rapid development of AI technologies has also raised concerns about job displacement and the balance between automation and human participation in creative processes. Ensuring that AI augments, rather than undermines, human effort poses a significant challenge for developers and policymakers.

%\subsection{The human in the Loop -- AI and Creativity}
% ============================================================
\subsection{The future of AI technologies}

Several key technological issues remain which need to be addressed if AI is to deliver its full potential. These in particular relate to training data, computational complexity and their depth of reasoning or planning, and are discussed below.

A substantial amount of data is essential for training AI models in order to achieve high performance and good generalization. Major companies such as Google, Meta, and NVIDIA, with their respective models: BERT, Segment Anything, and Canvas, dominate this space, benefiting from leveraged resources to gather data and process it to train sophisticated models. However,  in November 2024, Bloomberg reported that OpenAI, Anthropic, and Google are all experiencing relatively slow growth in the performance of their AI models, with one of the key challenges being training data\footnote{\url{https://www.bloomberg.com/news/articles/2024-11-13/openai-google-and-anthropic-are-struggling-to-build-more-advanced-ai}}.

LLMs excel in applications involving complex tasks, advanced reasoning, data analysis, and understanding context. 
However, these models typically require high computational resources or cloud computing for development, operation and fine-tuning. A new trend emerging alongside LLMs is the development of Small language models (SLMs), such as Phi-3 by Microsoft\footnote{\url{https://azure.microsoft.com/en-us/blog/introducing-phi-3-redefining-whats-possible-with-slms/}}. SLMs offer promising solutions for regulated industries and sectors encountering scenarios where high-quality results are essential while keeping data 'on-site'. Their potential is particularly relevant when deploying more capable SLMs on smartphones and other mobile devices, allowing them to operate `at the edge' without relying on cloud connectivity. Recent highly successful platforms, such as  DeepSeek-V3 \cite{deepseekv3} and Qwen2.5-Max \cite{qwen25}, are based on Mixture-of-Experts (MoE) models, which tackle complex problems by dividing them into simpler sub-tasks, each handled by a specialized ``expert."

Despite evident advancements in AI, current models still struggle with tasks requiring planning or deep reasoning and are prone to errors when encountering unexpected data. This, in turn, reduces the confidence of users and trust in the results.  AI algorithms can learn through reinforcement learning, but this process often identifies the best outcome as an anomaly rather than the norm. Yann LeCun, Professor at NYU and Chief AI Scientist at Meta, noted that while LLMs show a degree of comprehension in processing and generating text, their understanding lacks depth, often leading to results that defy common sense\footnote{\url{https://twitter.com/ylecun/status/1728496457601183865}}. He advocates for self-supervised learning as a pivotal future direction for AI, emphasizing its potential to derive insights from unlabeled data. Concurrently, Andrew Ng, Adjunct Professor at Stanford University and Founder of DeepLearning.AI, sees iterative AI agentic workflows\footnote{\url{https://www.youtube.com/watch?v=sal78ACtGTc}} as a key advancement for enhancing AI tool capabilities through an interactive approach by AI agents. These workflows involve autonomous agents that interactively learn from experience, understand natural language, and execute tasks on behalf of users.

The increasing openness of code and datasets is seen by many as a catalyst for accelerating AI advancements, with major firms like Microsoft, Google, and Meta supporting open access technologies. However, this openness also introduces security risks, necessitating new regulatory measures to monitor models post-release, to standardize documentation, and to assess the safety of  software code and training data disclosure.

Finally, as stated in \cite{Jeary2024}, the rapid advancement of AI technologies has revolutionized cultural experiences, often referred to as `CreaTech'—the convergence of the creative and digital sectors \cite{CreativeIndustriesCouncil2021}. Such innovations not only reshape how people engage with art and creative work (e.g., through AR/VR/MR) but also drive the evolution of the technologies themselves.

%%===========================================================================================%%

\bmhead{Research funding}

This work has been funded by the UKRI MyWorld Strength in Places Programme (SIPF00006/1).

\bmhead{Data Availability}
No datasets were generated or analysed during the current study.

\bmhead{Author Contributions Statement}
N.A. wrote the main manuscript text in section 1, 2,  3.1-3.5,3.8, 4, prepared all figures.
F.Z. wrote the main manuscript text in section 3.6-3.7.
D.B. wrote the main manuscript text in section 4.
All authors reviewed the manuscript.

\bibliography{literature_review}% common bib file

@inproceedings{zhong:LDB:2024,
  title={{LDB: A} Large Language Model Debugger via Verifying Runtime Execution Step-by-step},
  author={Li Zhong and Zilong Wang and Jingbo Shang},
  booktitle={Proceedings of the 62nd Annual Meeting of the Association for Computational Linguistics: Findings.},
  year={2024}
}

@InProceedings{Gong:Continuous:2023,
    author    = {Gong, Rui and Wang, Qin and Danelljan, Martin and Dai, Dengxin and Van Gool, Luc},
    title     = {Continuous Pseudo-Label Rectified Domain Adaptive Semantic Segmentation With Implicit Neural Representations},
    booktitle = {Proceedings of the IEEE/CVF Conference on Computer Vision and Pattern Recognition (CVPR)},
    month     = {June},
    year      = {2023},
    pages     = {7225-7235}
}

@article{Yang2025,
  author    = {Jing Yang},
  title     = {Exploring the Productivity of Generative AI-Powered Ad Campaigns: A Consumer Response Perspective},
  journal   = {Journal of Interactive Advertising},
  year      = {2025},
  volume    = {25},
  number    = {3},
  pages     = {222--239},
  doi       = {10.1080/15252019.2025.2551598}
}

@inproceedings{Fang:Fast:2022,
author = {Fang, Jiemin and Yi, Taoran and Wang, Xinggang and Xie, Lingxi and Zhang, Xiaopeng and Liu, Wenyu and Nie\ss{}ner, Matthias and Tian, Qi},
title = {Fast Dynamic Radiance Fields with Time-Aware Neural Voxels},
year = {2022},
note = {https://doi.org/10.1145/3550469.3555383},
doi = {10.1145/3550469.3555383},
booktitle = {SIGGRAPH Asia 2022 Conference Papers},
articleno = {11},
numpages = {9}
}

@ARTICLE{Stefanini:From:2023,
  author={Stefanini, Matteo and Cornia, Marcella and Baraldi, Lorenzo and Cascianelli, Silvia and Fiameni, Giuseppe and Cucchiara, Rita},
  journal={IEEE Transactions on Pattern Analysis and Machine Intelligence}, 
  title={From Show to Tell: A Survey on Deep Learning-Based Image Captioning}, 
  year={2023},
  volume={45},
  number={1},
  pages={539-559},
  doi={10.1109/TPAMI.2022.3148210}}

@ARTICLE{Zhang:vision:2024,
  author={Zhang, Jingyi and Huang, Jiaxing and Jin, Sheng and Lu, Shijian},
  journal={IEEE Transactions on Pattern Analysis and Machine Intelligence}, 
  title={Vision-Language Models for Vision Tasks: A Survey}, 
  year={2024},
  volume={},
  number={},
  pages={1-20},
  doi={10.1109/TPAMI.2024.3369699}}

@InProceedings{Mildenhall:NeRFDark:2022,
    author    = {Mildenhall, Ben and Hedman, Peter and Martin-Brualla, Ricardo and Srinivasan, Pratul P. and Barron, Jonathan T.},
    title     = {{NeRF in the Dark: High} Dynamic Range View Synthesis From Noisy Raw Images},
    booktitle = {Proceedings of the IEEE/CVF Conference on Computer Vision and Pattern Recognition (CVPR)},
    month     = {June},
    year      = {2022},
    pages     = {16190-16199}
}

@inproceedings{pumarola:DNeRF:2020,
    title={{D-NeRF: Neural} Radiance Fields for Dynamic Scenes},
    author={Pumarola, Albert and Corona, Enric and Pons-Moll, Gerard and Moreno-Noguer, Francesc},
    booktitle={Proceedings of the IEEE/CVF Conference on Computer Vision and Pattern Recognition},
    year={2020}
}

@article{Anantrasirichai:AI:2022,
  title={Artificial intelligence in the creative industries: a review},
  author={Anantrasirichai, N. and Bull, D.},
  journal={Artificial Intelligence Review},
  volume={55},
  pages={589--656},
  year={2022},
  publisher={Springer},
  doi={10.1007/s10462-021-10039-7}
}

@inproceedings{Mirowski:cowriting:2023,
author = {Mirowski, Piotr and Mathewson, Kory W. and Pittman, Jaylen and Evans, Richard},
title = {Co-Writing Screenplays and Theatre Scripts with Language Models: Evaluation by Industry Professionals},
year = {2023},
isbn = {9781450394215},
doi = {10.1145/3544548.3581225},
booktitle = {Proceedings of the 2023 CHI Conference on Human Factors in Computing Systems},
articleno = {355},
numpages = {34},
}

@article{Metzler:Rethinking:2021,
author = {Metzler, Donald and Tay, Yi and Bahri, Dara and Najork, Marc},
title = {Rethinking search: making domain experts out of dilettantes},
year = {2021},
issue_date = {June 2021},
volume = {55},
number = {1},
doi = {10.1145/3476415.3476428},
journal = {SIGIR Forum},
month = {jul},
articleno = {13},
numpages = {27}
}

@inproceedings{zhao:videoprism:2024,
  title={{VideoPrism: A} Foundational Visual Encoder for Video Understanding},
  author={Zhao, Long and Gundavarapu, Nitesh B. and Yuan, Liangzhe and Zhou, Hao and Yan, Shen and Jennifer J. and others},
  booktitle={Proceedings of the 41st International Conference on Machine Learning},
  year={2024}
}

@inproceedings{li:unigen:2024,
  title={{UniGen: A} Unified Generative Framework for Retrieval and Question Answering with Large Language Models},
  author={Li, X and Zhou, Y and Dou, Z},
  booktitle={Proceedings of the AAAI Conference on Artificial Intelligence},
  volume={38},
  number={8},
  pages={8688--8696},
  year={2024}
}

@InProceedings{Jin:DiffusionRet:2023,
    author    = {Jin, Peng and Li, Hao and Cheng, Zesen and Li, Kehan and Ji, Xiangyang and Liu, Chang and Yuan, Li and Chen, Jie},
    title     = {{DiffusionRet: Generative} Text-Video Retrieval with Diffusion Model},
    booktitle = {Proceedings of the IEEE/CVF International Conference on Computer Vision (ICCV)},
    month     = {October},
    year      = {2023},
    pages     = {2470-2481}
}

@inproceedings{Lu:content:2023,
author = {Lu, Daohan and Wang, Sheng-Yu and Kumari, Nupur and Agarwal, Rohan and Tang, Mia and Bau, David and Zhu, Jun-Yan},
title = {Content-based Search for Deep Generative Models},
year = {2023},
doi = {10.1145/3610548.3618189},
booktitle = {SIGGRAPH Asia 2023 Conference Papers},
articleno = {71},
}

@inproceedings{Rajput:recommender:2023,
 author = {Rajput, Shashank and Mehta, Nikhil and Singh, Anima and Hulikal Keshavan, Raghunandan and Vu, Trung and  others},
 booktitle = {Advances in Neural Information Processing Systems},
 pages = {10299--10315},
 title = {Recommender Systems with Generative Retrieval},
 volume = {36},
 year = {2023}
}

@article{guo:exploring:2024,
  title={Exploring the Interaction of Creative Writers with {AI}-Powered Writing Tools},
  author={Guo, Alicia and Pataranutaporn, Pat and Maes, Pattie},
  year={2024},
  journal={arXiv:2402.12814}

}

@article{zhao:survey:2023,
  title={A Survey of Large Language Models},
  author={Zhao, Wayne Xin and Zhou, Kun and Li, Junyi and Tang, Tianyi and others},
  year={2023},
  journal={arXiv:2303.18223}
}

@article{Chang:Survey:2024,
author = {Chang, Yupeng and Wang, Xu and Wang, Jindong and Wu, Yuan and Yang, Linyi and others},
title = {A Survey on Evaluation of Large Language Models},
year = {2024},
issue_date = {June 2024},
publisher = {Association for Computing Machinery},
address = {New York, NY, USA},
volume = {15},
number = {3},
doi = {10.1145/3641289},
journal = {ACM Trans. Intell. Syst. Technol.},
month = {mar},
articleno = {39},
numpages = {45},
}

@article{openai:gpt4:2023,
  title={{GPT-4} Technical Report},
  author={{OpenAI} and Achiam, Josh and Adler, Steven and Agarwal, Sandhini and others},
  year={2023},
  journal={arXiv: 2303.08774}
}

@InProceedings{Hou:promptboosting:2023,
  title = 	 {{P}rompt{B}oosting: Black-Box Text Classification with Ten Forward Passes},
  author =       {Hou, Bairu and O'Connor, Joe and Andreas, Jacob and Chang, Shiyu and Zhang, Yang},
  booktitle = 	 {Proceedings of the 40th International Conference on Machine Learning},
  pages = 	 {13309--13324},
  year = 	 {2023},
  volume = 	 {202},
  month = 	 {23--29 Jul},
}

@article{AI2025125952,
title = {Contrastive multi-graph learning with neighbor hierarchical sifting for semi-supervised text classification},
journal = {Expert Systems with Applications},
volume = {266},
pages = {125952},
year = {2025},
issn = {0957-4174},
doi = {https://doi.org/10.1016/j.eswa.2024.125952},
author = {Wei Ai and Jianbin Li and Ze Wang and Yingying Wei and Tao Meng and Keqin Li}
}

@inproceedings{Ryu:Cinema:2025,
author = {Ryu, Jeongwoo and Kim, Kyusik and Heo, Dongseok and Song, Hyungwoo and Oh, Changhoon and Suh, Bongwon},
title = {Cinema Multiverse Lounge: Enhancing Film Appreciation via Multi-Agent Conversations},
year = {2025},
publisher = {ACM},
doi = {10.1145/3706598.3713641},
booktitle = {Proceedings of the 2025 CHI Conference on Human Factors in Computing Systems},
articleno = {409},
numpages = {22},
location = {
},
series = {CHI '25}
}

@article{Li:From:2025,
author = {Li, Xiaoxi and Jin, Jiajie and Zhou, Yujia and Zhang, Yuyao and Zhang, Peitian and Zhu, Yutao and Dou, Zhicheng},
title = {From Matching to Generation: A Survey on Generative Information Retrieval},
year = {2025},
issue_date = {May 2025},
publisher = {Association for Computing Machinery},
address = {New York, NY, USA},
volume = {43},
number = {3},
issn = {1046-8188},
note = {https://doi.org/10.1145/3722552},
doi = {10.1145/3722552},
journal = {ACM Trans. Inf. Syst.},
month = may,
articleno = {83},
numpages = {62},
}

@incollection{jiang2025domain,
  author    = {Ning Jiang and Sepideh Hasanzadeh and Vincent G. Duffy},
  title     = {Domain-Tailored Generative AI for Personalized Assistant},
  booktitle = {HCI International 2024 – Late Breaking Papers},
  editor    = {Vincent G. Duffy},
  series    = {Lecture Notes in Computer Science},
  volume    = {15376},
  year      = {2025},
  pages     = {227--237},
  doi       = {10.1007/978-3-031-76809-5_17},
}

@article{safonov2025ntire,
  title   = {{NTIRE} 2025 Challenge on {UGC} Video Enhancement: Methods and Results},
  author  = {Nikolay Safonov and Alexey Bryncev and Andrey Moskalenko and Dmitry Kulikov and others},
  journal = {arXiv preprint arXiv:2505.03007},
  year    = {2025}
}

@article{aggarwal2025evolution,
  author  = {Ankur Aggarwal},
  title   = {Evolution of Recommendation Systems in the Age of Generative AI},
  journal = {International Journal of Science and Research Archive},
  volume  = {14},
  number  = {01},
  pages   = {485--492},
  year    = {2025},
  doi     = {10.30574/ijsra.2025.14.1.0061},
  note     = {https://doi.org/10.30574/ijsra.2025.14.1.0061}
}

@InProceedings{ren:hypersd:2024,
      title={{Hyper-SD: Trajectory} Segmented Consistency Model for Efficient Image Synthesis}, 
      author={Yuxi Ren and Xin Xia and Yanzuo Lu and Jiacheng Zhang and Jie Wu and Pan Xie and Xing Wang and Xuefeng Xiao},
      year={2024},
      booktitle={Advances in Neural Information Processing Systems}
}

@article{krugmann:sentiment:2024,
  title={Sentiment Analysis in the Age of Generative {AI}},
  author={Krugmann, J. O. and Hartmann, J.},
  journal={Customer Needs and Solutions},
  volume={11},
  number={3},
  year={2024}
}

@article{feizi:Online:2023,
  title={Online Advertisements with LLMs: Opportunities and Challenges},
  author={Feizi, Soheil and Hajiaghayi, MohammadTaghi and Rezaei, Keivan and Shin, Suho},
  journal={arXiv preprint arXiv:2311.07601},
  year={2023}
}

@article{Hartmann:More:2023,
title = {More than a Feeling: Accuracy and Application of Sentiment Analysis},
journal = {International Journal of Research in Marketing},
volume = {40},
number = {1},
pages = {75-87},
year = {2023},
doi = {https://doi.org/10.1016/j.ijresmar.2022.05.005},
author = {Jochen Hartmann and Mark Heitmann and Christian Siebert and Christina Schamp},
}

@ARTICLE{Mao:Biases:2023,
  author={Mao, Rui and Liu, Qian and He, Kai and Li, Wei and Cambria, Erik},
  journal={IEEE Transactions on Affective Computing}, 
  title={The Biases of Pre-Trained Language Models: An Empirical Study on Prompt-Based Sentiment Analysis and Emotion Detection}, 
  year={2023},
  volume={14},
  number={3},
  pages={1743-1753},
  doi={10.1109/TAFFC.2022.3204972}}

@inproceedings{kang:gigagan:2023,
  author    = {Kang, Minguk and Zhu, Jun-Yan and Zhang, Richard and Park, Jaesik and Shechtman, Eli and Paris, Sylvain and Park, Taesung},
  title     = {Scaling up GANs for Text-to-Image Synthesis},
  booktitle = {Proceedings of the IEEE Conference on Computer Vision and Pattern Recognition (CVPR)},
  year      = {2023},
}

@article{xu:videogigagan:2024,
  title={{VideoGigaGAN: Towards} Detail-rich Video Super-Resolution},
  author={Xu, Yiran and Park, Taesung and Zhang, Richard and Zhou, Yang and Shechtman, Eli and Liu, Feng and Huang, Jia-Bin and Liu, Difan},
  year={2024},
  journal={arXiv:2404.12388},
  archivePrefix={arXiv},
  primaryClass={cs.CV}
}

@INPROCEEDINGS{Shi:ChatGraph:2023,
  author={Shi, Yucheng and Ma, Hehuan and Zhong, Wenliang and Tan, Qiaoyu and Mai, Gengchen and Li, Xiang and Liu, Tianming and Huang, Junzhou},
  booktitle={2023 IEEE International Conference on Data Mining Workshops (ICDMW)}, 
  title={{ChatGraph: Interpretable} Text Classification by Converting ChatGPT Knowledge to Graphs}, 
  year={2023},
  volume={},
  number={},
  pages={515-520},
  doi={10.1109/ICDMW60847.2023.00073}}

@article{YAO:Survey:2024,
title = {A Survey on Large Language Model (LLM) Security and Privacy: The Good, The Bad, and The Ugly},
journal = {High-Confidence Computing},
volume = {4},
number = {2},
pages = {100211},
year = {2024},
doi = {https://doi.org/10.1016/j.hcc.2024.100211},
author = {Yifan Yao and Jinhao Duan and Kaidi Xu and Yuanfang Cai and Zhibo Sun and Yue Zhang},
}

@article{ippolito:creative:2022,
  title={Creative Writing with an {AI}-Powered Writing Assistant: Perspectives from Professional Writers},
  author={Ippolito, Daphne and Yuan, Ann and Coenen, Andy and Burnam, Sehmon},
  year={2022},
  journal={arXiv:2211.05030}
}

@inproceedings{
sun:text:2023,
title={Text Classification via Large Language Models},
author={Xiaofei Sun and Xiaoya Li and Jiwei Li and Fei Wu and Shangwei Guo and Tianwei Zhang and Guoyin Wang},
booktitle={The 2023 Conference on Empirical Methods in Natural Language Processing},
year={2023}
}

@techreport{Beckett:Generating:2023,
  title={Generating Change: {A} global survey of what news organisations are doing with {AI}},
  author={Charlie Beckett and Mira Yaseen},
  institution={JournalismAI},
  pages={},
  year={2023},
}

@InProceedings{Wang:multi:2021,
    author    = {Wang, Dan and Cui, Xinrui and Chen, Xun and Zou, Zhengxia and Shi, Tianyang and Salcudean, Septimiu and Wang, Z. Jane and Ward, Rabab},
    title     = {Multi-View {3D} Reconstruction With Transformers},
    booktitle = {Proceedings of the IEEE/CVF International Conference on Computer Vision (ICCV)},
    month     = {October},
    year      = {2021},
    pages     = {5722-5731}
}

@inproceedings{Saragadam:wire:2023,
              title={WIRE: Wavelet Implicit Neural Representations},
              author={Saragadam, Vishwanath and LeJeune, Daniel and Tan, Jasper and Balakrishnan, Guha and Veeraraghavan, Ashok and Baraniuk, Richard G},
              booktitle={Conf. Computer Vision and Pattern Recognition},
              year={2023}
            }

@inproceedings{Ke:SAM-HQ:2023,
 author = {Ke, Lei and Ye, Mingqiao and Danelljan, Martin and liu, Yifan and Tai, Yu-Wing and Tang, Chi-Keung and Yu, Fisher},
 booktitle = {Advances in Neural Information Processing Systems},
 pages = {29914--29934},
 title = {Segment Anything in High Quality},
 volume = {36},
 year = {2023}
}

@InProceedings{Kirillov:SAM:2023,
    author    = {Kirillov, Alexander and Mintun, Eric and Ravi, Nikhila and Mao, Hanzi and Rolland, Chloe and Gustafson, Laura and Xiao, Tete and Whitehead, Spencer and Berg, Alexander C. and Lo, Wan-Yen and Dollar, Piotr and Girshick, Ross},
    title     = {Segment Anything},
    booktitle = {Proceedings of the IEEE/CVF International Conference on Computer Vision (ICCV)},
    month     = {October},
    year      = {2023},
    pages     = {4015-4026}
}

@INPROCEEDINGS{Wang:Painter:2023,
  author={Wang, Xinlong and Wang, Wen and Cao, Yue and Shen, Chunhua and Huang, Tiejun},
  booktitle={2023 IEEE/CVF Conference on Computer Vision and Pattern Recognition (CVPR)}, 
  title={Images Speak in Images: A Generalist Painter for In-Context Visual Learning}, 
  year={2023},
  volume={},
  number={},
  pages={6830-6839},
  doi={10.1109/CVPR52729.2023.00660}}

@INPROCEEDINGS{Wang:SegGPT:2023,
  author={Wang, Xinlong and Zhang, Xiaosong and Cao, Yue and Wang, Wen and Shen, Chunhua and Huang, Tiejun},
  booktitle={2023 IEEE/CVF International Conference on Computer Vision (ICCV)}, 
  title={{SegGPT}: Towards Segmenting Everything In Context}, 
  year={2023},
  volume={},
  number={},
  pages={1130-1140},
  doi={10.1109/ICCV51070.2023.00110}}

@inproceedings{Baranchuk:label:2022,
      title={Label-Efficient Semantic Segmentation with Diffusion Models},
      author={Dmitry Baranchuk and Andrey Voynov and Ivan Rubachev and Valentin Khrulkov and Artem Babenko},
      booktitle={International Conference on Learning Representations},
      year={2022}
}

@InProceedings{Xu:Open:2023,
    author    = {Xu, Jiarui and Liu, Sifei and Vahdat, Arash and Byeon, Wonmin and Wang, Xiaolong and De Mello, Shalini},
    title     = {Open-Vocabulary Panoptic Segmentation With Text-to-Image Diffusion Models},
    booktitle = {Proceedings of the IEEE/CVF Conference on Computer Vision and Pattern Recognition (CVPR)},
    month     = {June},
    year      = {2023},
    pages     = {2955-2966}
}

@inproceedings{Bowen:Marked:2022,
  title={Masked-attention Mask Transformer for Universal Image Segmentation},
  author={Bowen Cheng and Ishan Misra and Alexander G. Schwing and Alexander Kirillov and Rohit Girdhar},
  booktitle={Proceedings of the IEEE/CVF Conference on Computer Vision and Pattern Recognition (CVPR)},
  year={2022}
}

@inproceedings{cui:mixformer:2022,
  title={Mixformer: End-to-end tracking with iterative mixed attention},
  author={Cui, Yutao and Jiang, Cheng and Wang, Limin and Wu, Gangshan},
  booktitle={Proceedings of the IEEE/CVF Conference on Computer Vision and Pattern Recognition},
  pages={13608--13618},
  year={2022}
}

@InProceedings{Meinhardt:TrackFormer:2022,
    author    = {Meinhardt, Tim and Kirillov, Alexander and Leal-Taix\'e, Laura and Feichtenhofer, Christoph},
    title     = {TrackFormer: Multi-Object Tracking With Transformers},
    booktitle = {Proceedings of the IEEE/CVF Conference on Computer Vision and Pattern Recognition (CVPR)},
    month     = {June},
    year      = {2022},
    pages     = {8844-8854}
}

@INPROCEEDINGS {Zhang:MOTRv2:2023,
author = {Y. Zhang and T. Wang and X. Zhang},
booktitle = {2023 IEEE/CVF Conference on Computer Vision and Pattern Recognition (CVPR)},
title = {{MOTRv2: Bootstrapping} End-to-End Multi-Object Tracking by Pretrained Object Detectors},
year = {2023},
volume = {},
issn = {},
pages = {22056-22065},
doi = {10.1109/CVPR52729.2023.02112}
}

@article{Li:Transformer:2023,
title = {Transformer for object detection: Review and benchmark},
journal = {Engineering Applications of Artificial Intelligence},
volume = {126},
pages = {107021},
year = {2023},
doi = {https://doi.org/10.1016/j.engappai.2023.107021},
note = {https://www.sciencedirect.com/science/article/pii/S0952197623012058},
author = {Yong Li and Naipeng Miao and Liangdi Ma and Feng Shuang and Xingwen Huang},
}

@INPROCEEDINGS{Chen:SeqTrack:2023,
  author={Chen, Xin and Peng, Houwen and Wang, Dong and Lu, Huchuan and Hu, Han},
  booktitle={2023 IEEE/CVF Conference on Computer Vision and Pattern Recognition (CVPR)}, 
  title={SeqTrack: Sequence to Sequence Learning for Visual Object Tracking}, 
  year={2023},
  volume={},
  number={},
  pages={14572-14581},
  doi={10.1109/CVPR52729.2023.01400}}

@inproceedings{zeng:motr:2022,
  title={MOTR: End-to-End Multiple-Object Tracking with TRansformer},
  author={Zeng, Fangao and Dong, Bin and Zhang, Yuang and Wang, Tiancai and Zhang, Xiangyu and Wei, Yichen},
  booktitle={European Conference on Computer Vision (ECCV)},
  year={2022}
}

@inproceedings{peng2024rmt,
  title={{RMT-BVQA}: Recurrent memory transformer-based blind video quality assessment for enhanced video content},
  author={Peng, Tianhao and Feng, Chen and Danier, Duolikun and Zhang, Fan and Vallade, Benoit and Mackin, Alex and Bull, David},
   booktitle={European Conference on Computer Vision (ECCV) Workshop on Advances in Image Manipulation},
  year={2024}
}

@inproceedings{zhao2023quality,
  title={Quality-aware pre-trained models for blind image quality assessment},
  author={Zhao, Kai and Yuan, Kun and Sun, Ming and Li, Mading and Wen, Xing},
  booktitle={Proceedings of the IEEE/CVF conference on computer vision and pattern recognition},
  pages={22302--22313},
  year={2023}
}

@article{ge:yolox:2021,
  title={YOLOX: Exceeding YOLO Series in 2021},
  author={Ge, Zheng and Liu, Songtao and Wang, Feng and Li, Zeming and Sun, Jian},
  journal={arXiv preprint arXiv:2107.08430},
  year={2021}
}

@ARTICLE{Zou:object:2023,
  author={Zou, Zhengxia and Chen, Keyan and Shi, Zhenwei and Guo, Yuhong and Ye, Jieping},
  journal={Proceedings of the IEEE}, 
  title={Object Detection in 20 Years: A Survey}, 
  year={2023},
  volume={111},
  number={3},
  pages={257-276},
  doi={10.1109/JPROC.2023.3238524}}

@inproceedings{cheng:xmem:2022,
  title={Xmem: Long-term video object segmentation with an Atkinson-Shiffrin memory model},
  author={Cheng, Ho Kei and Schwing, Alexander G.},
  booktitle={European Conference on Computer Vision (ECCV)},
  volume={13688},
  pages={640--658},
  year={2022}
}

@ARTICLE{yang:track:2023,
      title={Track Anything: Segment Anything Meets Videos}, 
      author={Jinyu Yang and Mingqi Gao and Zhe Li and Shang Gao and Fangjing Wang and Feng Zheng},
      year={2023},
      journal={arXiv:2304.11968},
}

@InProceedings{Jung:AnyFlow:2023,
    author    = {Jung, Hyunyoung and Hui, Zhuo and Luo, Lei and Yang, Haitao and Liu, Feng and Yoo, Sungjoo and Ranjan, Rakesh and Demandolx, Denis},
    title     = {{AnyFlow: Arbitrary} Scale Optical Flow With Implicit Neural Representation},
    booktitle = {Proceedings of the IEEE/CVF Conference on Computer Vision and Pattern Recognition (CVPR)},
    month     = {June},
    year      = {2023},
    pages     = {5455-5465}
}

@InProceedings{Guo:neural:2022,
    author    = {Guo, Haoyu and Peng, Sida and Lin, Haotong and Wang, Qianqian and Zhang, Guofeng and Bao, Hujun and Zhou, Xiaowei},
    title     = {Neural {3D} Scene Reconstruction With the Manhattan-World Assumption},
    booktitle = {Proceedings of the IEEE/CVF Conference on Computer Vision and Pattern Recognition (CVPR)},
    month     = {June},
    year      = {2022},
    pages     = {5511-5520}
}

@article{azzarelli:waveplanes:2023,
  title={{WavePlanes: A} Compact Wavelet Representation for Dynamic Neural Radiance Fields},
  author={Azzarelli, Adrian and Anantrasirichai, Nantheera and Bull, David R},
  journal={arXiv preprint arXiv:2312.02218},
  year={2023}
}

@InProceedings{Goel:Interactive:2023,
    author    = {Goel, Rahul and Sirikonda, Dhawal and Saini, Saurabh and Narayanan, P. J.},
    title     = {Interactive Segmentation of Radiance Fields},
    booktitle = {Proceedings of the IEEE/CVF Conference on Computer Vision and Pattern Recognition (CVPR)},
    month     = {June},
    year      = {2023},
    pages     = {4201-4211}
}

@INPROCEEDINGS{Gu:Diffusioninst:2024,
  author={Gu, Zhangxuan and Chen, Haoxing and Xu, Zhuoer},
  booktitle={ICASSP 2024 - 2024 IEEE International Conference on Acoustics, Speech and Signal Processing (ICASSP)}, 
  title={{Diffusioninst: Diffusion} Model for Instance Segmentation}, 
  year={2024},
  volume={},
  number={},
  pages={2730-2734},
  doi={10.1109/ICASSP48485.2024.10447191}}

@InProceedings{Wu:DiffuMask:2023,
    author    = {Wu, Weijia and Zhao, Yuzhong and Shou, Mike Zheng and Zhou, Hong and Shen, Chunhua},
    title     = {{DiffuMask: Synthesizing} Images with Pixel-level Annotations for Semantic Segmentation Using Diffusion Models},
    booktitle = {Proceedings of the IEEE/CVF International Conference on Computer Vision (ICCV)},
    month     = {October},
    year      = {2023},
    pages     = {1206-1217}
}

@InProceedings{Yan:Universal:2023,
    author    = {Yan, Bin and Jiang, Yi and Wu, Jiannan and Wang, Dong and Luo, Ping and Yuan, Zehuan and Lu, Huchuan},
    title     = {Universal Instance Perception As Object Discovery and Retrieval},
    booktitle = {Proceedings of the IEEE/CVF Conference on Computer Vision and Pattern Recognition (CVPR)},
    month     = {June},
    year      = {2023},
    pages     = {15325-15336}
}

@inproceedings{Zou:Segment:2023,
 author = {Zou, Xueyan and Yang, Jianwei and Zhang, Hao and Li, Feng and Li, Linjie and Wang, Jianfeng and Wang, Lijuan and Gao, Jianfeng and Lee, Yong Jae},
 booktitle = {Advances in Neural Information Processing Systems},
 pages = {19769--19782},
 publisher = {Curran Associates, Inc.},
 title = {Segment Everything Everywhere All at Once},
 volume = {36},
 year = {2023}
}

@inproceedings{sitzmann:siren:2020,
                author = {Sitzmann, Vincent
                          and Martel, Julien N.P.
                          and Bergman, Alexander W.
                          and Lindell, David B.
                          and Wetzstein, Gordon},
                title = {Implicit Neural Representations
                          with Periodic Activation Functions},
                booktitle = {Proc. NeurIPS},
                year={2020}
            }

@article{Ren:Faster:2027,
  author={Ren, Shaoqing and He, Kaiming and Girshick, Ross and Sun, Jian},
  journal={IEEE Transactions on Pattern Analysis and Machine Intelligence}, 
  title={Faster R-CNN: Towards Real-Time Object Detection with Region Proposal Networks}, 
  year={2017},
  volume={39},
  number={6},
  pages={1137-1149},
  doi={10.1109/TPAMI.2016.2577031}}

@inproceedings{Carion:DERT:2020,
  title={End-to-end object detection with transformers},
  author={Carion, Nicolas and Massa, Francisco and Synnaeve, Gabriel and Usunier, Nicolas and Kirillov, Alexander and Zagoruyko, Sergey},
  booktitle={Eur. Conf. Comput. Vis.},
  pages={213--229},
  year={2020}
}

@inproceedings{Dosovitskiy:image:2021,
title={An Image is Worth 16$\times$16 Words: {T}ransformers for Image Recognition at Scale},
author={Alexey Dosovitskiy and Lucas Beyer and Alexander Kolesnikov and Dirk Weissenborn and Xiaohua Zhai and Thomas Unterthiner and Mostafa Dehghani and Matthias Minderer and Georg Heigold and Sylvain Gelly and Jakob Uszkoreit and Neil Houlsby},
booktitle={International Conference on Learning Representations},
year={2021}
}

@article{Li:HAM:2022,
title = {{HAM: Hybrid} attention module in deep convolutional neural networks for image classification},
journal = {Pattern Recognition},
volume = {129},
pages = {108785},
year = {2022},
author = {Guoqiang Li and Qi Fang and Linlin Zha and Xin Gao and Nenggan Zheng},
}

@article{Guo:Attention:2022,
  title={Attention mechanisms in computer vision: A survey},
  author={Guo, Ming-Hao and Xu, Tian-Xiang and Liu, Jia-Jun and et al.},
  journal={Computational Visual Media},
  volume={8},
  pages={331–368},
  year={2022}
}

@InProceedings{Woo:CBAM:2018,
author = {Woo, Sanghyun and Park, Jongchan and Lee, Joon-Young and Kweon, In So},
title = {{CBAM: Convolutional} Block Attention Module},
booktitle = {Proceedings of the European Conference on Computer Vision (ECCV)},
month = {September},
year = {2018}
}

@Article{UK:Large:2024,
  Title                    = {Large language models and generative {AI}},
  Author                   = {{Communications and Digital Committee}},
  Journal                  = {1st Report of Session 2023–24},
  Year                     = {2024},
}

@incollection{Goodfellow:GAN:2014,
title = {Generative Adversarial Nets},
author = {Goodfellow, Ian and Pouget-Abadie, Jean and Mirza, Mehdi and Xu, Bing and Warde-Farley, David and Ozair, Sherjil and Courville, Aaron and Bengio, Yoshua},
booktitle = {Advances in Neural Information Processing Systems 27},
editor = {Z. Ghahramani and M. Welling and C. Cortes and N. D. Lawrence and K. Q. Weinberger},
pages = {2672--2680},
year = {2014},
publisher = {Curran Associates, Inc.},
note = {http://papers.nips.cc/paper/5423-generative-adversarial-nets.pdf}
}

@ARTICLE{Saharia:image:2023,
  author={Saharia, Chitwan and Ho, Jonathan and Chan, William and Salimans, Tim and Fleet, David J. and Norouzi, Mohammad},
  journal={IEEE Transactions on Pattern Analysis and Machine Intelligence}, 
  title={Image Super-Resolution via Iterative Refinement}, 
  year={2023},
  volume={45},
  number={4},
  pages={4713-4726},
  doi={10.1109/TPAMI.2022.3204461}}

@inproceedings{Ye:FLASK:2024,
title={{FLASK}: Fine-grained Language Model Evaluation based on Alignment Skill Sets},
author={Seonghyeon Ye and Doyoung Kim and Sungdong Kim and Hyeonbin Hwang and Seungone Kim and Yongrae Jo and James Thorne and Juho Kim and Minjoon Seo},
booktitle={The Twelfth International Conference on Learning Representations},
year={2024},
note={https://openreview.net/forum?id=CYmF38ysDa}
}

@InProceedings{Gao:Implicit:2023,
    author    = {Gao, Sicheng and Liu, Xuhui and Zeng, Bohan and Xu, Sheng and Li, Yanjing and Luo, Xiaoyan and Liu, Jianzhuang and Zhen, Xiantong and Zhang, Baochang},
    title     = {Implicit Diffusion Models for Continuous Super-Resolution},
    booktitle = {Proceedings of the IEEE/CVF Conference on Computer Vision and Pattern Recognition (CVPR)},
    month     = {June},
    year      = {2023},
    pages     = {10021-10030}
}

@InProceedings{Chan:BasicVSR:2022,
    author    = {Chan, Kelvin C.K. and Zhou, Shangchen and Xu, Xiangyu and Loy, Chen Change},
    title     = {BasicVSR++: Improving Video Super-Resolution With Enhanced Propagation and Alignment},
    booktitle = {Proceedings of the IEEE/CVF Conference on Computer Vision and Pattern Recognition (CVPR)},
    month     = {June},
    year      = {2022},
    pages     = {5972-5981}
}

@article{quan:deep:2024,
  title={Deep Learning-Based Image and Video Inpainting: A Survey},
  author={Quan, W. and Chen, J. and Liu, Y. and et al.},
  journal={International Journal of Computer Vision},
  year={2024}
}

@article{HUANG:Sparse:2024,
title = {Sparse self-attention transformer for image inpainting},
journal = {Pattern Recognition},
volume = {145},
pages = {109897},
year = {2024},
issn = {0031-3203},
author = {Wenli Huang and Ye Deng and Siqi Hui and Yang Wu and Sanping Zhou and Jinjun Wang}
}

@InProceedings{Li:MAT:2022,
    author    = {Li, Wenbo and Lin, Zhe and Zhou, Kun and Qi, Lu and Wang, Yi and Jia, Jiaya},
    title     = {{MAT: Mask}-Aware Transformer for Large Hole Image Inpainting},
    booktitle = {Proceedings of the IEEE/CVF Conference on Computer Vision and Pattern Recognition (CVPR)},
    month     = {June},
    year      = {2022},
    pages     = {10758-10768}
}

@InProceedings{Zhou:ProPainter:2023,
    author    = {Zhou, Shangchen and Li, Chongyi and Chan, Kelvin C.K. and Loy, Chen Change},
    title     = {{ProPainter: Improving} Propagation and Transformer for Video Inpainting},
    booktitle = {Proceedings of the IEEE/CVF International Conference on Computer Vision (ICCV)},
    month     = {October},
    year      = {2023},
    pages     = {10477-10486}
}

@InProceedings{Ren:DLFormer:2022,
    author    = {Ren, Jingjing and Zheng, Qingqing and Zhao, Yuanyuan and Xu, Xuemiao and Li, Chen},
    title     = {{DLFormer: Discrete} Latent Transformer for Video Inpainting},
    booktitle = {Proceedings of the IEEE/CVF Conference on Computer Vision and Pattern Recognition (CVPR)},
    month     = {June},
    year      = {2022},
    pages     = {3511-3520}
}

@InProceedings{Liu:Reduce:2022,
    author    = {Liu, Qiankun and Tan, Zhentao and Chen, Dongdong and Chu, Qi and Dai, Xiyang and Chen, Yinpeng and Liu, Mengchen and Yuan, Lu and Yu, Nenghai},
    title     = {Reduce Information Loss in Transformers for Pluralistic Image Inpainting},
    booktitle = {Proceedings of the IEEE/CVF Conference on Computer Vision and Pattern Recognition (CVPR)},
    month     = {June},
    year      = {2022},
    pages     = {11347-11357}
}

@inproceedings{zheng:pluralistic:2019,
  title={Pluralistic Image Completion},
  author={Zheng, Chuanxia and Cham, Tat-Jen and Cai, Jianfei},
  booktitle={Proceedings of the IEEE Conference on Computer Vision and Pattern Recognition},
  pages={1438--1447},
  year={2019}
}

@inproceedings{singer:Make:2023,
title={{Make-A-Video: Text}-to-Video Generation without Text-Video Data},
author={Uriel Singer and Adam Polyak and Thomas Hayes and Xi Yin and Jie An and Songyang Zhang and Qiyuan Hu and Harry Yang and Oron Ashual and Oran Gafni and Devi Parikh and Sonal Gupta and Yaniv Taigman},
booktitle={The Eleventh International Conference on Learning Representations },
year={2023},
}

@book{encyclopedia_ai_v1,
  title = {Encyclopedia of Artificial Intelligence},
  volume = {1},
  editor = {Stuart C. Shapiro and David Eckroth },
  publisher = {John Wiley \& Sons},
  year = {1987},
  address = {New York}
}

@InProceedings{Fuoli:Fast:2023,
    author    = {Fuoli, Dario and Danelljan, Martin and Timofte, Radu and Van Gool, Luc},
    title     = {Fast Online Video Super-Resolution With Deformable Attention Pyramid},
    booktitle = {Proceedings of the IEEE/CVF Winter Conference on Applications of Computer Vision (WACV)},
    month     = {January},
    year      = {2023},
    pages     = {1735-1744}
}

@InProceedings{Liu:Learning:2022,
    author    = {Liu, Chengxu and Yang, Huan and Fu, Jianlong and Qian, Xueming},
    title     = {Learning Trajectory-Aware Transformer for Video Super-Resolution},
    booktitle = {Proceedings of the IEEE/CVF Conference on Computer Vision and Pattern Recognition (CVPR)},
    month     = {June},
    year      = {2022},
    pages     = {5687-5696}
}

@inproceedings{cao2025zero,
  author    = {Chao Cao and Hongguang Yue and Xinzhe Liu and Jing Yang},
  title     = {Zero-shot Video Restoration and Enhancement Using Pre-Trained Image Diffusion Model},
  booktitle = {Proceedings of the AAAI Conference on Artificial Intelligence},
  year      = {2025},
  volume    = {39},
  number    = {2},
  pages     = {1935--1943},
  doi       = {10.1609/aaai.v39i2.32189}
}

@ARTICLE{Shi:VmambaIR:2025,
  author={Shi, Yuan and Xia, Bin and Jin, Xiaoyu and Wang, Xing and Zhao, Tianyu and Xia, Xin and Xiao, Xuefeng and Yang, Wenming},
  journal={IEEE Transactions on Circuits and Systems for Video Technology}, 
  title={{VmambaIR: Visual} State Space Model for Image Restoration}, 
  year={2025},
  volume={},
  number={},
  doi={10.1109/TCSVT.2025.3530090}}

@inproceedings{wang2025seedvr,
      title={{SeedVR: Seeding} Infinity in Diffusion Transformer Towards Generic Video Restoration},
      author={Wang, Jianyi and Lin, Zhijie and Wei, Meng and Zhao, Yang and Yang, Ceyuan and Loy, Chen Change and Jiang, Lu},
      booktitle={IEEE Conference on Computer Vision and Pattern Recognition (CVPR)},
      year={2025}
   }

@article{moser:diffusion:2024,
  author={Moser, Brian B. and Shanbhag, Arundhati S. and Raue, Federico and Frolov, Stanislav and Palacio, Sebastian and Dengel, Andreas},
  journal={IEEE Transactions on Neural Networks and Learning Systems}, 
  title={Diffusion Models, Image Super-Resolution, and Everything: A Survey}, 
  year={2024},
  volume={},
  number={},
  pages={1-21},
  doi={10.1109/TNNLS.2024.3476671}}

@article{Bai:Train:2021,
  title={Training a Helpful and Harmless Assistant with Reinforcement Learning from Human Feedback},
  author={Bai, Yuntao and Jones, Andy and Ndousse, Kamal and others},
  journal={arXiv:2204.05862},
  year={2021}
}

@inproceedings{Lester:power:2021,
    title = "The Power of Scale for Parameter-Efficient Prompt Tuning",
    author = "Lester, Brian  and
      Al-Rfou, Rami  and
      Constant, Noah",
    booktitle = "Proceedings of the 2021 Conference on Empirical Methods in Natural Language Processing",
    month = nov,
    year = "2021",
    doi = "10.18653/v1/2021.emnlp-main.243",
    pages = "3045--3059",
}

@INPROCEEDINGS{Wang:Uformer:2022,
  author={Wang, Zhendong and Cun, Xiaodong and Bao, Jianmin and Zhou, Wengang and Liu, Jianzhuang and Li, Houqiang},
  booktitle={2022 IEEE/CVF Conference on Computer Vision and Pattern Recognition (CVPR)}, 
  title={Uformer: A General U-Shaped Transformer for Image Restoration}, 
  year={2022},
  volume={},
  number={},
  pages={17662-17672},}

@inproceedings{Zamir:Restormer:2022,
  author={Zamir, Syed Waqas and Arora, Aditya and Khan, Salman and Hayat, Munawar and Khan, Fahad Shahbaz and Yang, Ming–Hsuan},
  booktitle={2022 IEEE/CVF Conference on Computer Vision and Pattern Recognition (CVPR)}, 
  title={Restormer: Efficient Transformer for High-Resolution Image Restoration}, 
  year={2022},
  volume={},
  number={},
  pages={5718-5729}}

@inproceedings{Fridovich:kplanes:2023,
      title={K-Planes: Explicit Radiance Fields in Space, Time, and Appearance},
      author={{Sara Fridovich-Keil and Giacomo Meanti} and Frederik Rahbæk Warburg and Benjamin Recht and Angjoo Kanazawa},
      year={2023},
      booktitle={Proceedings of the IEEE/CVF Conference on Computer Vision and Pattern Recognition (CVPR)}
}

@InProceedings{Mildenhall:NeRF:2020,
author="Mildenhall, Ben
and Srinivasan, Pratul P.
and Tancik, Matthew
and Barron, Jonathan T.
and Ramamoorthi, Ravi
and Ng, Ren",
editor="Vedaldi, Andrea
and Bischof, Horst
and Brox, Thomas
and Frahm, Jan-Michael",
title="{NeRF: Representing} Scenes as Neural Radiance Fields for View Synthesis",
booktitle="Computer Vision -- ECCV 2020",
year="2020",
pages="405--421"
}

@article{Oquab:DINOv2:2024,
title={{DINOv2: Learning} Robust Visual Features without Supervision},
author={Maxime Oquab and Timoth{\'e}e Darcet and Th{\'e}o Moutakanni and Huy V. Vo and Marc Szafraniec and Vasil Khalidov and Pierre Fernandez and Daniel HAZIZA and Francisco Massa and Alaaeldin El-Nouby and Mido Assran and Nicolas Ballas and Wojciech Galuba and Russell Howes and Po-Yao Huang and Shang-Wen Li and Ishan Misra and Michael Rabbat and Vasu Sharma and Gabriel Synnaeve and Hu Xu and Herve Jegou and Julien Mairal and Patrick Labatut and Armand Joulin and Piotr Bojanowski},
journal={Transactions on Machine Learning Research},
issn={2835-8856},
year={2024},
note={}
}

@inproceedings{wynn:diffusionerf:2023,
 title   = {{DiffusioNeRF: Regularizing Neural Radiance Fields with Denoising Diffusion Models}},
 author  = {Jamie Wynn and
            Daniyar Turmukhambetov
           },
 booktitle = {Proceedings of the IEEE/CVF Conference on Computer Vision and Pattern Recognition (CVPR)},
 year = {2023}
}

@Article{kerbl:3Dgaussians:2023,
      author       = {Kerbl, Bernhard and Kopanas, Georgios and Leimk{\"u}hler, Thomas and Drettakis, George},
      title        = {{3D} Gaussian Splatting for Real-Time Radiance Field Rendering},
      journal      = {ACM Transactions on Graphics},
      number       = {4},
      volume       = {42},
      month        = {July},
      year         = {2023},
      note          = {https://repo-sam.inria.fr/fungraph/3d-gaussian-splatting/}
}

@INPROCEEDINGS{Barron:Mip-NeRF360:2022,
  author={Barron, Jonathan T. and Mildenhall, Ben and Verbin, Dor and Srinivasan, Pratul P. and Hedman, Peter},
  booktitle={2022 IEEE/CVF Conference on Computer Vision and Pattern Recognition (CVPR)}, 
  title={{Mip-NeRF 360: Unbounded} Anti-Aliased Neural Radiance Fields}, 
  year={2022},
  volume={},
  number={},
  pages={5460-5469},
  doi={10.1109/CVPR52688.2022.00539}}

@inproceedings{jiang:vrgs:2024,
author = {Jiang, Ying and Yu, Chang and Xie, Tianyi and Li, Xuan and Feng, Yutao and Wang, Huamin and Li, Minchen and Lau, Henry and Gao, Feng and Yang, Yin and Jiang, Chenfanfu},
title = {{VR-GS: A} Physical Dynamics-Aware Interactive Gaussian Splatting System in Virtual Reality},
year = {2024},
doi = {10.1145/3641519.3657448},
booktitle = {ACM SIGGRAPH 2024 Conference Papers}
}

@inproceedings{schoenberger:sfm:2016,
    author={Sch\"{o}nberger, Johannes Lutz and Frahm, Jan-Michael},
    title={Structure-from-Motion Revisited},
    booktitle={Conference on Computer Vision and Pattern Recognition (CVPR)},
    year={2016},
}

@InProceedings{Zhang:Inversion:2023,
    author    = {Zhang, Yuxin and Huang, Nisha and Tang, Fan and Huang, Haibin and Ma, Chongyang and Dong, Weiming and Xu, Changsheng},
    title     = {Inversion-Based Style Transfer With Diffusion Models},
    booktitle = {Proceedings of the IEEE/CVF Conference on Computer Vision and Pattern Recognition (CVPR)},
    month     = {June},
    year      = {2023},
    pages     = {10146-10156}
}

@InProceedings{Yu:CoGS:2024,
    author    = {Yu, Heng and Julin, Joel and Milacski, Zolt\'an A. and Niinuma, Koichiro and Jeni, L\'aszl\'o A.},
    title     = {{CoGS: Controllable} Gaussian Splatting},
    booktitle = {Proceedings of the IEEE/CVF Conference on Computer Vision and Pattern Recognition (CVPR)},
    month     = {June},
    year      = {2024},
    pages     = {21624-21633}
}

@InProceedings{Huang:SCGS:2024,
    author    = {Huang, Yi-Hua and Sun, Yang-Tian and Yang, Ziyi and Lyu, Xiaoyang and Cao, Yan-Pei and Qi, Xiaojuan},
    title     = {{SC-GS: Sparse}-Controlled Gaussian Splatting for Editable Dynamic Scenes},
    booktitle = {Proceedings of the IEEE/CVF Conference on Computer Vision and Pattern Recognition (CVPR)},
    month     = {June},
    year      = {2024},
    pages     = {4220-4230}
}

@InProceedings{wu:4dgaussians:2024,
      title={4D Gaussian Splatting for Real-Time Dynamic Scene Rendering},
      author={Wu, Guanjun and Yi, Taoran and Fang, Jiemin and Xie, Lingxi and Zhang, Xiaopeng and Wei Wei and Liu, Wenyu and Tian, Qi and Wang Xinggang},
      booktitle={Proceedings of the IEEE/CVF Conference on Computer Vision and Pattern Recognition (CVPR)},
      year={2024}
    }

@article{ren:dreamgaussian4d:2023,
  title={{DreamGaussian4D: Generative} 4D Gaussian Splatting},
  author={Ren, Jiawei and Pan, Liang and Tang, Jiaxiang and Zhang, Chi and Cao, Ang and Zeng, Gang and Liu, Ziwei},
  journal={arXiv preprint arXiv:2312.17142},
  year={2023}
}

@INPROCEEDINGS{Choi:Exploring:2023,
  author={Choi, Myungsub and Lee, Hana and Lee, Hyong-euk},
  booktitle={2023 IEEE/CVF International Conference on Computer Vision (ICCV)}, 
  title={Exploring Positional Characteristics of Dual-Pixel Data for Camera Autofocus}, 
  year={2023},
  volume={},
  number={},
  pages={13112-13122},
  doi={10.1109/ICCV51070.2023.01210}}

@article{yang:realworld:2023,
  title={Real-World Denoising via Diffusion Model},
  author={Yang, Cheng and Liang, Lijing and Su, Zhixun},
  journal={arXiv preprint arXiv:2305.04457},
  year={2023}
}

@ARTICLE{Song:vision:2023,
  author={Song, Yuda and He, Zhuqing and Qian, Hui and Du, Xin},
  journal={IEEE Transactions on Image Processing}, 
  title={Vision Transformers for Single Image Dehazing}, 
  year={2023},
  volume={32},
  number={},
  pages={1927-1941},
  doi={10.1109/TIP.2023.3256763}}

@article{yang:ldp:2023,
  author = {Hao Yang and Liyuan Pan and Yan Yang and Richard Hartley and Miaomiao Liu},
  title = {{LDP}: Language-driven Dual-Pixel Image Defocus Deblurring Network},
  journal = {arXiv: 2307.09815},
  year = {2023},
  eprint = {}
}

@InProceedings{Yang:K3DN:2023,
    author    = {Yang, Yan and Pan, Liyuan and Liu, Liu and Liu, Miaomiao},
    title     = {{K3DN: Disparity}-Aware Kernel Estimation for Dual-Pixel Defocus Deblurring},
    booktitle = {Proceedings of the IEEE/CVF Conference on Computer Vision and Pattern Recognition (CVPR)},
    month     = {June},
    year      = {2023},
    pages     = {13263-13272}
}

@inproceedings{xu:VASA-1:2024,
title={{VASA}-1: Lifelike Audio-Driven Talking Faces Generated in Real Time},
author={Sicheng Xu and Guojun Chen and Yu-Xiao Guo and Jiaolong Yang and Chong Li and Zhenyu Zang and Yizhong Zhang and Xin Tong and Baining Guo},
booktitle={The Thirty-eighth Annual Conference on Neural Information Processing Systems},
year={2024}
}

@inproceedings{chung:human-loop:2021,
  author    = {Neo Christopher Chung},
  title     = {Human in the Loop for Machine Creativity},
  booktitle = {AAAI Conference on Human Computation and Crowdsourcing},
  year      = {2021}
}

@inproceedings{Gupta:Photorealistic:2024,
author = {Gupta, Agrim and Yu, Lijun and Sohn, Kihyuk and Gu, Xiuye and Hahn, Meera and Li, Fei-Fei and Essa, Irfan and Jiang, Lu and Lezama, Jos\'{e}},
title = {Photorealistic Video Generation with Diffusion Models},
year = {2024},
doi = {10.1007/978-3-031-72986-7_23},
booktitle = {European Conference Computer Vision},
pages = {393–411},
}

@inproceedings{Shi:Motion-I2V:2024,
author = {Shi, Xiaoyu and Huang, Zhaoyang and Wang, Fu-Yun and Bian, Weikang and Li, Dasong and Zhang, Yi and Zhang, Manyuan and Cheung, Ka Chun and See, Simon and Qin, Hongwei and Dai, Jifeng and Li, Hongsheng},
title = {{Motion-I2V: Consistent} and Controllable Image-to-Video Generation with Explicit Motion Modeling},
year = {2024},
doi = {10.1145/3641519.3657497},
booktitle = {ACM SIGGRAPH Conference Papers},
articleno = {111},
numpages = {11},
}

@InProceedings{Hu_2024_CVPR,
    author    = {Hu, Li},
    title     = {{Animate Anyone: Consistent} and Controllable Image-to-Video Synthesis for Character Animation},
    booktitle = {Proceedings of the IEEE/CVF Conference on Computer Vision and Pattern Recognition (CVPR)},
    month     = {June},
    year      = {2024},
    pages     = {8153-8163}
}

@article{corona:vlogger:2024,
title={VLOGGER: Multimodal Diffusion for Embodied Avatar Synthesis},
author={Corona, Enric and Zanfir, Andrei and Bazavan, Eduard Gabriel and Kolotouros, Nikos and Alldieck, Thiemo and Sminchisescu, Cristian},
journal={	arXiv:2403.08764 },
year={2024}
}

@InProceedings{Chai:StableVideo:2023,
    author    = {Chai, Wenhao and Guo, Xun and Wang, Gaoang and Lu, Yan},
    title     = {{StableVideo: Text}-driven Consistency-aware Diffusion Video Editing},
    booktitle = {Proceedings of the IEEE/CVF International Conference on Computer Vision (ICCV)},
    month     = {October},
    year      = {2023},
    pages     = {23040-23050}
}

@InProceedings{Pan:Deep:2023,
    author    = {Pan, Jinshan and Xu, Boming and Dong, Jiangxin and Ge, Jianjun and Tang, Jinhui},
    title     = {Deep Discriminative Spatial and Temporal Network for Efficient Video Deblurring},
    booktitle = {Proceedings of the IEEE/CVF Conference on Computer Vision and Pattern Recognition (CVPR)},
    month     = {June},
    year      = {2023},
    pages     = {22191-22200}
}

@InProceedings{Deng:StyTr2:2022,
    author    = {Deng, Yingying and Tang, Fan and Dong, Weiming and Ma, Chongyang and Pan, Xingjia and Wang, Lei and Xu, Changsheng},
    title     = {{StyTr2: Image} Style Transfer With Transformers},
    booktitle = {Proceedings of the IEEE/CVF Conference on Computer Vision and Pattern Recognition (CVPR)},
    month     = {June},
    year      = {2022},
    pages     = {11326-11336}
}

@inproceedings{
tang:dreamgaussian:2024,
title={DreamGaussian: Generative Gaussian Splatting for Efficient {3D} Content Creation},
author={Jiaxiang Tang and Jiawei Ren and Hang Zhou and Ziwei Liu and Gang Zeng},
booktitle={The Twelfth International Conference on Learning Representations},
year={2024}
}

@inproceedings{
Qian:Magic123:2024,
title={{Magic123: One} Image to High-Quality {3D} Object Generation Using Both 2D and {3D} Diffusion Priors},
author={Qian, Guocheng and Mai, Jinjie and Hamdi, Abdullah and Ren, Jian and Siarohin, Aliaksandr and Li, Bing and Lee, Hsin-Ying and Skorokhodov, Ivan and Wonka, Peter and Tulyakov, Sergey and Ghanem, Bernard},
booktitle={The Twelfth International Conference on Learning Representations (ICLR)},
year={2024}
}

@article{mueller:instant:2022,
    author = {Thomas M\"uller and Alex Evans and Christoph Schied and Alexander Keller},
    title = {Instant Neural Graphics Primitives with a Multiresolution Hash Encoding},
    journal = {ACM Trans. Graph.},
    issue_date = {July 2022},
    volume = {41},
    number = {4},
    month = jul,
    year = {2022},
    pages = {102:1--102:15},
    articleno = {102},
    doi = {10.1145/3528223.3530127},
}

@inproceedings{Yang:depthanything:2024,
      title={Depth Anything: Unleashing the Power of Large-Scale Unlabeled Data}, 
      author={Yang, Lihe and Kang, Bingyi and Huang, Zilong and Xu, Xiaogang and Feng, Jiashi and Zhao, Hengshuang},
      booktitle={IEEE/CVF Conference on Computer Vision and Pattern Recognition (CVPR)},
      year={2024}
}

@INPROCEEDINGS{Yang:depthanythingv2:2024,
  title={Depth Anything V2},
  author={Yang, Lihe and Kang, Bingyi and Huang, Zilong and Zhao, Zhen and Xu, Xiaogang and Feng, Jiashi and Zhao, Hengshuang},
  booktitle={Advances in Neural Information Processing
Systems},
  year={2024}
}

@INPROCEEDINGS{Ji:DDP:2023,
  author={Ji, Yuanfeng and Chen, Zhe and Xie, Enze and Hong, Lanqing and Liu, Xihui and Liu, Zhaoqiang and Lu, Tong and Li, Zhenguo and Luo, Ping},
  booktitle={2023 IEEE/CVF International Conference on Computer Vision (ICCV)}, 
  title={{DDP: Diffusion} Model for Dense Visual Prediction}, 
  year={2023},
  volume={},
  number={},
  pages={21684-21695},
  doi={10.1109/ICCV51070.2023.01987}}

@inproceedings{Chen:Vision:2023,
  author    = {Zhe Chen and Yuchen Duan and Wenhai Wang and Junjun He and Tong Lu and Jifeng Dai and Yu Qiao},
  title     = {Vision Transformer Adapter for Dense Predictions},
  booktitle = {Proceedings of the International Conference on Learning Representations (ICLR)},
  year      = {2023}
}

@InProceedings{Zhang:Lite:2023,
    author    = {Zhang, Ning and Nex, Francesco and Vosselman, George and Kerle, Norman},
    title     = {{Lite-Mono: A} Lightweight CNN and Transformer Architecture for Self-Supervised Monocular Depth Estimation},
    booktitle = {Proceedings of the IEEE/CVF Conference on Computer Vision and Pattern Recognition (CVPR)},
    month     = {June},
    year      = {2023},
    pages     = {18537-18546}
}

@InProceedings{Ke:Repurposing:2024,
    author    = {Ke, Bingxin and Obukhov, Anton and Huang, Shengyu and Metzger, Nando and Daudt, Rodrigo Caye and Schindler, Konrad},
    title     = {Repurposing Diffusion-Based Image Generators for Monocular Depth Estimation},
    booktitle = {Proceedings of the IEEE/CVF Conference on Computer Vision and Pattern Recognition (CVPR)},
    month     = {June},
    year      = {2024},
    pages     = {9492-9502}
}

@inproceedings{xie2022neural,
  title={Neural fields in visual computing and beyond},
  author={Xie, Yiheng and Takikawa, Towaki and Saito, Shunsuke and Litany, Or and Yan, Shiqin and Khan, Numair and Tombari, Federico and Tompkin, James and Sitzmann, Vincent and Sridhar, Srinath},
  booktitle={Computer Graphics Forum},
  volume={41(2)},
  pages={641--676},
  year={2022},
  organization={Wiley Online Library}
}

@article{Yang:diffusion:2023,
author = {Yang, Ling and Zhang, Zhilong and Song, Yang and Hong, Shenda and Xu, Runsheng and Zhao, Yue and Zhang, Wentao and Cui, Bin and Yang, Ming-Hsuan},
title = {Diffusion Models: A Comprehensive Survey of Methods and Applications},
year = {2023},
issue_date = {April 2024},
volume = {56},
number = {4},
doi = {10.1145/3626235},
journal = {ACM Comput. Surv.},
month = {nov},
articleno = {105},
numpages = {39}
}

@inproceedings{Jia:VPT:2022,
  title={Visual Prompt Tuning},
  author={Jia, Menglin and Tang, Luming and Chen, Bor-Chun and Cardie, Claire and Belongie, Serge and Hariharan, Bharath and Lim, Ser-Nam},
  booktitle={European Conference on Computer Vision (ECCV)},
  year={2022}
}

@ARTICLE{Cao:survey:2024,
  author={Cao, Hanqun and Tan, Cheng and Gao, Zhangyang and Xu, Yilun and Chen, Guangyong and Heng, Pheng-Ann and Li, Stan Z.},
  journal={IEEE Transactions on Knowledge and Data Engineering}, 
  title={A Survey on Generative Diffusion Models}, 
  year={2024},
  volume={},
  number={},
  pages={1-20},
  doi={10.1109/TKDE.2024.3361474}}

@article{Lian:LLMG:2024,
title={{LLM}-grounded Diffusion: Enhancing Prompt Understanding of Text-to-Image Diffusion Models with Large Language Models},
author={Long Lian and Boyi Li and Adam Yala and Trevor Darrell},
journal={Transactions on Machine Learning Research},
issn={2835-8856},
year={2024},
note={Featured Certification}
}

@inproceedings{Wang:SIMVLM:2022,
  title={{SIMVLM: Simple} Visual Language Model Pretraining with Weak Supervision},
  author={Wang, Zirui and Yu, Jiahui and Yu, Adams Wei and Dai, Zihang and Tsvetkov, Yulia and Cao, Yuan},
  booktitle={International Conference on Learning Representations (ICLR)},
  year={2022}
}

@INPROCEEDINGS{Xu:NeuralLift:2023,
  author={Xu, Dejia and Jiang, Yifan and Wang, Peihao and Fan, Zhiwen and Wang, Yi and Wang, Zhangyang},
  booktitle={2023 IEEE/CVF Conference on Computer Vision and Pattern Recognition (CVPR)}, 
  title={{NeuralLift-360: Lifting} an in-the-Wild 2D Photo to A {3D} Object with 360° Views}, 
  year={2023},
  volume={},
  number={},
  pages={4479-4489},
  doi={10.1109/CVPR52729.2023.00435}}

@inproceedings{gal:Image:2023,
title={An Image is Worth One Word: Personalizing Text-to-Image Generation using Textual Inversion},
author={Rinon Gal and Yuval Alaluf and Yuval Atzmon and Or Patashnik and Amit Haim Bermano and Gal Chechik and Daniel Cohen-or},
booktitle={The Eleventh International Conference on Learning Representations },
year={2023}
}

@INPROCEEDINGS{Melas:RealFusion:2023,
  author={Melas-Kyriazi, Luke and Laina, Iro and Rupprecht, Christian and Vedaldi, Andrea},
  booktitle={2023 IEEE/CVF Conference on Computer Vision and Pattern Recognition (CVPR)}, 
  title={RealFusion 360$^\circ$ Reconstruction of Any Object from a Single Image}, 
  year={2023},
  volume={},
  number={},
  pages={8446-8455},
  doi={10.1109/CVPR52729.2023.00816}}

@article{molad:dreamix:2023,
  author    = {Eyal Molad and Eliahu Horwitz and Dani Valevski and Alex Rav Acha and Yossi Matias and Yael Pritch and Yaniv Leviathan and Yedid Hoshen},
  title     = {Dreamix: Video Diffusion Models are General Video Editors},
  journal   = {arXiv: 2302.01329},
  year      = {2023},
}

@inproceedings{hong:cogvideo:2023,
title={CogVideo: Large-scale Pretraining for Text-to-Video Generation via Transformers},
author={Wenyi Hong and Ming Ding and Wendi Zheng and Xinghan Liu and Jie Tang},
booktitle={The Eleventh International Conference on Learning Representations },
year={2023}
}

@inproceedings{Liu:FETV:2023,
 author = {Liu, Yuanxin and Li, Lei and Ren, Shuhuai and Gao, Rundong and Li, Shicheng and Chen, Sishuo and Sun, Xu and Hou, Lu},
 booktitle = {Advances in Neural Information Processing Systems},
 pages = {62352--62387},
 title = {{FETV: A} Benchmark for Fine-Grained Evaluation of Open-Domain Text-to-Video Generation},
 volume = {36},
 year = {2023}
}

@article{XU2025103402,
title = {Integrating augmented reality and LLM for enhanced cognitive support in critical audio communications},
journal = {International Journal of Human-Computer Studies},
volume = {194},
pages = {103402},
year = {2025},
issn = {1071-5819},
doi = {https://doi.org/10.1016/j.ijhcs.2024.103402},
author = {Fang Xu and Tianyu Zhou and Tri Nguyen and Haohui Bao and Christine Lin and Jing Du},
}

@article{zhao2024clear,
  title     = {{CleAR}: Robust Context-Guided Generative Lighting Estimation for Mobile Augmented Reality},
  author    = {Zhao, Yiqin and Dasari, Mallesham and Guo, Tian},
  journal   = {arXiv preprint arXiv:2411.02179},
  year      = {2024}
}

@inproceedings{li2024learning,
author = {Li, Yongqi and Yang, Nan and Wang, Liang and Wei, Furu and Li, Wenjie},
title = {Learning to Rank in Generative Retrieval},
booktitle = {AAAI 2024},
year = {2024},
}

@article{sajja2024ai,
  title={Artificial Intelligence-Enabled Intelligent Assistant for Personalized and Adaptive Learning in Higher Education},
  author={Sajja, R. and Sermet, Y. and Cikmaz, M. and Cwiertny, D. and Demir, I.},
  journal={Information},
  volume={15},
  number={10},
  pages={596},
  year={2024},
  doi={10.3390/info15100596},
}

@inproceedings{Lee:design:2024,
author = {Lee, Mina and Gero, Katy Ilonka and Chung, John Joon Young and Shum, Simon Buckingham and Raheja, Vipul and Shen, Hua and Venugopalan, Subhashini and Wambsganss, Thiemo and Zhou, David and Alghamdi, Emad A. and August, Tal and Bhat, Avinash and Choksi, Madiha Zahrah and Dutta, Senjuti and Guo, Jin L.C. and Hoque, Md Naimul and Kim, Yewon and Knight, Simon and Neshaei, Seyed Parsa and Shibani, Antonette and Shrivastava, Disha and Shroff, Lila and Sergeyuk, Agnia and Stark, Jessi and Sterman, Sarah and Wang, Sitong and Bosselut, Antoine and Buschek, Daniel and Chang, Joseph Chee and Chen, Sherol and Kreminski, Max and Park, Joonsuk and Pea, Roy and Rho, Eugenia Ha Rim and Shen, Zejiang and Siangliulue, Pao},
title = {A Design Space for Intelligent and Interactive Writing Assistants},
year = {2024},
doi = {10.1145/3613904.3642697},
booktitle = {Proceedings of the 2024 CHI Conference on Human Factors in Computing Systems},
articleno = {1054},
numpages = {35}
}

@InProceedings{Chung_2024_CVPR,
    author    = {Chung, Jiwoo and Hyun, Sangeek and Heo, Jae-Pil},
    title     = {Style Injection in Diffusion: A Training-free Approach for Adapting Large-scale Diffusion Models for Style Transfer},
    booktitle = {Proceedings of the IEEE/CVF Conference on Computer Vision and Pattern Recognition (CVPR)},
    month     = {June},
    year      = {2024},
    pages     = {8795-8805}
}

@inproceedings{lin2024lowlight,
  title={Low-light Video Enhancement with Conditional Diffusion Models and Wavelet Interscale Attentions},
  author={Lin, R. and Sun, Q. and Anantrasirichai, N.},
  booktitle={Proceedings of the 21st ACM SIGGRAPH Conference on Visual Media Production},
  pages={1--10},
  year={2024}
}

@INPROCEEDINGS{Yu:Bidirectionally:2023,
  author={Yu, Wing-Yin and Po, Lai-Man and Cheung, Ray C.C. and Zhao, Yuzhi and Xue, Yu and Li, Kun},
  booktitle={2023 IEEE/CVF International Conference on Computer Vision (ICCV)}, 
  title={Bidirectionally Deformable Motion Modulation For Video-based Human Pose Transfer}, 
  year={2023},
  volume={},
  number={},
  pages={7468-7478},
  doi={10.1109/ICCV51070.2023.00690}}

@InProceedings{wang:disco:2024,
author    = {Wang, Tan and Li, Linjie and Lin, Kevin and Zhai, Yuanhao and Lin, Chung-Ching and Yang, Zhengyuan and Zhang, Hanwang and Liu, Zicheng and Wang, Lijuan},
    title     = {DisCo: Disentangled Control for Realistic Human Dance Generation},
    booktitle = {Proceedings of the IEEE/CVF Conference on Computer Vision and Pattern Recognition (CVPR)},
    month     = {June},
    year      = {2024},
    pages     = {9326-9336}
}

@InProceedings{Azadi:Make:2023,
    author    = {Azadi, Samaneh and Shah, Akbar and Hayes, Thomas and Parikh, Devi and Gupta, Sonal},
    title     = {{Make-An-Animation: Large}-Scale Text-conditional {3D} Human Motion Generation},
    booktitle = {Proceedings of the IEEE/CVF International Conference on Computer Vision (ICCV)},
    month     = {October},
    year      = {2023},
    pages     = {15039-15048}
}

@inproceedings{villegas:phenaki:2023,
title={{Phenaki: Variable} Length Video Generation from Open Domain Textual Descriptions},
author={Ruben Villegas and Mohammad Babaeizadeh and Pieter-Jan Kindermans and Hernan Moraldo and Han Zhang and Mohammad Taghi Saffar and Santiago Castro and Julius Kunze and Dumitru Erhan},
booktitle={International Conference on Learning Representations},
year={2023}
}

@article{wang:modelscope:2023,
  author    = {Jiuniu Wang and Hangjie Yuan and Dayou Chen and Yingya Zhang and Xiang Wang and Shiwei Zhang},
  title     = {ModelScope Text-to-Video Technical Report},
  journal   = {arXiv: 2308.06571},
  year      = {2023},
}

@inproceedings{wu:tune:2023,
  title={{Tune-a-video: One-shot} tuning of image diffusion models for text-to-video generation},
  author={Wu, Jay Zhangjie and Ge, Yixiao and Wang, Xintao and Lei, Stan Weixian and Gu, Yuchao and Shi, Yufei and Hsu, Wynne and Shan, Ying and Qie, Xiaohu and Shou, Mike Zheng},
  booktitle={Proceedings of the IEEE/CVF International Conference on Computer Vision},
  pages={7623--7633},
  year={2023}
}

@article{Bommasani2021FoundationModels,
title={On the Opportunities and Risks of Foundation Models},
author={Rishi Bommasani and Drew A. Hudson and Ehsan Adeli and Russ Altman and Simran Arora and others},
journal={arXiv:2108.07258 },
year={2021},
note={https://crfm.stanford.edu/assets/report.pdf}
}

@misc{CreativeIndustriesCouncil2021,
  author       = {Creative Industries Council},
  title        = {How CreaTech added 1+1 to make £981m},
  year         = {2021},
  note          = {\url{https://www.thecreativeindustries.co.uk/site-content/how-createch-added-one-plus-one-to-make-ps981m} (Accessed: 2025-01-10)}
}

@misc{Jeary2024,
  author       = {Lois Jeary and Devyani Gajjar},
  title        = {Artificial intelligence and new technology in creative industries},
  year         = {2024},
  month        = {October},
  day          = {7},
  publisher    = {UK Parliament Horizon Scanning Post},
  note          = {\url{https://post.parliament.uk/artificial-intelligence-and-new-technology-in-creative-industries/} (Accessed: 2025-01-10)},
}

@article{Azzarelli2024,
  author    = {Adrian Azzarelli and Nantheera Anantrasirichai and David R. Bull},
  title     = {Exploring Dynamic Novel View Synthesis Technologies for Cinematography},
  journal   = {arXiv: 2412.17532},
  year      = {2024},
}

@inproceedings{Zhang:DINet:2023,
  title={{DINet: Deformation} inpainting network for realistic face visually dubbing on high resolution video},
  author={Zhang, Zhimeng and Hu, Zhipeng and Deng, Wenjin and Fan, Changjie and Lv, Tangjie and Ding, Yu},
  booktitle={Proceedings of the AAAI Conference on Artificial Intelligence},
  volume={37},
  number={3},
  pages={3543--3551},
  year={2023}
}

@inproceedings{Alayrac:Flamingo:2022,
 author = {Alayrac, Jean-Baptiste and Donahue, Jeff and Luc, Pauline and Miech, Antoine and Barr, Iain and Hasson, Yana and Lenc, Karel and Mensch, Arthur and Millican, Katherine and Reynolds, Malcolm and Ring, Roman and Rutherford, Eliza and Cabi, Serkan and Han, Tengda and Gong, Zhitao and Samangooei, Sina and Monteiro, Marianne and Menick, Jacob L and Borgeaud, Sebastian and Brock, Andy and Nematzadeh, Aida and Sharifzadeh, Sahand and Bi\'{n}kowski, Miko\l aj and Barreira, Ricardo and Vinyals, Oriol and Zisserman, Andrew and Simonyan, Kar\'{e}n},
 booktitle = {Advances in Neural Information Processing Systems},
 pages = {23716--23736},
 title = {Flamingo: a Visual Language Model for Few-Shot Learning},
 volume = {35},
 year = {2022}
}

@inproceedings{esser:scaling:2024,
  title={Scaling Rectified Flow Transformers for High-Resolution Image Synthesis},
  author={Esser, Patrick and Kulal, Sumith and Blattmann, Andreas and Entezari, Rahim and M{\"u}ller, Jonas and others},
  booktitle={Proceedings of the 41 st International Conference on Machine Learning},
  year={2024}
}

@InProceedings{Brooks:InstructPix2Pix:2023,
    author    = {Brooks, Tim and Holynski, Aleksander and Efros, Alexei A.},
    title     = {{InstructPix2Pix: Learning} To Follow Image Editing Instructions},
    booktitle = {Proceedings of the IEEE/CVF Conference on Computer Vision and Pattern Recognition (CVPR)},
    month     = {June},
    year      = {2023},
    pages     = {18392-18402}
}

@InProceedings{Gandikota:Unified:2024,
    author    = {Gandikota, Rohit and Orgad, Hadas and Belinkov, Yonatan and Materzy\'nska, Joanna and Bau, David},
    title     = {Unified Concept Editing in Diffusion Models},
    booktitle = {Proceedings of the IEEE/CVF Winter Conference on Applications of Computer Vision (WACV)},
    month     = {January},
    year      = {2024},
    pages     = {5111-5120}
}

@INPROCEEDINGS{Choi:ILVR:2021,
  author={Choi, Jooyoung and Kim, Sungwon and Jeong, Yonghyun and Gwon, Youngjune and Yoon, Sungroh},
  booktitle={2021 IEEE/CVF International Conference on Computer Vision (ICCV)}, 
  title={{ILVR: Conditioning} Method for Denoising Diffusion Probabilistic Models}, 
  year={2021},
  volume={},
  number={},
  pages={14347-14356},
  doi={10.1109/ICCV48922.2021.01410}}

@inproceedings{Song:Score:2021,
  title={Score-Based Generative Modeling through Stochastic Differential Equations},
  author={Yang Song and Jascha Sohl-Dickstein and Diederik P Kingma and Abhishek Kumar and Stefano Ermon and Ben Poole},
  booktitle={International Conference on Learning Representations},
  year={2021},
}

@inproceedings{Dhariwal:Diffusion:2021,
title={Diffusion Models Beat {GAN}s on Image Synthesis},
author={Prafulla Dhariwal and Alexander Quinn Nichol},
booktitle={Advances in Neural Information Processing Systems},
editor={A. Beygelzimer and Y. Dauphin and P. Liang and J. Wortman Vaughan},
year={2021},
}

@INPROCEEDINGS{Rombach:LDM:2022,
  author={Rombach, Robin and Blattmann, Andreas and Lorenz, Dominik and Esser, Patrick and Ommer, Björn},
  booktitle={IEEE/CVF Conference on Computer Vision and Pattern Recognition (CVPR)}, 
  title={High-Resolution Image Synthesis with Latent Diffusion Models}, 
  year={2022},
  volume={},
  number={},
  pages={10674-10685},
  doi={10.1109/CVPR52688.2022.01042}}

@article{ravi2024sam2,
  title={SAM 2: Segment Anything in Images and Videos},
  author={Ravi, Nikhila and Gabeur, Valentin and Hu, Yuan-Ting and Hu, Ronghang and Ryali, Chaitanya and Ma, Tengyu and Khedr, Haitham and R{\"a}dle, Roman and Rolland, Chloe and Gustafson, Laura and Mintun, Eric and Pan, Junting and Alwala, Kalyan Vasudev and Carion, Nicolas and Wu, Chao-Yuan and Girshick, Ross and Doll{\'a}r, Piotr and Feichtenhofer, Christoph},
  journal={arXiv preprint arXiv:2408.00714},
  year={2024}
}

@inproceedings{zou2024deturb,
  title     = {{DeTurb: Atmospheric} Turbulence Mitigation with Deformable 3D Convolutions and 3D Swin Transformers},
  author    = {Zou, Z. and Anantrasirichai, N.},
  booktitle = {Proceedings of the Asian Conference on Computer Vision (ACCV)},
  year      = {2024}
}

@inproceedings{Wang2025,
  author    = {H. Wang and N. Anantrasirichai and F. Zhang and D. Bull},
  title     = {{UW-GS: Distractor}-Aware 3D Gaussian Splatting for Enhanced Underwater Scene Reconstruction},
  booktitle = {Proceedings of the IEEE/CVF Winter Conference on Applications of Computer Vision (WACV)},
  year      = {2025},
}

@inproceedings{lv2:detrs:2024,
      title={DETRs Beat YOLOs on Real-time Object Detection},
      author={Yian Zhao and Wenyu Lv and Shangliang Xu and Jinman Wei and Guanzhong Wang and Qingqing Dang and Yi Liu and Jie Chen},
      year={2024},
      booktitle = {IEEE Conference on Computer Vision and Pattern Recognition (CVPR)},
}

@INPROCEEDINGS{Li:Your:2023,
  author={Li, Alexander C. and Prabhudesai, Mihir and Duggal, Shivam and Brown, Ellis and Pathak, Deepak},
  booktitle={2023 IEEE/CVF International Conference on Computer Vision (ICCV)}, 
  title={Your Diffusion Model is Secretly a Zero-Shot Classifier}, 
  year={2023},
  volume={},
  number={},
  pages={2206-2217},
  doi={10.1109/ICCV51070.2023.00210}}

@InProceedings{Fang:Data:2024,
    author    = {Fang, Haoyang and Han, Boran and Zhang, Shuai and Zhou, Su and Hu, Cuixiong and Ye, Wen-Ming},
    title     = {Data Augmentation for Object Detection via Controllable Diffusion Models},
    booktitle = {Proceedings of the IEEE/CVF Winter Conference on Applications of Computer Vision (WACV)},
    month     = {January},
    year      = {2024},
    pages     = {1257-1266}
}

@inproceedings{Wu:datasetDM:2023,
 author = {Wu, Weijia and Zhao, Yuzhong and Chen, Hao and Gu, Yuchao and Zhao, Rui and He, Yefei and Zhou, Hong and Shou, Mike Zheng and Shen, Chunhua},
 booktitle = {Advances in Neural Information Processing Systems},
 pages = {54683--54695},
 publisher = {Curran Associates, Inc.},
 title = {{DatasetDM: Synthesizing} Data with Perception Annotations Using Diffusion Models},
 volume = {36},
 year = {2023}
}

@article{Luo:DiffusionTrack:2024, title={{DiffusionTrack: Diffusion} Model for Multi-Object Tracking}, volume={38}, note={https://ojs.aaai.org/index.php/AAAI/article/view/28192}, DOI={10.1609/aaai.v38i5.28192}, number={5}, journal={Proceedings of the AAAI Conference on Artificial Intelligence}, 
author={Luo, Run and Song, Zikai and Ma, Lintao and Wei, Jinlin and Yang, Wei and Yang, Min}, year={2024}, month={Mar.}, pages={3991-3999} }

@INPROCEEDINGS{Chen:DiffusionDet:2023,
  author={Chen, Shoufa and Sun, Peize and Song, Yibing and Luo, Ping},
  booktitle={IEEE/CVF International Conference on Computer Vision (ICCV)}, 
  title={DiffusionDet: Diffusion Model for Object Detection}, 
  year={2023},
  volume={},
  number={},
  pages={19773-19786},
  doi={10.1109/ICCV51070.2023.01816}}

@InProceedings{Zhang:DiffusionTracker:2024,
author="Zhang, Runqing
and Cai, Dunbo
and Qian, Ling
and Du, Yujian
and Lu, Huijun
and Zhang, Yijun",
title="{DiffusionTracker: Targets} Denoising Based on Diffusion Model for Visual Tracking",
booktitle="Pattern Recognition and Computer Vision",
year="2024",
pages="225--237",
}

@article{Karim:Current:2023,
title = {Current advances and future perspectives of image fusion: A comprehensive review},
journal = {Information Fusion},
volume = {90},
pages = {185-217},
year = {2023},
issn = {1566-2535},
doi = {https://doi.org/10.1016/j.inffus.2022.09.019},
author = {Shahid Karim and Geng Tong and Jinyang Li and Akeel Qadir and Umar Farooq and Yiting Yu},
}

@INPROCEEDINGS{Chen:Learning:2021,
  author={Chen, Yinbo and Liu, Sifei and Wang, Xiaolong},
  booktitle={2021 IEEE/CVF Conference on Computer Vision and Pattern Recognition (CVPR)}, 
  title={Learning Continuous Image Representation with Local Implicit Image Function}, 
  year={2021},
  volume={},
  number={},
  pages={8624-8634},
  doi={10.1109/CVPR46437.2021.00852}}

@ARTICLE{Zhang:visible:2023,
  author={Zhang, Xingchen and Demiris, Yiannis},
  journal={IEEE Transactions on Pattern Analysis and Machine Intelligence}, 
  title={Visible and Infrared Image Fusion Using Deep Learning}, 
  year={2023},
  volume={45},
  number={8},
  pages={10535-10554},
  doi={10.1109/TPAMI.2023.3261282}}

@ARTICLE{Rao:TGFuse:2023,
  author={Rao, Dongyu and Xu, Tianyang and Wu, Xiao-Jun},
  journal={IEEE Transactions on Image Processing}, 
  title={{TGFuse: An} Infrared and Visible Image Fusion Approach Based on Transformer and Generative Adversarial Network}, 
  year={2023},
  volume={},
  number={},
  pages={1-1},
  doi={10.1109/TIP.2023.3273451}}

@ARTICLE{Ma:SwinFusion:2022,
  author={Ma, Jiayi and Tang, Linfeng and Fan, Fan and Huang, Jun and Mei, Xiaoguang and Ma, Yong},
  journal={IEEE/CAA Journal of Automatica Sinica}, 
  title={{SwinFusion: Cross-domain} Long-range Learning for General Image Fusion via Swin Transformer}, 
  year={2022},
  volume={9},
  number={7},
  pages={1200-1217},
  doi={10.1109/JAS.2022.105686}}

@InProceedings{Liu:Multi:2023,
    author    = {Liu, Jinyuan and Liu, Zhu and Wu, Guanyao and Ma, Long and Liu, Risheng and Zhong, Wei and Luo, Zhongxuan and Fan, Xin},
    title     = {Multi-interactive Feature Learning and a Full-time Multi-modality Benchmark for Image Fusion and Segmentation},
    booktitle = {Proceedings of the IEEE/CVF International Conference on Computer Vision (ICCV)},
    month     = {October},
    year      = {2023},
    pages     = {8115-8124}
}

@InProceedings{Zhao:DDFM:2023,
    author    = {Zhao, Zixiang and Bai, Haowen and Zhu, Yuanzhi and Zhang, Jiangshe and Xu, Shuang and Zhang, Yulun and Zhang, Kai and Meng, Deyu and Timofte, Radu and Van Gool, Luc},
    title     = {{DDFM: Denoising} Diffusion Model for Multi-Modality Image Fusion},
    booktitle = {Proceedings of the IEEE/CVF International Conference on Computer Vision (ICCV)},
    month     = {October},
    year      = {2023},
    pages     = {8082-8093}
}

@article{LI2024102147,
title = {{CrossFuse: A} novel cross attention mechanism based infrared and visible image fusion approach},
journal = {Information Fusion},
volume = {103},
pages = {102147},
year = {2024},
issn = {1566-2535},
doi = {https://doi.org/10.1016/j.inffus.2023.102147},
author = {Hui Li and Xiao-Jun Wu},
}

@ARTICLE{10902142,
  author={He, Chunming and Shen, Yuqi and Fang, Chengyu and Xiao, Fengyang and Tang, Longxiang and Zhang, Yulun and Zuo, Wangmeng and Guo, Zhenhua and Li, Xiu},
  journal={IEEE Transactions on Pattern Analysis and Machine Intelligence}, 
  title={Diffusion Models in Low-Level Vision: A Survey}, 
  year={2025},
  volume={47},
  number={6},
  pages={4630-4651}
}

@inproceedings{zhan2024kfd,
  author    = {Yifan Zhan and Zhuoxiao Li and Muyao Niu and Zhihang Zhong and Shohei Nobuhara and Ko Nishino and Yinqiang Zheng},
  title     = {{KFD-NeRF}: Rethinking Dynamic NeRF with Kalman Filter},
  booktitle = {Proceedings of the European Conference on Computer Vision (ECCV)},
  year      = {2024},
}

@INPROCEEDINGS{9879447,
  author={Hu, Tao and Liu, Shu and Chen, Yilun and Shen, Tiancheng and Jia, Jiaya},
  booktitle={2022 IEEE/CVF Conference on Computer Vision and Pattern Recognition (CVPR)}, 
  title={{EfficientNeRF - Efficient} Neural Radiance Fields}, 
  year={2022},
  volume={},
  number={},
  pages={12892-12901},
  doi={10.1109/CVPR52688.2022.01256}}

@InProceedings{Tang_2024_CVPR,
    author    = {Tang, Xiao and Yang, Min and Sun, Penghui and Li, Hui and Dai, Yuchao and Zhu, Feng and Lee, Hojae},
    title     = {PaReNeRF: Toward Fast Large-scale Dynamic NeRF with Patch-based Reference},
    booktitle = {Proceedings of the IEEE/CVF Conference on Computer Vision and Pattern Recognition (CVPR)},
    month     = {June},
    year      = {2024},
    pages     = {5428-5438}
}

@InProceedings{Liu_2024_CVPR,
    author    = {Liu, Jia-Wei and Cao, Yan-Pei and Wu, Jay Zhangjie and Mao, Weijia and Gu, Yuchao and Zhao, Rui and Keppo, Jussi and Shan, Ying and Shou, Mike Zheng},
    title     = {DynVideo-E: Harnessing Dynamic NeRF for Large-Scale Motion- and View-Change Human-Centric Video Editing},
    booktitle = {Proceedings of the IEEE/CVF Conference on Computer Vision and Pattern Recognition (CVPR)},
    month     = {June},
    year      = {2024},
    pages     = {7664-7674}
}

@article{im2025gate3d,
  title     = {{GATE3D: Generalized Attention-based Task-synergized Estimation in 3D}},
  author    = {Eunsoo Im and Changhyun Jee and Jung Kwon Lee},
  journal   = {arXiv preprint arXiv:2504.11014},
  year      = {2025}
}

@ARTICLE{10637966,
  author={Song, Ziying and Liu, Lin and Jia, Feiyang and Luo, Yadan and Jia, Caiyan and Zhang, Guoxin and Yang, Lei and Wang, Li},
  journal={IEEE Transactions on Intelligent Transportation Systems}, 
  title={Robustness-Aware 3D Object Detection in Autonomous Driving: A Review and Outlook}, 
  year={2024},
  volume={25},
  number={11},
  pages={15407-15436},
  doi={10.1109/TITS.2024.3439557}}

@InProceedings{Xie:DiffusionTrack:2024,
    author    = {Xie, Fei and Wang, Zhongdao and Ma, Chao},
    title     = {{DiffusionTrack: Point} Set Diffusion Model for Visual Object Tracking},
    booktitle = {Proceedings of the IEEE/CVF Conference on Computer Vision and Pattern Recognition (CVPR)},
    month     = {June},
    year      = {2024},
    pages     = {19113-19124}
}

@inproceedings{wang:yolov10:2024,
  title={{YOLOv10: Real}-Time End-to-End Object Detection},
  author={Wang, Ao and Chen, Hui and Liu, Lihao and Chen, Kai and Lin, Zijia and Han, Jungong and Ding, Guiguang},
  booktitle = {Advances in Neural Information Processing Systems},
  year={2024}
}

@inproceedings{Cen:Segment:2023,
 author = {Cen, Jiazhong and Zhou, Zanwei and Fang, Jiemin and yang, chen and Shen, Wei and Xie, Lingxi and Jiang, Dongsheng and ZHANG, XIAOPENG and Tian, Qi},
 booktitle = {Advances in Neural Information Processing Systems},
 pages = {25971--25990},
 title = {Segment Anything in {3D with NeRFs}},
 volume = {36},
 year = {2023}
}

@ARTICLE{Tous:Lester:2024,
author = {Ruben Tous},
title = {{Lester: Rotoscope} animation through video object segmentation and tracking},
year = {2024},
journal = {Algorithms },
volume ={17},
number = {8},
doi = {https://doi.org/10.3390/a17080330}
}

@inproceedings{radford2021learning,
  title={Learning Transferable Visual Models From Natural Language Supervision},
  author={Alec Radford and Jong Wook Kim and Chris Hallacy and Aditya Ramesh and Gabriel Goh and Sandhini Agarwal and Girish Sastry and Amanda Askell and Pamela Mishkin and Jack Clark and Gretchen Krueger and Ilya Sutskever},
  booktitle={International Conference on Machine Learning},
  year={2021}
}

@inproceedings{wei2024vary,
  title={{Vary: Scaling} up the Vision Vocabulary for Large Vision-Language Models},
  author={Wei, Haoran and Kong, Lingyu and Chen, Jinyue and Zhao, Liang and Ge, Zheng and Yang, Jinrong and Sun, Jianjian and Han, Chunrui and Zhang, Xiangyu},
  booktitle={European Conference on Computer Vision},
  year={2024}
}

@article{guo2024liveportrait,
    title   = {{LivePortrait: Efficient} Portrait Animation with Stitching and Retargeting Control},
    author  = {Guo, Jianzhu and Zhang, Dingyun and Liu, Xiaoqiang and Zhong, Zhizhou and Zhang, Yuan and Wan, Pengfei and Zhang, Di},
    journal = {arXiv preprint arXiv:2407.03168},
    year    = {2024}
}

@InProceedings{Dickstein:Deep:2015,
  title = 	 {Deep Unsupervised Learning using Nonequilibrium Thermodynamics},
  author = 	 {Sohl-Dickstein, Jascha and Weiss, Eric and Maheswaranathan, Niru and Ganguli, Surya},
  booktitle = 	 {Proceedings of the 32nd International Conference on Machine Learning},
  pages = 	 {2256--2265},
  year = 	 {2015},
  volume = 	 {37},
  month = 	 {07--09 Jul},
}

@INPROCEEDINGS{Ho:DDPM:2020,
  title={Denoising Diffusion Probabilistic Models},
  author={Jonathan Ho and Ajay Jain and Pieter Abbeel},
  year={2020},
booktitle = {Advances in Neural Information Processing Systems },
pages = {6840–6851.},
}

@InProceedings{Xu:Video:2023,
    author    = {Xu, Jiaqi and Hu, Xiaowei and Zhu, Lei and Dou, Qi and Dai, Jifeng and Qiao, Yu and Heng, Pheng-Ann},
    title     = {Video Dehazing via a Multi-Range Temporal Alignment Network With Physical Prior},
    booktitle = {Proceedings of the IEEE/CVF Conference on Computer Vision and Pattern Recognition (CVPR)},
    month     = {June},
    year      = {2023},
    pages     = {18053-18062}
}

@inproceedings{mao:single:2022,
  title={Single Frame Atmospheric Turbulence Mitigation: A Benchmark Study and a New Physics-Inspired Transformer Model},
  author={Mao, Zhiyuan and Jaiswal, Ajay and Wang, Zhangyang and Chan, Stanley H.},
  booktitle={European Conference on Computer Vision (ECCV)},
  year={2022}
}

@article{Atmospheric:2023,
title = {Atmospheric turbulence removal with complex-valued convolutional neural network},
journal = {Pattern Recognition Letters},
volume = {171},
pages = {69-75},
year = {2023},
doi = {https://doi.org/10.1016/j.patrec.2023.05.017},
author = {Nantheera Anantrasirichai},
}

@InProceedings{Nair:AT-DDPM:2023,
    author    = {Nair, Nithin Gopalakrishnan and Mei, Kangfu and Patel, Vishal M.},
    title     = {{AT-DDPM: Restoring} Faces Degraded by Atmospheric Turbulence Using Denoising Diffusion Probabilistic Models},
    booktitle = {Proceedings of the IEEE/CVF Winter Conference on Applications of Computer Vision (WACV)},
    month     = {January},
    year      = {2023},
    pages     = {3434-3443}
}

@ARTICLE{Wu:brief:2023,
  author={Wu, Tianyu and He, Shizhu and Liu, Jingping and Sun, Siqi and Liu, Kang and Han, Qing-Long and Tang, Yang},
  journal={IEEE/CAA Journal of Automatica Sinica}, 
  title={A Brief Overview of {ChatGPT}: The History, Status Quo and Potential Future Development}, 
  year={2023},
  volume={10},
  number={5},
  pages={1122-1136},
  doi={10.1109/JAS.2023.123618}}

@ARTICLE{Zhang:Image:2024,
  author={Zhang, Xingguang and Mao, Zhiyuan and Chimitt, Nicholas and Chan, Stanley H.},
  journal={IEEE Transactions on Computational Imaging}, 
  title={Imaging Through the Atmosphere Using Turbulence Mitigation Transformer}, 
  year={2024},
  volume={10},
  number={},
  pages={115-128},
  doi={10.1109/TCI.2024.3354421}}

@InProceedings{Jiang:NeRT:2023,
    author    = {Jiang, Weiyun and Boominathan, Vivek and Veeraraghavan, Ashok},
    title     = {{NeRT: Implicit} Neural Representations for Unsupervised Atmospheric Turbulence Mitigation},
    booktitle = {Proceedings of the IEEE/CVF Conference on Computer Vision and Pattern Recognition (CVPR) Workshops},
    month     = {June},
    year      = {2023},
    pages     = {4236-4243}
}

@InProceedings{Jaiswal:Physics:2023,
    author    = {Jaiswal, Ajay and Zhang, Xingguang and Chan, Stanley H. and Wang, Zhangyang},
    title     = {Physics-Driven Turbulence Image Restoration with Stochastic Refinement},
    booktitle = {Proceedings of the IEEE/CVF International Conference on Computer Vision (ICCV)},
    month     = {October},
    year      = {2023},
    pages     = {12170-12181}
}

@article{Reed:Generalist:2022,
title={A Generalist Agent},
author={Scott Reed and Konrad Zolna and Emilio Parisotto and Sergio G{\'o}mez Colmenarejo and Alexander Novikov and Gabriel Barth-maron and Mai Gim{\'e}nez and Yury Sulsky and Jackie Kay and Jost Tobias Springenberg and Tom Eccles and Jake Bruce and Ali Razavi and Ashley Edwards and Nicolas Heess and Yutian Chen and Raia Hadsell and Oriol Vinyals and Mahyar Bordbar and Nando de Freitas},
journal={Transactions on Machine Learning Research},
issn={2835-8856},
year={2022},
}

@book{Bull:intelligent:2021,
  title={Intelligent image and video compression: communicating pictures},
  author={Bull, David and Zhang, Fan},
  year={2021},
  publisher={Academic Press}
}

@ARTICLE {Selva:video:2023,
author = {J. Selva and A. S. Johansen and S. Escalera and K. Nasrollahi and T. B. Moeslund and A. Clapes},
journal = {IEEE Transactions on Pattern Analysis and Machine Intelligence},
title = {Video Transformers: A Survey},
year = {2023},
volume = {45},
number = {11},
issn = {1939-3539},
pages = {12922-12943},
doi = {10.1109/TPAMI.2023.3243465},
month = {nov}
}

@article{Khan:Transformers:2022,
author = {Khan, Salman and Naseer, Muzammal and Hayat, Munawar and Zamir, Syed Waqas and Khan, Fahad Shahbaz and Shah, Mubarak},
title = {Transformers in Vision: A Survey},
journal={ACM computing surveys (CSUR)},
year = {2022},
volume = {54},
number = {10s},
issn = {0360-0300},
note = {https://doi.org/10.1145/3505244},
doi = {10.1145/3505244},
month = {sep},
articleno = {200},
numpages = {41},
}

@inproceedings{Oord:Neural:2017,
author = {van den Oord, Aaron and Vinyals, Oriol and Kavukcuoglu, Koray},
title = {Neural discrete representation learning},
year = {2017},
booktitle = {Proceedings of the 31st International Conference on Neural Information Processing Systems},
pages = {6309–6318},
numpages = {10}
}

@INPROCEEDINGS{Moliner:solving:2023,
  author={Moliner, Eloi and Lehtinen, Jaakko and Välimäki, Vesa},
  booktitle={IEEE International Conference on Acoustics, Speech and Signal Processing (ICASSP)}, 
  title={Solving Audio Inverse Problems with a Diffusion Model}, 
  year={2023},
  volume={},
  number={},
  pages={1-5},
  doi={10.1109/ICASSP49357.2023.10095637}}

@ARTICLE{Yu:DBT:2022,
  author={Yu, Guochen and Li, Andong and Wang, Hui and Wang, Yutian and Ke, Yuxuan and Zheng, Chengshi},
  journal={IEEE/ACM Transactions on Audio, Speech, and Language Processing}, 
  title={DBT-Net: Dual-Branch Federative Magnitude and Phase Estimation With Attention-in-Attention Transformer for Monaural Speech Enhancement}, 
  year={2022},
  volume={30},
  number={},
  pages={2629-2644},
  doi={10.1109/TASLP.2022.3195112}}

@INPROCEEDINGS{Wang:One:2022,
  author={Wang, Zhichao and Xie, Qicong and Li, Tao and Du, Hongqiang and Xie, Lei and Zhu, Pengcheng and Bi, Mengxiao},
  booktitle={IEEE International Conference on Acoustics, Speech and Signal Processing (ICASSP)}, 
  title={One-Shot Voice Conversion For Style Transfer Based On Speaker Adaptation}, 
  year={2022},
  pages={6792-6796},
  doi={10.1109/ICASSP43922.2022.9746405}}

@inproceedings{Huang:GenerSpeech:2022,
  title={{GenerSpeech: Towards} Style Transfer for Generalizable Out-Of-Domain Text-to-Speech},
  author={Huang, Rongjie and Ren, Yi and Liu, Jinglin and Cui, Chenye and Zhao, Zhou},
  booktitle={Advances in Neural Information Processing Systems},
year={2022}
}

@ARTICLE{Yang:Diffsound:2023,
  author={Yang, Dongchao and Yu, Jianwei and Wang, Helin and Wang, Wen and Weng, Chao and Zou, Yuexian and Yu, Dong},
  journal={IEEE/ACM Transactions on Audio, Speech, and Language Processing}, 
  title={Diffsound: Discrete Diffusion Model for Text-to-Sound Generation}, 
  year={2023},
  volume={31},
  number={},
  pages={1720-1733},
  doi={10.1109/TASLP.2023.3268730}}

@inproceedings{Li:Diffusion:2022,
 author = {Li, Xiang and Thickstun, John and Gulrajani, Ishaan and Liang, Percy S and Hashimoto, Tatsunori B},
 booktitle = {Advances in Neural Information Processing Systems},
 pages = {4328--4343},
 title = {Diffusion-LM Improves Controllable Text Generation},
 volume = {35},
 year = {2022}
}

@INPROCEEDINGS{Neimark:video:2021,
  author={Neimark, Daniel and Bar, Omri and Zohar, Maya and Asselmann, Dotan},
  booktitle={2021 IEEE/CVF International Conference on Computer Vision Workshops (ICCVW)}, 
  title={Video Transformer Network}, 
  year={2021},
  volume={},
  number={},
  pages={3156-3165},
  doi={10.1109/ICCVW54120.2021.00355}}

@inproceedings{Fan:SUNet:2022,
  title={{SUNet: swin} transformer UNet for image denoising},
  author={Fan, Chi-Mao and Liu, Tsung-Jung and Liu, Kuan-Hsien},
  booktitle={2022 IEEE International Symposium on Circuits and Systems (ISCAS)},
  pages={2333--2337},
  year={2022},
  organization={IEEE}
}

@inproceedings{Liu:Swin:2021,
  title={{Swin Transformer: Hierarchical} Vision Transformer using Shifted Windows},
  author={Liu, Ze and Lin, Yutong and Cao, Yue and Hu, Han and Wei, Yixuan and Zhang, Zheng and Lin, Stephen and Guo, Baining},
  booktitle={Proceedings of the IEEE/CVF International Conference on Computer Vision (ICCV)},
  year={2021}
}

@article{bochkovskiy2020yolov4,
  title={{YOLOv4: Optimal} speed and accuracy of object detection},
  author={Bochkovskiy, A. and Wang, C.-Y. and Liao, H.-Y. M.},
  journal={arXiv preprint arXiv:2004.10934},
  year={2020}
}

@inproceedings{Zhu:Deformable:2021,
  title={{Deformable DETR: Deformable} Transformers for End-to-End Object Detection},
  author={Zhu, Xizhou and Su, Weijie and Lu, Lewei and Li, Bin and Wang, Xiaogang and Dai, Jifeng},
  booktitle={International Conference on Learning Representations (ICLR)},
  year={2021}
}

@article{ZHOU:Bridging:2025,
title = {Bridging the metrics gap in image style transfer: A comprehensive survey of models and criteria},
journal = {Neurocomputing},
volume = {624},
pages = {129430},
year = {2025},
issn = {0925-2312},
doi = {https://doi.org/10.1016/j.neucom.2025.129430},
author = {Xiaotong Zhou and Yuhui Zheng and Junming Yang}
}

@INPROCEEDINGS{Liang:SwinIR:2021,
  author={Liang, Jingyun and Cao, Jiezhang and Sun, Guolei and Zhang, Kai and Van Gool, Luc and Timofte, Radu},
  booktitle={2021 IEEE/CVF International Conference on Computer Vision Workshops (ICCVW)}, 
  title={SwinIR: Image Restoration Using Swin Transformer}, 
  year={2021},
  volume={},
  number={},
  pages={1833-1844},
  doi={10.1109/ICCVW54120.2021.00210}}

@INPROCEEDINGS{Liu:Swinv2:2022,
  author={Liu, Ze and Hu, Han and Lin, Yutong and Yao, Zhuliang and Xie, Zhenda and Wei, Yixuan and Ning, Jia and Cao, Yue and Zhang, Zheng and Dong, Li and Wei, Furu and Guo, Baining},
  booktitle={2022 IEEE/CVF Conference on Computer Vision and Pattern Recognition (CVPR)}, 
  title={Swin Transformer V2: Scaling Up Capacity and Resolution}, 
  year={2022},
  volume={},
  number={},
  pages={11999-12009},
  doi={10.1109/CVPR52688.2022.01170}}

@article{CHUA:AI:2023,
title = {AI-enabled investment advice: Will users buy it?},
journal = {Computers in Human Behavior},
volume = {138},
pages = {107481},
year = {2023},
doi = {https://doi.org/10.1016/j.chb.2022.107481},
author = {Alton Y.K. Chua and Anjan Pal and Snehasish Banerjee},
}

@techreport{brynjolfsson:generative:2023,
  title={Generative {AI} at Work},
  author={Brynjolfsson, Erik and Li, Danielle and Raymond, Lindsey},
institution={National Bureau of Economic Research},
  year={2023},
  month={April}
}

@article{King:Sasha:2024,
author = {King, Evan and Yu, Haoxiang and Lee, Sangsu and Julien, Christine},
title = {{Sasha: Creative} Goal-Oriented Reasoning in Smart Homes with Large Language Models},
year = {2024},
volume = {8},
number = {1},
doi = {10.1145/3643505},
journal = {Proc. ACM Interact. Mob. Wearable Ubiquitous Technol.},
month = {mar},
}

@inproceedings{Huang:MonoDTR:2022,
    author = {Kuan-Chih Huang and Tsung-Han Wu and Hung-Ting Su and Winston H. Hsu},
    title = {{MonoDTR: Monocular} {3D} Object Detection with Depth-Aware Transformer},
    booktitle = {Proceedings of the IEEE/CVF Conference on Computer Vision and Pattern Recognition (CVPR)},
    year = {2022}    
}

@inproceedings{yang:Holodeck:2024,
      title={{Holodeck: Language} Guided Generation of {3D} Embodied {AI} Environments}, 
      author={Yue Yang and Fan-Yun Sun and Luca Weihs and Eli VanderBilt and Alvaro Herrasti and Winson Han and Jiajun Wu and Nick Haber and Ranjay Krishna and Lingjie Liu and Chris Callison-Burch and Mark Yatskar and Aniruddha Kembhavi and Christopher Clark},
    booktitle = {Proceedings of the IEEE/CVF Conference on Computer Vision and Pattern Recognition (CVPR)},
    year = {2024}   
}

@inproceedings{deitke:Objaverse:2023,
  title={{Objaverse-XL: A} Universe of 10M+ {3D} Objects},
  author={Deitke, Matt and Liu, Ruoshi and Wallingford, Matthew and Ngo, Huong and Michel, Oscar and Kusupati, Aditya and Fan, Alan and Laforte, Christian and Voleti, Vikram and Gadre, Samir Yitzhak and VanderBilt, Eli and Kembhavi, Aniruddha and Vondrick, Carl and Gkioxari, Georgia and Ehsani, Kiana and Schmidt, Ludwig and Farhadi, Ali},
  booktitle={Advances in Neural Information Processing Systems (NeurIPS)},
  year={2023},
  month={7},
  day={11}
}

@inproceedings{Youk:FMA:2024,
  author    = {Geunhyuk Youk and Jihyong Oh and Munchnote Kim},
  title     = {{FMA-Net: Flow}-Guided Dynamic Filtering and Iterative Feature Refinement with Multi-Attention for Joint Video Super-Resolution and Deblurring},
  booktitle = {Proceedings of the IEEE/CVF Conference on Computer Vision and Pattern Recognition (CVPR)},
  year      = {2024},
 }

@inproceedings{Yin:CLE:2023,
author = {Yin, Yuyang and Xu, Dejia and Tan, Chuangchuang and Liu, Ping and Zhao, Yao and Wei, Yunchao},
title = {{CLE Diffusion: Controllable} Light Enhancement Diffusion Model},
year = {2023},
doi = {10.1145/3581783.3612145},
booktitle = {Proceedings of the 31st ACM International Conference on Multimedia},
pages = {8145–8156}
}

@inproceedings{li:GRL:2023,
  title={Efficient and Explicit Modelling of Image Hierarchies for Image Restoration},
  author={Yawei Li and Yuchen Fan and Xiaoyu Xiang and Denis Demandolx, Rakesh Ranjan and Radu Timofte and and Luc Van Gool},
  booktitle={Proceedings of the IEEE Conference on Computer Vision and Pattern Recognition},
  year={2023}
}

@inproceedings{Vaswani:attention:2017,
 author = {Vaswani, Ashish and Shazeer, Noam and Parmar, Niki and Uszkoreit, Jakob and Jones, Llion and Gomez, Aidan N and Kaiser, \L ukasz and Polosukhin, Illia},
 booktitle = {Advances in Neural Information Processing Systems},
 pages = {},
 title = {Attention is All you Need},
 volume = {30},
 year = {2017}
}

@ARTICLE{Liang:VRT:2024,
  author={Liang, Jingyun and Cao, Jiezhang and Fan, Yuchen and Zhang, Kai and Ranjan, Rakesh and Li, Yawei and Timofte, Radu and Van Gool, Luc},
  journal={IEEE Transactions on Image Processing}, 
  title={VRT: A Video Restoration Transformer}, 
  year={2024},
  volume={33},
  number={},
  pages={2171-2182},
  doi={10.1109/TIP.2024.3372454}}

@InProceedings{Lu:Transformer:2022,
    author    = {Lu, Zhisheng and Li, Juncheng and Liu, Hong and Huang, Chaoyan and Zhang, Linlin and Zeng, Tieyong},
    title     = {Transformer for Single Image Super-Resolution},
    booktitle = {Proceedings of the IEEE/CVF Conference on Computer Vision and Pattern Recognition (CVPR) Workshops},
    month     = {June},
    year      = {2022},
    pages     = {457-466}
}

@INPROCEEDINGS{Conde:Efficient:2023,
  author={Conde, Marcos V. and Zamfir, Eduard and Timofte, Radu and Motilla, Daniel and others},
  booktitle={2023 IEEE/CVF Conference on Computer Vision and Pattern Recognition Workshops (CVPRW)}, 
  title={Efficient Deep Models for Real-Time 4K Image Super-Resolution. NTIRE 2023 Benchmark and Report}, 
  year={2023},
  volume={},
  number={},
  pages={1495-1521},
  doi={10.1109/CVPRW59228.2023.00154}}

@InProceedings{Chen:Activating:2023,
    author    = {Chen, Xiangyu and Wang, Xintao and Zhou, Jiantao and Qiao, Yu and Dong, Chao},
    title     = {Activating More Pixels in Image Super-Resolution Transformer},
    booktitle = {Proceedings of the IEEE/CVF Conference on Computer Vision and Pattern Recognition (CVPR)},
    month     = {June},
    year      = {2023},
    pages     = {22367-22377}
}

@INPROCEEDINGS {Bilecen:Bicubic:2023,
author = {B. Bilecen and M. Ayazoglu},
booktitle = {2023 IEEE/CVF Conference on Computer Vision and Pattern Recognition Workshops (CVPRW)},
title = {{Bicubic++: Slim,} Slimmer, Slimmest Designing an Industry-Grade Super-Resolution Network},
year = {2023},
volume = {},
issn = {},
pages = {1623-1632},
doi = {10.1109/CVPRW59228.2023.00164},
month = {jun}
}

@inproceedings{liang:RVRT:2022,
  title={Recurrent video restoration transformer with guided deformable attention},
  author={Liang, J. and Fan, Y. and Xiang, X. and Ranjan, R. and Ilg, E. and Green, S. and Cao, J. and Zhang, K. and Timofte, R. and Gool, LV},
  booktitle={Advances in Neural Information Processing Systems},
  year={2022}
}

@InProceedings{Kingma:auto:2014,
  Title                    = {Auto-Encoding Variational Bayes},
  Author                   = { D.P. Kingma and M. Welling},
  Booktitle                = {International Conference on Learning Representations},
  Year                     = {2014}
}

@INPROCEEDINGS{STMAD, 
author={P. V. {Vu} and C. T. {Vu} and D. M. {Chandler}}, 
booktitle={IEEE ICIP}, 
title={A spatiotemporal most-apparent-distortion model for video quality assessment}, 
year={2011}, 
pages={2505-2508}, 
month={Sep.},}

@ARTICLE{Bovik_SSIM, 
author={ Z. {Wang} and A. C. {Bovik} and H. R. {Sheikh} and E. P. {Simoncelli}}, 
journal={IEEE Trans. Image Process.}, 
title={Image quality assessment: from error visibility to structural similarity},
year={2004}, 
volume={13}, 
number={4}, 
pages={600-612}, 
month={April},}

@inproceedings{Bovik_MSSSIM,
  title={Multi-scale structural similarity for image quality assessment},
  author={Wang, Z. and Simoncelli, E. P. and Bovik, A. C.},
  booktitle={Asilomar Conf. Signals Syst. Comput.},
  pages={1398--1402},
  year={2003},
}

@MISC{VMAFblog,
  title = {{The NETFLIX tech blog: Toward a practical perceptual video quality metric}}, 
  author={Z. Li and A. Aaron and I. Katsavounidis and A. Moorthy and M. Manohara},
  howpublished= {\url{http://techblog.netflix.com/2016/06/toward-practical-perceptual-video.html},
  note = {[Online; accessed 2018-08-04]},}
}

@inproceedings{Lin:SPATIO:2024,
  title={A SPATIO-TEMPORAL ALIGNED SUNET MODEL FOR LOW-LIGHT VIDEO ENHANCEMENT},
  author={Lin, Ruirui and Anantrasirichai, Nantheera and Malyugina, Alexandra and Bull, David},
  booktitle={IEEE International Conference on Image Processing},
  year={2024},
}

@article{huang2025bayesian,
  author    = {Guoxi Huang and Nantheera Anantrasirichai and Fei Ye and Zipeng Qi and RuiRui Lin and Qirui Yang and David Bull},
  title     = {Bayesian Neural Networks for One-to-Many Mapping in Image Enhancement},
  journal   = {arXiv preprint},
  volume    = {arXiv:2501.14265},
  year      = {2025},
  month     = {January},
  day       = {24},
}

@InProceedings{Zhou:LEDNet:2022,
author="Zhou, Shangchen
and Li, Chongyi
and Change Loy, Chen",
editor="Avidan, Shai
and Brostow, Gabriel
and Ciss{\'e}, Moustapha
and Farinella, Giovanni Maria
and Hassner, Tal",
title="LEDNet: Joint Low-Light Enhancement and Deblurring in the Dark",
booktitle="Computer Vision -- ECCV 2022",
year="2022",
pages="573--589",
}

@INPROCEEDINGS{Xu:SNR:2022,
  author={Xu, Xiaogang and Wang, Ruixing and Fu, Chi-Wing and Jia, Jiaya},
  booktitle={2022 IEEE/CVF Conference on Computer Vision and Pattern Recognition (CVPR)}, 
  title={SNR-Aware Low-light Image Enhancement}, 
  year={2022},
  volume={},
  number={},
  pages={17693-17703},
  doi={10.1109/CVPR52688.2022.01719}}

@inproceedings{Wang:Ultra:2023,
  title={Ultra-high-definition low-light image enhancement: A benchmark and transformer-based method},
  author={Wang, Tao and Zhang, Kaihao and Shen, Tianrun and Luo, Wenhan and Stenger, Bjorn and Lu, Tong},
  booktitle={Proceedings of the AAAI Conference on Artificial Intelligence},
  volume={37},
  number={3},
  pages={2654--2662},
  year={2023}
}

@inproceedings{HOU:Global:2023,
 author = {HOU, Jinhui and Zhu, Zhiyu and Hou, Junhui and LIU, Hui and Zeng, Huanqiang and Yuan, Hui},
 booktitle = {Advances in Neural Information Processing Systems},
 pages = {79734--79747},
 title = {Global Structure-Aware Diffusion Process for Low-light Image Enhancement},
 volume = {36},
 year = {2023}
}

@ARTICLE{Kugarajeevan:Transformers:2023,
  author={Kugarajeevan, Janani and Kokul, Thanikasalam and Ramanan, Amirthalingam and Fernando, Subha},
  journal={IEEE Access}, 
  title={Transformers in Single Object Tracking: An Experimental Survey}, 
  year={2023},
  volume={11},
  number={},
  pages={80297-80326},
  doi={10.1109/ACCESS.2023.3298440}}

@INPROCEEDINGS{Mayer:Transforming:2022,
  author={Mayer, Christoph and Danelljan, Martin and Bhat, Goutam and Paul, Matthieu and Paudel, Danda Pani and Yu, Fisher and Van Gool, Luc},
  booktitle={2022 IEEE/CVF Conference on Computer Vision and Pattern Recognition (CVPR)}, 
  title={Transforming Model Prediction for Tracking}, 
  year={2022},
  volume={},
  number={},
  pages={8721-8730},
  doi={10.1109/CVPR52688.2022.00853}}

@InProceedings{Yang:Implicit:2023,
    author    = {Yang, Shuzhou and Ding, Moxuan and Wu, Yanmin and Li, Zihan and Zhang, Jian},
    title     = {Implicit Neural Representation for Cooperative Low-light Image Enhancement},
    booktitle = {Proceedings of the IEEE/CVF International Conference on Computer Vision (ICCV)},
    month     = {October},
    year      = {2023},
    pages     = {12918-12927}
}

@InProceedings{Yi:Diff:2023,
    author    = {Yi, Xunpeng and Xu, Han and Zhang, Hao and Tang, Linfeng and Ma, Jiayi},
    title     = {Diff-Retinex: Rethinking Low-light Image Enhancement with A Generative Diffusion Model},
    booktitle = {Proceedings of the IEEE/CVF International Conference on Computer Vision (ICCV)},
    month     = {October},
    year      = {2023},
    pages     = {12302-12311}
}

@article{Jiang:Low:2023,
  title={Low-light image enhancement with wavelet-based diffusion models},
  author={Jiang, Hai and Luo, Ao and Fan, Haoqiang and Han, Songchen and Liu, Shuaicheng},
  journal={ACM Transactions on Graphics (TOG)},
  volume={42},
  number={6},
  pages={1--14},
  year={2023}
}

@InProceedings{Fei:Generative:2023,
    author    = {Fei, Ben and Lyu, Zhaoyang and Pan, Liang and Zhang, Junzhe and Yang, Weidong and Luo, Tianyue and Zhang, Bo and Dai, Bo},
    title     = {Generative Diffusion Prior for Unified Image Restoration and Enhancement},
    booktitle = {Proceedings of the IEEE/CVF Conference on Computer Vision and Pattern Recognition (CVPR)},
    month     = {June},
    year      = {2023},
    pages     = {9935-9946}
}

@article{anantrasirichai:BVI:2024,
  title={{BVI-Lowlight: Fully} Registered Benchmark Dataset for Low-Light Video Enhancement},
  author={Anantrasirichai, Nantheera and Lin, Ruirui and Malyugina, Alexandra and Bull, David},
  journal={arXiv preprint arXiv:2402.01970},
  year={2024}
}

@article{Kim:Controllable:2024,
title = {Controllable Style Transfer via Test-time Training of Implicit Neural Representation},
journal = {Pattern Recognition},
volume = {146},
pages = {109988},
year = {2024},
doi = {https://doi.org/10.1016/j.patcog.2023.109988},
author = {Sunwoo Kim and Youngjo Min and Younghun Jung and Seungryong Kim},
}

@inproceedings{Moon:generalizable:2023,
  title={Generalizable Style Transfer for Implicit Neural Representation},
  author={Jaeho Moon and Taehong Moon and Wonyong Seo},
  booktitle={International Conference on Learning Representations (ICLR)},
  year={2023}
}

@article{chen2025ntire,
  title   = {NTIRE 2025 Challenge on Image Super-Resolution (×4): Methods and Results},
  author  = {Zheng Chen and Kai Liu and Jue Gong and Jingkai Wang and others},
  journal = {arXiv preprint arXiv:2504.14582},
  year    = {2025}
}

@article{Lin:BVI-RLV:2024,
  title={{BVI-RLV: A} Fully Registered Dataset and Benchmarks for Low-Light Video Enhancement},
  author={R Lin and N Anantrasirichai and G Huang and J Lin and Q Sun and A Malyugina and DR Bull},
  journal={arXiv preprint arXiv:2407.03535},
  year={2024}
}

@inproceedings{Morris:DaBiT:2024,
  title={{DaBiT: Depth} and Blur informed Transformer for Video Deblurring},
  author={C Morris and N Anantrasirichai and F Zhang and D Bull},
  booktitle={IEEE/CVF Winter Conference on Applications of Computer Vision Workshop},
  year={2025}
}

@article{sun2025tenth,
  title   = {The Tenth NTIRE 2025 Image Denoising Challenge Report},
  author  = {Lei Sun and Hang Guo and Bin Ren and others},
  journal = {arXiv preprint arXiv:2504.12276},
  year    = {2025}
}

@inproceedings{fang2025guided,
  author    = {Weizhi Fang and Jingfan Fan and Yezhi Zheng and Jianbo Weng and Yijun Tai and Jun Li},
  title     = {Guided Real Image Dehazing Using YCbCr Color Space},
  booktitle = {Proceedings of the AAAI Conference on Artificial Intelligence},
  volume    = {39},
  number    = {3},
  pages     = {2906--2914},
  year      = {2025},
  doi       = {10.1609/aaai.v39i3.32297}
}

@ARTICLE{Jin:Masked:2025,
  author={Jin, Yi and Ma, Xiaoxiao and Zhang, Rui and Chen, Huaian and Gu, Yuxuan and Ling, Pengyang and Chen, Enhong},
  journal={IEEE Transactions on Multimedia}, 
  title={Masked Video Pretraining Advances Real-World Video Denoising}, 
  year={2025},
  volume={27},
  number={},
  pages={622-636},
  doi={10.1109/TMM.2024.3521818}}

@article{elharrouss2025transformer,
  author    = {Omar Elharrouss and Rania Damseh and Abdelkader Nasreddine Belkacem and Somaya Al-Maadeed and Adel Saleous and Salah Bourennane},
  title     = {Transformer-based Image and Video Inpainting: Current Challenges and Future Directions},
  journal   = {Artificial Intelligence Review},
  volume    = {58},
  pages     = {124},
  year      = {2025},
  doi       = {10.1007/s10462-024-11075-9}
}

@ARTICLE{Zhang:DVISp:2025,
  author={Zhang, Tao and Tian, Xingye and Zhou, Yikang and Ji, Shunping and Wang, Xuebo and Tao, Xin and Zhang, Yuan and Wan, Pengfei and Wang, Zhongyuan and Wu, Yu},
  journal={IEEE Transactions on Pattern Analysis and Machine Intelligence}, 
  title={{DVIS++: Improved} Decoupled Framework for Universal Video Segmentation}, 
  year={2025},
  doi={10.1109/TPAMI.2025.3552694}}

@article{HE2025102722,
title = {Deep learning based 3D segmentation in computer vision: A survey},
journal = {Information Fusion},
volume = {115},
pages = {102722},
year = {2025},
issn = {1566-2535},
doi = {https://doi.org/10.1016/j.inffus.2024.102722},
author = {Yong He and Hongshan Yu and Xiaoyan Liu and Zhengeng Yang and Wei Sun and Saeed Anwar and Ajmal Mian}
}

@article{KHEDDAR2025103347,
title = {Transformers and large language models for efficient intrusion detection systems: A comprehensive survey},
journal = {Information Fusion},
volume = {124},
pages = {103347},
year = {2025},
issn = {1566-2535},
doi = {https://doi.org/10.1016/j.inffus.2025.103347},
note = {https://www.sciencedirect.com/science/article/pii/S1566253525004208},
author = {Hamza Kheddar}
}

@ARTICLE{Zhang:DiffAD:2025,
  author={Zhang, Hui and Wang, Zheng and Zeng, Dan and Wu, Zuxuan and Jiang, Yu-Gang},
  journal={IEEE Transactions on Pattern Analysis and Machine Intelligence}, 
  title={{DiffusionAD: Norm}-Guided One-Step Denoising Diffusion for Anomaly Detection}, 
  year={2025},
  volume={},
  number={},
  pages={1-13},
  doi={10.1109/TPAMI.2025.3570494}}

@article{WU2025102965,
title = {{MAFCD: Multi}-level and adaptive conditional diffusion model for anomaly detection},
journal = {Information Fusion},
volume = {118},
pages = {102965},
year = {2025},
issn = {1566-2535},
doi = {https://doi.org/10.1016/j.inffus.2025.102965},
note = {https://www.sciencedirect.com/science/article/pii/S1566253525000387},
author = {Zhichao Wu and Li Zhu and Zitao Yin and Xirong Xu and Jianmin Zhu and Xiaopeng Wei and Xin Yang}
}

@article{tian2025yolov12,
  title={YOLOv12: Attention-Centric Real-Time Object Detectors},
  author={Tian, Yunjie and Ye, Qixiang and Doermann, David},
  journal={arXiv preprint arXiv:2502.12524},
  year={2025}
}

@article{LIU:DSEM:2025,
title = {{DSEM-NeRF: Multimodal} feature fusion and global–local attention for enhanced 3D scene reconstruction},
journal = {Information Fusion},
volume = {115},
pages = {102752},
year = {2025},
issn = {1566-2535},
doi = {https://doi.org/10.1016/j.inffus.2024.102752},
note = {https://www.sciencedirect.com/science/article/pii/S156625352400530X},
author = {Dong Liu and Zhiyong Wang and Peiyuan Chen}
}

@inproceedings{kong2025efficient,
  author    = {Hongwei Kong and Xiaodong Yang and Xiaogang Wang},
  title     = {Efficient Gaussian Splatting for Monocular Dynamic Scene Rendering via Sparse Time-Variant Attribute Modeling},
  booktitle = {Proceedings of the AAAI Conference on Artificial Intelligence},
  volume    = {39},
  number    = {4},
  pages     = {4374--4382},
  year      = {2025}
}

@ARTICLE{Yue:RViDeformer:2025,
  author={Yue, Huanjing and Cao, Cong and Liao, Lei and Yang, Jingyu},
  journal={IEEE Transactions on Circuits and Systems for Video Technology}, 
  title={{RViDeformer: Efficient} Raw Video Denoising Transformer with a Larger Benchmark Dataset}, 
  year={2025},
  volume={},
  number={},
  pages={1-1},
  doi={10.1109/TCSVT.2025.3553160}}

@inproceedings{feng2025residual,
  author    = {Haotian Feng and Haoran Zhou and Tian Ye and Shuntao Chen and Linghao Zhu},
  title     = {Residual Diffusion Deblurring Model for Single Image Defocus Deblurring},
  booktitle = {Proceedings of the AAAI Conference on Artificial Intelligence},
  volume    = {39},
  number    = {3},
  pages     = {2960--2968},
  year      = {2025},
  doi       = {10.1609/aaai.v39i3.32303}
}

@article{Azzarelli:Reviewing:2024,
  title={Intelligent Cinematography: a review of AI research for cinematographic production},
  author={A Azzarelli and N Anantrasirichai and DR Bull},
  journal={Artificial Intelligence Review},
  year={2025},
volume = {58},
number = {108}
}

@article{Feng_Ma_2025, title={{DiT4Edit: Diffusion} Transformer for Image Editing}, volume={39}, note={https://ojs.aaai.org/index.php/AAAI/article/view/32304}, DOI={10.1609/aaai.v39i3.32304}, number={3}, journal={Proceedings of the AAAI Conference on Artificial Intelligence}, author={Feng, Kunyu and Ma, Yue and Wang, Bingyuan and Qi, Chenyang and Chen, Haozhe and Chen, Qifeng and Wang, Zeyu}, year={2025}, month={Apr.}, pages={2969-2977} }

@inproceedings{kang2025exploring,
  title={Exploring Enhanced Contextual Information for Video-Level Object Tracking}, 
  author={Ben Kang and Xin Chen and Simiao Lai and Yang Liu and Yi Liu and Dong Wang},
  booktitle={Proceedings of the AAAI Conference on Artificial Intelligence},
  year={2025}
}

@article{Liu_Ma_2025, title={{LLM4GEN: Leveraging} Semantic Representation of LLMs for Text-to-Image Generation}, volume={39}, note={https://ojs.aaai.org/index.php/AAAI/article/view/32588}, DOI={10.1609/aaai.v39i5.32588}, number={5}, journal={Proceedings of the AAAI Conference on Artificial Intelligence}, author={Liu, Mushui and Ma, Yuhang and Yang, Zhen and Dan, Jun and Yu, Yunlong and Zhao, Zeng and Hu, Zhipeng and Liu, Bai and Fan, Changjie}, year={2025}, month={Apr.}, pages={5523-5531} }

@article{wang2025lavie,
  title   = {{LaVie: High}-Quality Video Generation with Cascaded Latent Diffusion Models},
  author  = {Yaohui Wang and Xinyuan Chen and Xin Ma and Shangchen Zhou and Ziqi Huang and Yi Wang and Ceyuan Yang and Yinan He and Jiashuo Yu and Peiqing Yang and Yuwei Guo and Tianxing Wu and Chenyang Si and Yuming Jiang and Cunjian Chen and Chen Change Loy and Bo Dai and Dahua Lin and Yu Qiao and Ziwei Liu},
  journal = {International Journal of Computer Vision},
  volume  = {133},
  number  = {5},
  pages   = {3059--3078},
  year    = {2025},
  doi     = {10.1007/s11263-024-02295-1},
}

@inproceedings{wu2025customcrafter,
  author    = {Tao Wu and Yong Zhang and Xintao Wang and Xianpan Zhou and Guangcong Zheng and Zhongang Qi and Ying Shan and Xi Li},
  title     = {{CustomCrafter: Customized Video Generation with Preserving Motion and Concept Composition Abilities}},
  booktitle = {Proceedings of the AAAI Conference on Artificial Intelligence},
  volume    = {39},
  number    = {8},
  pages     = {8469--8477},
  year      = {2025},
  doi       = {10.1609/aaai.v39i8.32914}
}

@inproceedings{junkawitsch2025eva,
  author    = {Hendrik Junkawitsch and Guoxing Sun and Heming Zhu and Christian Theobalt and Marc Habermann},
  title     = {{EVA: Expressive Virtual Avatars from Multi-view Videos}},
  booktitle = {ACM SIGGRAPH 2025 Conference Proceedings},
  year      = {2025},
  publisher = {ACM},
}

@ARTICLE{Huang:Diff:2025,
  author={Huang, Yi and Huang, Jiancheng and Liu, Yifan and Yan, Mingfu and Lv, Jiaxi and Liu, Jianzhuang and Xiong, Wei and Zhang, He and Cao, Liangliang and Chen, Shifeng},
  journal={IEEE Transactions on Pattern Analysis and Machine Intelligence}, 
  title={Diffusion Model-Based Image Editing: A Survey}, 
  year={2025},
  volume={47},
  number={6},
  pages={4409-4437},
  doi={10.1109/TPAMI.2025.3541625}}

@INPROCEEDINGS{Evans:stable:2025,
  author={Evans, Zach and Parker, Julian D. and Carr, CJ and Zukowski, Zack and Taylor, Josiah and Pons, Jordi},
  booktitle={ICASSP 2025 - 2025 IEEE International Conference on Acoustics, Speech and Signal Processing (ICASSP)}, 
  title={Stable Audio Open}, 
  year={2025},
  volume={},
  number={},
  pages={1-5},
  doi={10.1109/ICASSP49660.2025.10888461}}

@ARTICLE{10521791,
  author={Fei, Ben and Xu, Jingyi and Zhang, Rui and Zhou, Qingyuan and Yang, Weidong and He, Ying},
  journal={IEEE Transactions on Visualization and Computer Graphics}, 
  title={3D Gaussian Splatting as New Era: A Survey}, 
  year={2024},
  volume={},
  number={},
  pages={1-20},
  doi={10.1109/TVCG.2024.3397828}}

@article{Zhu:INFP:2024,
  author    = {Yongming Zhu and Longhao Zhang and Zhengkun Rong and Tianshu Hu and Shuang Liang and Zhipeng Ge},
  title     = {{INFP: Audio}-Driven Interactive Head Generation in Dyadic Conversations},
  journal   = {arXiv:2412.04037},
  year      = {2024}
}

@misc{zhu2024vision,
  title={Vision mamba: Efficient visual representation learning with bidirectional state space model},
  author={Zhu, Lianghui and Liao, Bencheng and Zhang, Qian and Wang, Xinlong and Liu, Wenyu and Wang, Xinggang},
  journal={arXiv preprint arXiv:2401.09417},
  year={2024}
}

\end{document}